\documentclass[lettersize,journal]{IEEEtran}
\usepackage{amsmath}
\usepackage{booktabs}
\usepackage[switch]{lineno}
\usepackage{multirow}
\usepackage{xcolor}
\definecolor{deepred}{RGB}{200,0,0}

\usepackage{graphicx}
\usepackage{bbding}
\usepackage{bm}
\usepackage{amsthm}
\usepackage{subfigure}
\usepackage{amsfonts}
\usepackage{algorithmic}
\usepackage{algorithm}
\usepackage{enumerate}
\newtheorem{remark}{Remark}
\newtheorem{theorem}{\indent Theorem}
\newtheorem*{prf}{\indent Proof}
\newtheorem{lemma}{Lemma}
\usepackage{hyperref} 
\usepackage{placeins}
\usepackage{titlesec}

\usepackage{makecell}
\usepackage{array}
\usepackage{xspace}
\usepackage{orcidlink} 
\usepackage[table]{xcolor}

\makeatletter
\DeclareRobustCommand\onedot{\futurelet\@let@token\@onedot}
\def\@onedot{\ifx\@let@token.\else.\null\fi\xspace}

\makeatother

\newtheorem{definition}{Definition}

\hypersetup{
    colorlinks=true, 
    linkcolor=blue,  
    citecolor=blue,   
    urlcolor=blue     
}

\begin{document}

\title{One-Shot Federated Clustering of \\Non-Independent Completely Distributed Data}

\author{
    Yiqun Zhang,~\IEEEmembership{Senior Member,~IEEE}, 
    Shenghong Cai, 
    Zihua Yang, 
    Sen Feng,\\ 
    Yuzhu Ji,~\IEEEmembership{Member,~IEEE},
    Haijun Zhang,~\IEEEmembership{Senior Member,~IEEE}

    \thanks{Manuscript received 18 August 2025; revised 26 October 2025 and 10 December 2025; accepted 31 December 2025. This work was supported in part by the National Natural Science Foundation of China under Grants: 62476063 and 62302104; the National Key Research and Development Program of China under Grant: 2025YFE0101100; the Natural Science Foundation of Guangdong Province under Grants: 2025A1515011293 and 2023A1515012884; the Science and Technology Program of Guangzhou under Grant: SL2023A04J01625; and the Shenzhen Science and Technology Program under Grant: JCYJ20240813104843058. \textit{(Corresponding author: Yuzhu Ji; Haijun Zhang.)}}
    \thanks{Yiqun Zhang, Zihua Yang, Sen Feng, and Yuzhu Ji are with the School of Computer Science and Technology, Guangdong University of Technology, Guangzhou 510006, China (e-mail: yqzhang@gdut.edu.cn; \{3122004153, 2112305084\}@mail2.gdut.edu.cn; yuzhu.ji@gdut.edu.cn)}
    \thanks{Shenghong Cai is with the Department of Computer Science, Beijing Normal-Hong Kong Baptist University, Zhuhai 519087, Guangdong, China (e-mail: lsshenghongcai@bnbu.edu.cn)}
    \thanks{Haijun Zhang is with the Department of Computer Science, Harbin Institute of Technology, Shenzhen 518000, China. (e-mail: hjzhang@hit.edu.cn).}
}

\maketitle

\markboth{IEEE Internet of Things Journal, January~2026}%
{One-Shot Federated Clustering of Non-Independent Completely Distributed Data}

\begin{abstract}
Federated Learning (FL) that extracts data knowledge while protecting the privacy of multiple clients has achieved remarkable results in distributed privacy-preserving IoT systems, including smart traffic flow monitoring, smart grid load balancing, and so on. Since most data collected from edge devices are unlabeled, unsupervised Federated Clustering (FC) is becoming increasingly popular for exploring pattern knowledge from complex distributed data.
However, due to the lack of label guidance, the common Non-Independent and Identically Distributed (Non-IID) issue of clients have greatly challenged FC by posing the following problems: 
How to fuse pattern knowledge (i.e., cluster distribution) from Non-IID clients;
How are the cluster distributions among clients related; and 
How does this relationship connect with the global knowledge fusion?
In this paper, a more tricky but overlooked phenomenon in Non-IID is revealed, which bottlenecks the clustering performance of the existing FC approaches. That is, different clients could fragment a cluster, and accordingly, a more generalized Non-IID concept, i.e., Non-ICD (Non-Independent \textbf{Completely} Distributed), is derived.
To tackle the above FC challenges, a new framework named GOLD (Global Oriented Local Distribution Learning) is proposed. GOLD first finely explores the potential incomplete local cluster distributions of clients, then uploads the distribution summarization to the server for global fusion, and finally performs local cluster enhancement under the guidance of the global distribution.
Extensive experiments, including significance tests, ablation studies, scalability evaluations, qualitative results, etc., have been conducted to show the superiority of GOLD.
\end{abstract}
\IEEEpeerreviewmaketitle

\section{Introduction}

\begin{figure}
    \centering
    \subfigure[FC with Non-IID clients (a1) and Non-ICD clients (a2).]{\includegraphics[width=0.95\linewidth]{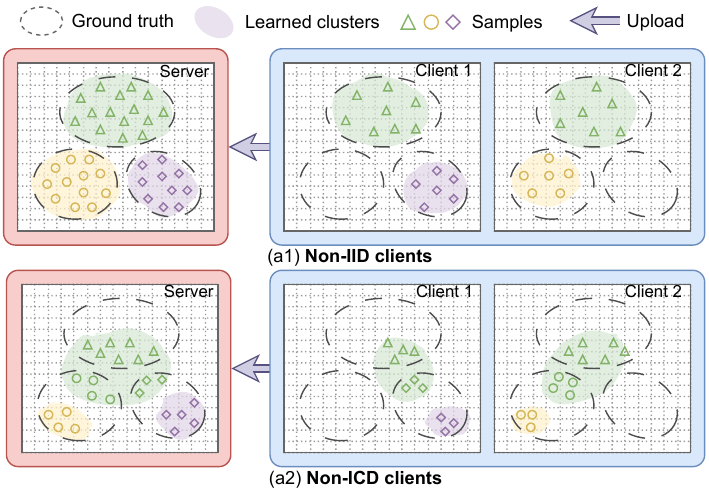}}
    \subfigure[Purity (left) and ARI (right) performance comparison of the proposed GOLD with existing counterparts under different degrees of Non-ICD ($\lambda$).]{\includegraphics[width=0.95\linewidth]{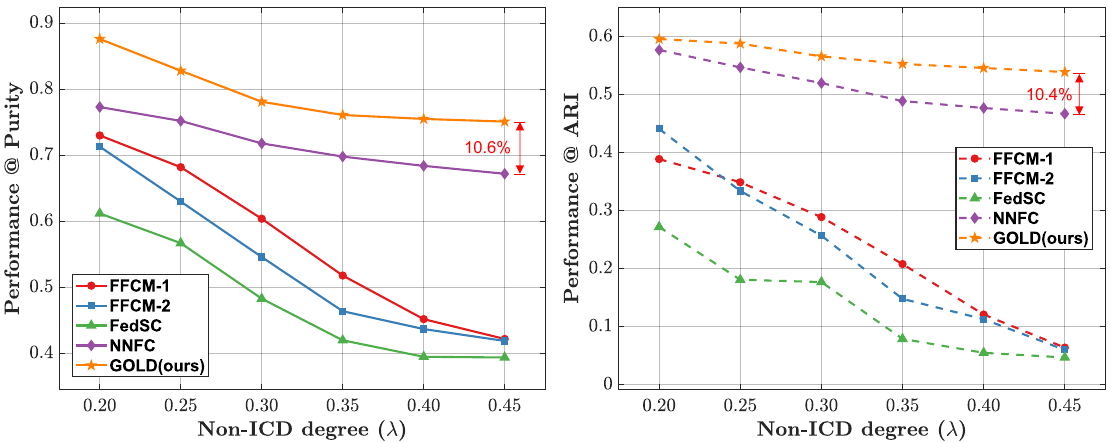}}
    \caption{Non-ICD phenomenon and its impact on Federated Clustering (FC) performance. Most FC studies address Non-IID by assuming that the global optimal number of clusters is known, and clusters of each client are balanced in scale, as shown in (a1). By contrast, a real cluster may be incomplete on certain clients, and clients may contain clusters that comprise non-adjacent subclusters as indicated in (a2). As a result, existing FC approaches can be easily misled by the Non-ICD to form improper clusters at the server. As can be seen in (b), the clustering performance of existing state-of-the-art FC methods decreases sharply with the increase of Non-ICD degree.}
    \label{fig:nonicd}
\end{figure}

\IEEEPARstart{F}{ederated} Learning (FL) aims to extract knowledge from the data of multiple clients, while protecting the clients' privacy \cite{yang2019federated,chen2022feddual}. Recently, FL plays a critical role in real data analysis systems such as edge device networks, where resolving the trade-off among computational efficiency, privacy protection, and elimination of data silos are essential \cite{liu2023fedforgery,yang2024federated}. As the number of edge devices increases and data continues to be collected in parallel, sample labeling becomes extremely expensive, hindering the exploration and understanding of data distribution, and thus limits the application of FL \cite{zhou2022pflf,zhang2024prototype,le2023privacy}. Under such circumstances, the role of Federated Clustering (FC) becomes more prominent for achieving universal FL \cite{briggs2020federated}. The common goal of FC tasks is to fuse data information of clients to obtain global cluster distributions while protecting the sample-level privacy of clients~\cite{ye2023adaptive}. The main FC challenges can be summarized as: 1) The lack of label guidance and the communication restriction prevent clients from effectively cooperating in learning cluster distributions, and 2) the common Non-IID data~\cite{qin2023fedapen,chen2024lightweight} of clients further exacerbates the difficulty of 1).

The current FC methods extract distribution information from the clients and then upload it to the server to seek complementarity of the cluster distributions on each client~\cite{li2021survey,li2022federated}.
In the literature, sparse subspace clustering and generative models have been extended for FC by amplifying the statistical information of client data into a new metric space, thus ensuring privacy protection in iterative FL. Spectral Clustering (SC) is also considered promising in FC as its feature reconstruction can simultaneously serve the privacy protection and high-dimensional data processing needs of FC \cite{wang2020federated,hernandez2021federated}. A more advanced SC-based approach named FedSC \cite{qiao2024federated} further introduces a kernel-based non-linear matrix factorization technique to serve the client's local feature extraction, and performs global reconstruction with differential privacy protection on the server. To enhance the robustness of FC in coping with Non-IID clients, FFCM \cite{stallmann2022towards} has been proposed to adapt the soft-partitional Fuzzy $c$-Means (FCM) clustering algorithm \cite{pedrycz2021federated} into the FC framework. Since all the above FC approaches involve multi-round client-server communications for iterative optimization, the higher computation overhead and privacy leakage risks limit their applications.

To better protect privacy and lower communication costs, one-shot FC paradigm that achieves global clustering with one-time communication has attracted more attention~\cite{yang2022k,li2021survey}. $k$-Fed \cite{dennis2021heterogeneity} is a representative one-shot extension of $k$-means for FC. $k$-Fed first performs independent local clustering on each client and then uploads the cluster centers to the server for global distribution combination. Most recently, OSFSC \cite{xie2023fed} further extends SC into a one-shot FC approach to overcome the problem of degraded clustering performance on high-dimensional data. Both $k$-Fed and OSFSC address the challenge of Non-IID by assuming that the global cluster distribution can be complemented by a sufficient number of clients. However, the number of active clients is usually not guaranteed in real applications. 
Moreover, most of the existing FC approaches implicitly assume that the true number of clusters $k^*$ is given in advance, which is usually not the case in practical FC tasks.

Recent FC advances~\cite{wang2025one,0003SPRM23} further enhance clustering performance while ensuring data privacy, but typically assuming each cluster distribution is locally complete, as illustrated in Fig.~\ref{fig:nonicd}(a1). In fact, the problem of incomplete clusters is pervasive, caused by the multi-granularity nature of real-world clusters. That is, a coarse-grained cluster at a high abstraction level usually consists of several fine-grained micro-clusters (also called subclusters interchangeably). The subclusters may be distributed across clients, complicating the formation of a global, coarse-grained clusters in FC.
For example, in smart home environments, a user behavior cluster can be fragmented across different IoT devices: motion sensors capture activity patterns in specific rooms, smart appliances record usage behaviors in the kitchen, while thermostats observe temperature control patterns. As a result, each client captures distinct-grained subclusters, making clusters often searched based on unaligned clients' granularities and perspectives on the server, as shown in Fig.~\ref{fig:nonicd}(a2).

To intuitively reveal the common yet overlooked incomplete cluster distribution pattern, the average Jensen-Shannon divergence between the cluster distributions of clients (denoted as $\lambda$ and formally defined with analysis in Section~\ref{sec:problem}) can be utilized to quantify the degree of cluster incompleteness across clients. A higher $\lambda$ indicates greater cluster fragmentation and more severe granularity misalignment across clients, which is formally named as \textbf{Non-Independent Completely Distribution (Non-ICD)}. The vulnerability of existing methods under Non-ICD issue can be intuitively illustrated through a demo experiment as shown in Fig.~\ref{fig:nonicd}(b). With gradually increasing $\lambda$, state-of-the-art FC methods experience steep performance degradation in both Purity and ARI metrics. In contrast, the proposed GOLD method exhibits a moderate declining trend and achieves at least 10\% improvement compared to the others. This clearly demonstrates that the Non-ICD phenomenon poses a significant bottleneck to practical FC tasks, and a new Non-ICD FC method is urgently needed to bridge the application gap of existing FC approaches.

This paper studies the Non-ICD problem and proposes a novel one-shot FC framework called \textbf{G}lobal \textbf{O}riented \textbf{L}ocal \textbf{D}istribution learning (GOLD)
to address the challenges brought by the Non-ICD and the unknown optimal number of clusters $k^*$. 
A concise design lies in that the Competitive Penalized Learning (CPL) runs through the entire cluster exploration process across the client and server, letting candidate clusters compete with each other to attract data samples. The clusters with low compactness can thus be iteratively eliminated, thereby automatically obtaining a proper number of compact clusters. From the client aspect, CPL captures dense micro-clusters, circumventing the risk of mis-partitioning of compact distributions. From the server aspect, the recursively launched CPL can explore a series of proper $k^*$s corresponding to different cluster granularities.
One-shot communication strategy is further adopted to relieve the privacy leakage risk caused by the finely explored nested clusters, whilst the multi-granular cluster structure exploration at the server effectively hedges the information loss caused by the one-shot design. As a result, GOLD effectively enhances the FC performance on the server and clients. More importantly, it yields an interpretable nested cluster relationship to reveal complex data distributions. The main contributions of this work are summarized in four-fold:

\begin{enumerate}
    \item The problem of Non-ICD that a cluster distribution can be fragmented by different clients has been defined and explored for the first time in FL. It is revealed that the common Non-ICD is a generalized but more challenging form of Non-IID, and bottlenecks the performance of FC.
    \item A new FC framework that simultaneously enhances accuracy and security has been proposed. It adopts a one-shot client-server communication mechanism with low information throughput for efficient learning, and employs CPL for sophisticated and robust FC.
    \item CPL is flexibly utilized on both the client and server sides. That is, Fine-grained CPL (FCPL) at clients for personalized cluster distribution exploration, and Multi-granularity CPL (MCPL) at the server for Non-ICD clients' distribution integration.
    \item A new insight that optimal $k^*$s may exist at different granularities guides the informative data encoding for representation enhancement. Comprehensive experiments validate the soundness of the learned $k^*$s, and their contributions in facilitating accurate and interpretable FC. 
\end{enumerate}

\section{Related Work}
This section reviews the federated clustering, Non-IID data clustering, and clustering with an unknown number of clusters, that are highly related to the research topic of this paper.

\subsection{Distributed and Federated Clustering}

The evolution from distributed to FC reflects the growing tension between data utility and privacy in IoT deployments. Early distributed clustering methods like distributed matrix factorization-based clustering \cite{wang2020federatedArxiv}, an extension for unsupervised learning, have shown promise in large-scale applications. Parallel $k$-Means \cite{kumar2020federated} and DSC \cite{hernandez2021federated} leveraged multi-processor architectures to compute pairwise distances without centralizing data, yet still assumed trusted communication channels and homogeneous feature spaces. As privacy regulations tightened, these assumptions became untenable, leading to the federated paradigm in which only model parameters rather than raw data cross organizational boundaries.

This shift brought new algorithmic challenges. FC methods specifically tackle these challenges by focusing on maintaining data privacy while improving clustering performance. FedSC \cite{qiao2024federated} pioneers kernelized decomposition to approximate global similarity matrices without sharing local kernels, maintaining $O(n^2)$ complexity locally while achieving near-centralized accuracy. Fed-FCM \cite{pedrycz2021federated} and FFCM \cite{stallmann2022towards} take a simpler path by iteratively refining fuzzy membership matrices via weighted centroid averaging, trading optimality for interpretability. More recently, FeMIFuzzy \cite{ngo2023federated} pushes further by handling incomplete longitudinal data via multiple imputation and Sammon mapping in just two rounds. Yet a fundamental limitation persists: these methods still require $O(r\cdot c)$ communication rounds (where $r$ denotes the number of iterations and $c$ denotes the number of clients), creating both a performance bottleneck and an expanding attack surface. This motivates the emergence of one-shot approaches.

One-shot federated clustering completes clustering through a single round of client-server communication, reducing communication overhead (from $O(r \cdot c)$ to $O(c)$) and minimizing privacy exposure.
$k$-Fed \cite{dennis2021heterogeneity} achieves this by precomputing local clusters and transmitting only centroids, relying on Hungarian matching for alignment. 
OSFSC \cite{xie2023fed} adopts a subspace-based approach, learning a unified affinity matrix via spectral regularization to preserve local manifold structure. 
NN-FC \cite{wang2024one} adds differential privacy via Laplacian noise while maintaining stable distance relationships.
The existing advanced one-shot FC solutions either assume cluster correspondence ($k$-Fed) \cite{dennis2021heterogeneity}, enforce global structure (OSFSC) \cite{xie2023fed}, preserve only pairwise relationships (NN-FC) \cite{wang2024one}, adopt machine unlearning techniques for privacy protection enhancement~\cite{0003SPRM23} or determine global centroids by aggregating locally extracted density cores~\cite{wang2025one}. However, none of them considers the multi-granular cluster reality. For example, Client $A$ may observe cluster $C$ at fine resolution as several small subclusters $(c_1, c_2, \ldots)$, while Client $B$ observes cluster $C$ at a coarse resolution as a single aggregate. 
Such granularity misalignment, combined with cluster incompleteness, fundamentally bottlenecks the existing FC methods.

\subsection{Federated Clustering of Non-IID Data}
Non-IID is a key challenge of practical FC \cite{li2023differentially}, and early attempts to solve such a problem include spectral clustering \cite{wang2020federated,hernandez2021federated} and kernelized matrix decomposition \cite{qiao2024federated,yin2021comprehensive}. These methods typically assume complete yet biased clients' data distributions, and reconcile discrepancies through kernel alignment or shared low-rank factorization. 
IFCA \cite{ghosh2020efficient} and its neural extension UIFCA \cite{chung2022federated} model this explicitly by partitioning clients into meta-clusters based on their local distributions. These approaches are effective when heterogeneity follows clear patterns but struggle with arbitrary fragmentation. More sophisticated approaches recognize that clients might observe overlapping, partial cluster structures. FLSC \cite{li2021federated} allows soft client-cluster assignments, while hierarchical methods \cite{ma2024feduc} aggregate at multiple scales to capture both local and global patterns.

\subsection{Clustering with Unknown Number of Clusters}

In centralized settings, practitioners often sweep over candidate $k$ and select the optimal one using validity indices (e.g., Silhouette) or heuristics (e.g., Elbow). Density-based approaches \cite{zhangcais2024} trace how density peaks \cite{zou2025sdenk} emerge and merge, then identify a knee point~\cite{zhang2025federated} to determine $k^*$. $k$-means++ improves seeding to reduce sensitivity to $k$, while significance-based techniques \cite{hu2025significance} judge the significance of cluster distributions under the current $k$. AFCL \cite{zhang2025asynchronous} removes the fixed-$k$ assumption and discovers clusters asynchronously as clients join and leave, making it well-suited to dynamic IoT networks. However, most criteria select $k$ from a global, centralized perspective, whereas Non-ICD disperses structure across clients so that $k$ becomes identifiable only after cross-client fusion. Recently, learning-based methods explore a proper number of clusters by allowing cluster centers to be added, merged~\cite{zhang2025learningSOM}, or pruned during the data representation structure training \cite{cai2024robust}. Since clustering without knowing the optimal number of clusters remains a challenge in the clustering domain, the above techniques have not yet been adapted to the FC of Non-ICD data.

\section{Proposed Method}

In this section, we first formulate the FC problem under Non-ICD data, and then introduce the proposed GOLD framework, which consists of two key technical components: 1) Fine-grained Competitive Penalized Learning (FCPL) for local micro-cluster exploration on clients, and 2) Multi-granular Competitive Penalized Learning (MCPL) for global distribution aggregation on the server. In addition, an informative data encoding method called Representation Enhancement based on Multi-Granular Clusters (REMC) is proposed. The overview of GOLD is shown in Fig.~\ref{fig:method_diagram}, and the frequently used notations are summarized in Appendix~\ref{ap:1}.

\begin{figure*}[t]
    \centering
    \includegraphics[width=0.95\linewidth]{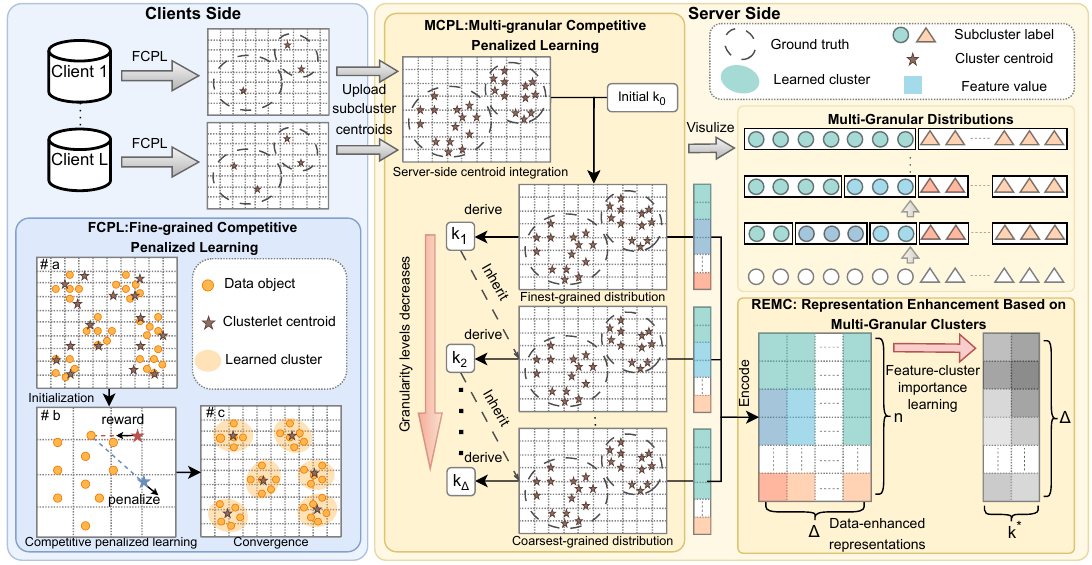}
    \caption{Overview of GOLD. Each client first employs FCPL to extract compact local subclusters from Non-ICD data, then transmits only the corresponding centroids to the server. The server executes MCPL, enabling the uploaded centroids to naturally aggregate across multiple granularity levels.
    The resulting multi-granular distributions are further encoded into an enhanced representation to drive the final feature-cluster importance learning-based clustering.}
    \label{fig:method_diagram}
\end{figure*}

\subsection{Problem Formulation}
\label{sec:problem}
Consider $L$ clients, indexed by $l \in \{ 1, \dots, L \}$, each with a private local dataset $\mathbf{X}^{(l)} \in \mathbb{R}^{n^{(l)} \times d}$, where $n^{(l)}$ is the number of local data objects and $d$ is the feature dimensionality. 
The cluster set of all data objects is denoted by $C = \{ C_j \mid 1 \le j \le k^* \}$, referred to as the target global clusters. This paper focuses on FC of Non-ICD data, where the local datasets from different clients are heterogeneous in terms of their cluster distributions. Accordingly, we first formally define the concept of Non-ICD and provide a quantity $\lambda$ to indicate its degree, and then formalize the FC problem with an objective function.

\begin{definition}
\label{def:non-icd}
\textbf{Non-ICD data distribution:}
    Given datasets $\mathbf{X}^{(l)} (1 \le l \le L)$ residing on $L$ clients. Suppose the clients demonstrate cluster incompleteness, i.e., a global cluster $C_j$ is decomposed into a set of subclusters at different granularities, dispersing across multiple clients. Since clients with incomplete clusters do not possess a complete view of $C_j$, the clients are deemed Non-ICD if one of the following conditions is satisfied:
    1) Granularity misalignment. Clients may have subclusters from the same global cluster $C_j$ at different levels of granularity, leading to inconsistent representations of the same cluster across clients.
    2) Non-adjacent subclusters. Subclusters belonging to the same global cluster $C_j$ may appear in non-adjacent regions of the feature space on the same client. Such separation can mislead the local learner to treat subclusters from the same source cluster as distinct clusters. 
\end{definition}

The Jensen-Shannon (JS) distance~\cite{sutter2020multimodal} is employed to quantify the degree of Non-ICD as the average information discrepancy between the probabilistic distributions of each pair of clients, which can be written as:
\begin{equation}
\label{lambda}
    \lambda = \frac{1}{L(L - 1)}  \sum_{1 \leq a < b \leq L} JS\left ( \hat{f}(\mathbf{X}^{(a)}), \hat{f}(\mathbf{X}^{(b)}) \right),
\end{equation}
where $\hat{f}(\mathbf{X}^{(l)})$ denotes the probabilistic distribution of the $l$-th client, estimated via Gaussian Kernel Density Estimation (KDE)~\cite{heer2021fast}, which is a non-parametric method that smoothly approximates the underlying probability density from observed samples. Specifically, $\hat{f}(\mathbf{X}^{(l)}) = \left [ \hat{f}(x^{(l)}_{1}),  \hat{f}(x^{(l)}_{2}), \dots \hat{f}(x^{(l)}_{n^{(l)}}) \right ]$, where each entry $\hat{f}(x^{(l)})$ is computed as:
\begin{equation}
\label{eq:kde}
\hat{f}(x^{(l)}) = \frac{1}{n^{(l)}h}  \sum_{t=1}^{n^{(l)}} \frac{1}{\sqrt{2 \pi h^2}} \exp \left(-\frac{(x^{(l)}-x_t^{(l)})^2}{2h^2}\right), 
\end{equation}
with $h$ denoting the bandwidth selected via cross-validated likelihood. After estimating $\hat{f}(\mathbf{X}^{(l)})$ for each client, the JS distance between the $a$-th and $b$-th clients is computed as:
\begin{align}
\label{eq:JS}
JS\left ( \hat{f}(\mathbf{X}^{(a)}), \hat{f}(\mathbf{X}^{(b)}) \right)
&= \frac{1}{2} \Big(
    KL(\hat{f}(\mathbf{X}^{(a)}) \| \hat{f}^{(a,b)}) \nonumber \\
&\quad + KL(\hat{f}(\mathbf{X}^{(b)}) \| \hat{f}^{(a,b)})
\Big),
\end{align}
where $\hat{f}^{(a,b)} = \frac{1}{2}\left( \hat{f}(\mathbf{X}^{(a)}) + \hat{f}(\mathbf{X}^{(b)}) \right)$ denotes the mean distribution of the two clients, and $KL(\hat{f}(\mathbf{X}^{(a)})||\hat{f}^{(a,b)})$ represents the Kullback-Leibler (KL) divergence~\cite{kim2021comparing}, defined as:
\begin{equation}
\label{eq:kl}
    KL(\hat{f}(\mathbf{X}^{(a)})||\hat{f}^{(a,b)}) = \sum_{x^{(a)} \in \mathbf{X}^{(a)}} \hat{f}(x^{(a)}) \log_2 \frac{\hat{f}(x^{(a)})}{\hat{f}^{(a,b)}}.
\end{equation}

\begin{remark}
\textbf{Physical meaning of the Non-ICD degree measure $\lambda$:}
According to Eqs.~(\ref{lambda})-(\ref{eq:kl}), $\lambda$ is in the interval $[0,1]$, indicating the averaged pairwise JS distances.
%, where the JS distance is computed through Eqs.~(\ref{eq:JS}) and (\ref{eq:kl}). 
Specifically, the value of $\lambda$ is dominated by the key term $\frac{\hat{f}(x^{(a)})}{\hat{f}^{(a,b)}}$ in Eq.~(\ref{eq:kl}), which is analyzed in the following three cases.
\textbf{(1) $\lambda = 0$} holds iff $\frac{\hat{f}(x^{(a)})}{\hat{f}^{(a,b)}} = 1$ for every pair of clients. That is, all clients exhibit identical data density at all locations in the metric space, indicating that the clients observe data from the same underlying distribution, corresponding to the pure IID scenario.
\textbf{(2) $\lambda \to 1$} occurs when $\frac{\hat{f}(x^{(a)})}{\hat{f}^{(a,b)}} = 2$ for any pair of the $L$ clients. This implies that every data object $x$ satisfies $\hat{f}(x^{(a)}) > 0$ and $\hat{f}(x^{(b)}) = 0$ for all $b \neq a$. Since each $x$ is observable by a single client, $\lambda \to 1$ indicates that the local datasets of any two clients tend to be completely disjoint, corresponding to the extreme Non-IID scenario.
\textbf{(3) $0 < \lambda < 1$} occurs when $1 \le \frac{\hat{f}(x^{(a)})}{\hat{f}^{(a,b)}} \le 2$, where $\hat{f}(x^{(a)}) > 0$, $\hat{f}(x^{(b)}) > 0$, and $\hat{f}(x^{(a)}) \neq \hat{f}(x^{(b)})$ hold. This case implies that the same data object follows heterogeneous underlying density distributions across the $a$-th and $b$-th clients. Moreover, a larger difference between $\hat{f}(x^{(a)})$ and $\hat{f}(x^{(b)})$ indicates a greater inconsistency in the cluster granularity levels corresponding to the same global cluster across clients. Meanwhile, under the case of $0 < \lambda < 1$, partial of the data objects may still satisfy $\frac{\hat{f}(x^{(a)})}{\hat{f}^{(a,b)}} = 2$, which indicates that these data objects are only observed by the $a$-th client and can not be observed by other clients. 
This corresponds to the incomplete clusters on clients, which is a typical characteristic of the Non-ICD.
\end{remark}

In FC tasks, the object-cluster affiliation is represented by a matrix $\mathbf{Q} \in \mathbb{R}^{n \times k}$, where each entry $q_{ij}$ satisfies:
\begin{equation}
\label{eq:affiliation}
    {\textstyle \sum_{j=1}^{k}} q_{ij} = 1 \text{ and } q_{ij} \in \{0, 1\}.
\end{equation}
The clustering objective of either clients or the server is to maximize the overall intra-cluster similarity, formulated as: 
\begin{equation}
\label{eq:objective_function}
    P(\mathbf{Q}, C) = \sum_{i=1}^{n} \sum_{j=1}^{k} q_{ij} s(x_i, C_j),
\end{equation}
where $s(x_i, C_j)$ measures the similarity between data object $x_i$ and cluster $C_j$, which can be defined as:
\begin{equation}
\label{eq:similarity}
    s(x_i,C_j)=\|\mathbf{x}_i-\mathbf{c}_j \|_2,
\end{equation}
In the one-shot federated setting, each client optimizes this objective function locally based on its dataset $\mathbf{X}^{(l)}$, and transmits the resulting cluster centroids $C^{(l)} = \{ C^{(l)}_j \mid 1 \le j \le k^{(l)} \}$ to the server for protecting privacy. The server then aggregates the received local cluster centroids and performs clustering to maximize the objective function.

\subsection{Micro-Cluster Exploration on Clients}

To enable a finer exploration of local distributions under Non-ICD data and to establish a solid foundation for effective global aggregation at the server, Fine-grained Competitive Penalized Learning (FCPL) is introduced. FCPL performs clustering without requiring prior knowledge of the number of clusters and effectively separates non-adjacent subclusters into compact, well-defined clusters.

For Non-ICD distributions, the presence of non-adjacent subclusters within a client may mislead clustering, potentially causing these subclusters to be erroneously assigned to separate clusters.
To address this, FCPL adopts Competitive Penalized Learning (CPL)~\cite{cheung2005maximum}, where candidate clusters compete during learning: prominent candidates that represent true subclusters are retained while redundant candidates are eliminated. This enables automatic determination of the optimal number of local subclusters.
In practice, for the $l$-th client, FCPL initializes a sufficiently large set of candidate clusters, 
$C^{(l)} = \{ C^{(l)}_j \mid 1 \le j \le k_0 \}$, with a large $k_0$, ensuring that subclusters are accurately captured as compact micro-clusters.
The significance of each candidate cluster is quantified by the weight set $W = \{ w_j \mid 1 \le j \le k_0 \}$, where $w_j$ denotes the weight of cluster $C^{(l)}_j$ in the overall clustering structure.
Since certain features may dominate specific clusters, a feature-cluster importance matrix $\mathbf{H} \in \mathbb{R}^{k^{(l)} \times d}$ is further introduced to enhance the measurement of object-cluster similarity, thereby redefining the similarity in Eq.~(\ref{eq:similarity}) as:
\begin{equation}
\label{eq:FCPL_similarity}
s(x^{(l)}_i,C^{(l)}_j)=\|\mathbf{h}_j \cdot (\mathbf{x}^{(l)}_i-\mathbf{c}^{(l)}_j) \|_2,
\end{equation}
where $\mathbf{h}_j$ is the feature-cluster importance vector for cluster $C^{(l)}_j$.
FCPL consists of two iterative stages: 1) cluster competition via CPL, and 2) feature-cluster importance update.

\textbf{Stage 1: Cluster Competition Via CPL}

According to the objective function in Eq.~(\ref{eq:objective_function}), each data object $x^{(l)}_i$ is assigned to its corresponding cluster, referred to as the winning cluster $C^{(l)}_v$, which is determined by:
\begin{equation}
\label{eq:v}
v = \arg \max_{1 \le j \le k_0} {\left [ \gamma_j w_j s(x^{(l)}_i, C^{(l)}_j) \right]},
\end{equation}
where $\gamma_j$ denotes the relative winning probability of $C_j^{(l)}$, computed as:
\begin{equation}
\label{eq:gamma}
\gamma_j=1-\frac{g_j}{ {\textstyle \sum_{t=1}^{k_0}g_t} }.
\end{equation}
Here, $g_j$ denotes the number of times $C^{(l)}_j$ has been selected as the winning cluster during the learning process, and it is updated as follows:
\begin{equation}
\label{eq:g}
g_v = g_v + 1.
\end{equation}
Meanwhile, the nearest rival cluster to $C_v^{(l)}$, denoted as $C_r^{(l)}$, is identified by:
\begin{equation}
\label{eq:r}
r = \arg \max_{1 \le j\le k_0, j \ne v} {\left [ \gamma_j w_j s(x^{(l)}_i, C^{(l)}_j) \right]}.
\end{equation}
\begin{remark}
\textbf{Fair Cluster Competition:}
    $\gamma_j$ mitigates the domination of frequently winning clusters by gradually reducing their relative winning probability, thereby ensuring a fair evaluation of all initialized clusters during redundant cluster elimination. Furthermore, it prevents initialized clusters located in marginal positions from being prematurely treated as ``dead units’’ without receiving any learning updates.
\end{remark}

To constrain the values of cluster weight $w_j$ within the interval $[0, 1]$, an intermediate variable set $\mathcal{W} = \{ \mathcal{W}_j \mid 1 \le j \le k_0 \}$ is introduced to indirectly update $W$. Specifically, the winning cluster $C^{(l)}_v$ is rewarded by increasing $\mathcal{W}_v$ as:
\begin{equation}
\label{eq:v_weight}
\mathcal{W}_v = \mathcal{W}_v + \eta,
\end{equation}
and the nearest rival cluster $C^{(l)}_r$ is penalized by:
\begin{equation}
\label{eq:client_penalize}
\mathcal{W}_r = \mathcal{W}_r - \eta \frac{s(x^{(l)}_i, C^{(l)}_r)}{s(x^{(l)}_i, C^{(l)}_v)},
\end{equation}
where $\eta$ is the learning rate. As a result, $C^{(l)}_v$ is rewarded by a small step $\eta$, making it more likely to be preserved as a dominant cluster, whereas $C^{(l)}_r$ is penalized by a step determined by the similarity ratio between the object and the two competing clusters.
The sigmoid is then applied to smoothly regularize $\mathcal{W}_j$ into the cluster weight $w_j$:
\begin{equation}
\label{eq:client_weight}
w_j = \frac{1}{1+e^{-10(\mathcal{W}_j + 5)}}.
\end{equation}

\textbf{Stage 2: Feature-Cluster Importance Update}

To update the feature-cluster importance matrix $\mathbf{H}$, each $(j,m)$-th entry $h_{jm}$ is determined by simultaneously considering the inter-cluster difference $\alpha^{(l)}_{jm}$ and the intra-cluster similarity $\beta^{(l)}_{jm}$. Specifically, $\alpha^{(l)}_{jm}$ quantifies the ability of the $m$-th feature to distinguish cluster $C_j^{(l)}$ from all other clusters, while $\beta^{(l)}_{jm}$ evaluates the compactness of $C_j^{(l)}$ along the $m$-th feature.
The inter-cluster difference $\alpha^{(l)}_{jm}$ is computed by comparing the distributions of the $m$-th feature inside and outside $C_j^{(l)}$. Both distributions are modeled as Gaussians, $\mathcal{N}(\mu, \sigma^2)$ and $\mathcal{N}(\bar{\mu}, \bar{\sigma}^2)$, respectively \cite{yuan2023spatio,jiang2023anomaly}. The dissimilarity is then quantified using the Hellinger distance \cite{akash2019inter,sengar2008detecting}, a bounded and symmetric metric measuring divergence between two probability distributions:
\begin{equation}
\label{eq:alpha_client}
\alpha^{(l)}_{jm} = \sqrt{1-
             \sqrt{\frac{2\sigma_{jm}\bar{\sigma}_{jm}}{\sigma_{jm}^2+\bar{\sigma}_{jm}^2}}
             \exp\left[-\frac{(\mu_{jm}-\bar{\mu}_{jm})^2}{4(\sigma_{jm}^2+\bar{\sigma}_{jm}^2)}\right]}
\end{equation}
with
\begin{equation}
\label{eq:mu_client}
\mu_{jm}=\frac{1}{\left | C^{(l)}_j \right | } \sum_{x^{(l)}_i\in C^{(l)}_j} x^{(l)}_{jm}, \ 
\bar{\mu}_{jm}=\frac{1}{\left | \bar{C}^{(l)}_j \right | } \sum_{x^{(l)}_i \notin  C^{(l)}_j} x^{(l)}_{jm},
\end{equation}
and
\begin{equation}
\label{eq:sigma1_client}
\sigma_{jm}^2=\frac{\sum_{x^{(l)}_i \in C^{(l)}_j}(x^{(l)}_{jm}-\mu_{jm})^2}{\left | C^{(l)}_j \right |-1}, 
\end{equation}
\begin{equation}
\label{eq:sigma2_cleint}
\bar{\sigma}_{jm}^2=\frac{\sum_{x^{(l)}_i \notin C^{(l)}_j}(x^{(l)}_{jm}-\bar{\mu}_{jm})^2}{\left | \bar{C}^{(l)}_j \right |-1}.
\end{equation}
The intra-cluster similarity $\beta^{(l)}_{jm}$ is estimated by the average distance within cluster $C_j^{(l)}$ along the $m$-th feature, given by:
\begin{equation}
\label{eq:beta_client}
\beta^{(l)}_{jm} = \frac{1}{|C^{(l)}_j|} \sqrt{\sum_{x^{(l)}_i\in C^{(l)}_j}\exp\left [ -0.5 (x^{(l)}_{jm}-c^{(l)}_{jm})^2 \right ]}.
\end{equation}
A large $\alpha^{(l)}_{jm}$ indicates that the $m$-th feature effectively distinguishes cluster $C^{(l)}_j$ from others, while a large $\beta^{(l)}_{jm}$ reflects high intra-cluster consistency along this feature. When both values are large, the $m$-th feature contributes significantly to the formation of $C^{(l)}_j$. Consequently, the importance $h_{jm}$ of the $m$-th feature to cluster $C^{(l)}_j$ can be calculated by:
\begin{equation}
\label{eq:client_feature_cluster_weight}
h_{jm} = \frac{\alpha^{(l)}_{jm} \beta^{(l)}_{jm}}{ {\textstyle \sum_{m=1}}^{d} \alpha^{(l)}_{jm} \beta^{(l)}_{jm}}. 
\end{equation}

To facilitate the aforementioned learning stages of FCPL, a large set of candidate clusters, $C^{(l)} = \{ C^{(l)}_j \mid 1 \le j \le k_0 \}$, is initialized with a sufficiently large $k_0$ to ensure comprehensive exploration of the local Non-ICD distributions. The two stages of FCPL are then iteratively performed until convergence. Once a cluster weight $w_j$ approaches 0, the corresponding cluster is eliminated, and its associated data objects are reassigned to the remaining candidate clusters. In this manner, the $l$-th client ultimately obtains $k^{(l)}$ clusters. To preserve privacy, only the cluster centroid set $C^{(l)} = \{ \mathbf{c}^{(l)}_j \mid 1 \le j \le k^{(l)} \}$ is uploaded to the server for global aggregation. 
\begin{remark}
\textbf{Secure Clients' Information Transmission:}
    GOLD requires a single communication round, where each client uploads its centroids $C^{(l)}$ to the server, ensuring that only a minimal amount of local information is shared across the federated network. Existing privacy-preserving techniques, such as homomorphic encryption \cite{acar2018survey} and differential privacy \cite{wei2020federated,li2023differentially}, can be incorporated into GOLD to enhance data confidentiality during the transmission of $C^{(l)}$.
\end{remark}

\subsection{Multi-Granular Cluster Learning on Server}

At this stage, the server collects cluster centroids from all clients. Stacking all local centroid sets $C^{(l)} (1 \le l \le L)$ forms a global data matrix $\mathbf{X}\in \mathbb{R}^{n \times d}$, where $n = \sum_{l = 1}^{L} n^{(l)}$ is the total number of centroids.
Due to the granularity diversity in local Non-ICD distributions, the integrated global distribution may exhibit multiple granularity levels. To mitigate potential granularity misalignment among the Non-ICD clients, MCPL is employed to automatically uncover multi-granular global distributions across different granularity levels.

To enable the global distribution to naturally form clusters at multiple granularities, it is essential to determine the number of clusters $k^*$ corresponding to each granularity level. However, these values are typically unknown. The proposed MCPL builds upon FCPL by initializing a substantially large set of candidate clusters $C = \{ C_j \mid 1 \le j \le k_0 \}$ with a large $k_0$ to capture a fine-grained cluster distribution. It then allows the global distribution to evolve into coarser-granular cluster distributions. Since the estimation of the object-cluster similarity forms the foundation of the learning process, it is essential to maintain its magnitude consistently across different granularities. This consistency enhances the capacity of MCPL to explore multi-granular global cluster distributions. Accordingly, Eq.~(\ref{eq:FCPL_similarity}) is reformulated to normalize object-cluster similarity across granularities by converting it into a probabilistic measure, as follows:
\begin{equation}
\label{eq:MCPL_similarity}
    s(x_i, C_j) = \frac {\exp\left [ -0.5\left \| \mathbf{h}_j \cdot (\mathbf{x}_i-\mathbf{c}_j) \right \|_2  \right ] } { {\textstyle \sum_{t=1}^{k_0}}\exp\left [ -0.5\left \| \mathbf{h}_j \cdot (\mathbf{x}_i-\mathbf{c}_{t}) \right \|_2  \right ]  }.
\end{equation}
FCPL leverages CPL's single-scale clustering to discover local subclusters, while MCPL exploits its recursive applicability for multi-granular global exploration. Such a same-base module design seamlessly connects the synchronization of learned distributions between the clients and server. The simple design also reduces the complexity of the model's architecture, which helps preserve interpretability throughout the FC process.

Building upon the normalized similarity measure in Eq.~(\ref{eq:MCPL_similarity}), MCPL begins with a substantially large initial cluster number $k_0$ and applies the CPL process to partition $\mathbf{X}$ into $k_1$ clusters, thereby producing the finest-grained object-cluster affiliation matrix $\mathbf{Q}_1$.
To discover coarser-grained distributions, this process is recursively applied, with each iteration initialized by inheriting the cluster number from the previously converged stage (e.g., $k_1$ obtained from the first learning stage) while reinitializing all internal learning factors, i.e., relative winning probability $\gamma_j \ (1 \le j \le k_1)$, cluster weights $W$, feature-cluster importance matrix $\mathbf{H}$. 
The recursion terminates when the cluster number stabilizes between consecutive iterations, i.e., $k_{new} = k_{old}$, and the objective function value converges, satisfying $|P_{new} - P_{old}| \le \epsilon $, where $\epsilon$  is a small threshold (see Algorithm~\ref{alg:GOLD_paper} Line 11).
As a result, MCPL produces multi-granular object-cluster affiliation matrices $Q = \{ \mathbf{Q}_1, \dots, \mathbf{Q}_{\Delta} \}$, corresponding to $\Delta$ granularity levels with cluster numbers $K = \{ k_1, \dots, k_{\Delta} \}$.

\begin{remark}
\textbf{Reinitialization for Structure Diversity:}
    At each learning stage of MCPL, only the number of clusters $k$ learned from the previous stage is inherited to initialize the next stage, but the corresponding cluster centroids are re-initialized. This repeated reinitialization promotes a thorough exploration of the global distribution, enabling each granularity level to more accurately capture the underlying structure and facilitating the discovery of multi-granular cluster distributions. In contrast, directly inheriting the converged centroids from earlier stages risks overfitting to the preceding structure, which can diminish structural diversity across different granularity levels.
\end{remark}

\subsection{Representation Enhancement Based on Multi-Granular Cluster Aggregation}
\label{sec:REME}
This section discusses how the server leverages multi-granular global cluster distributions to perform clustering with a target number of clusters $k^*$, and introduces Representation Enhancement Based on Multi-Granular Clusters (REMC), which informatively aggregates multi-granular cluster distributions to guide the clustering process.
REMC embeds the multi-granular object-cluster affiliation matrices $Q$ into a data-enhanced representation $\mathbf{X}^{(e)} \in \mathbb{R}^{n \times \Delta}$. Specifically, for an $n \times k_\delta$ matrix $\mathbf{Q}_\delta$ at granularity level $\delta$, it is encoded into the $\delta$-th feature of $\mathbf{X}^{(e)}$ by: 
\begin{equation}
\label{eq:encode}
x^{(e)}_{i\delta} = \sum_{j=1}^{k_\delta}jq_{ij}, \ \text{with} \ 1 \le i \le n, \ q_{ij} \in \mathbf{Q}_\delta.
\end{equation}

Since the $\delta$-th feature of $\mathbf{X}^{(e)}$ encodes cluster structural information of the $\delta$-th granularity level, the contribution of each granularity to the clustering process is different.
Accordingly, REMC performs clustering on $\mathbf{X}^{(e)}$ using the feature-cluster weight matrix $\mathbf{U} \in \mathbb{R}^{\Delta \times k^*}$.
Based on the objective in Eq.~(\ref{eq:objective_function}), the cluster assignment for each $x^{(e)}_i$ is determined by:
\begin{equation}
\label{eq:server_q}
q_{ij} =
\begin{cases}
1, & \text{if} \ j = \arg\max\limits_{1 \le t \le k^*} \ s(x_i^{(e)}, C_t) \\
0, & \text{otherwise},
\end{cases}
\end{equation}
where $s(x_i^{(e)}, C_t)$ computes the similarity between $x^{(e)}_i$ and $C_j$. Since each entry of the representation $\mathbf{X}^{(e)}$ encodes multi-granular cluster distributions information as a discrete vector, the similarity measure in Eq.~(\ref{eq:similarity}) is reformulated as follows:
\begin{equation}
\label{eq:server_similarity}
s(x^{(e)}_i, C_j) = \left \| \mathbf{u}_j \cdot  \neg (\mathbf{x}^{(e)}_i \oplus \mathbf{c}_j) \right \|_2,
\end{equation}
where $\mathbf{u}_j$ is the $j$-th row of $\mathbf{U}$, containing weights from features to cluster $C_j$. To measure $\mathbf{U}$, both the inter-cluster difference $\alpha_{j\delta}$~\cite{yuan2023spatio,jiang2023anomaly} and the intra-cluster similarity $\beta_{j\delta}$ should be considered simultaneously. $\alpha_{j\delta}$ is quantified using the Hellinger distance\cite{akash2019inter,sengar2008detecting}, defined as:
\begin{equation}
\label{eq:server_alpha}
\alpha_{j\delta}=
\frac{1}{\sqrt[]{2}}
\sqrt[]{\sum_{t=1}^{k_\delta}
\left(\frac{\Upsilon (x_{t\delta}^{(e)}, C_j)}{| C_j |} - 
 \frac{\Upsilon (x_{t\delta}^{(e)}, \bar{C}_j)}{| \bar{C}_j |}\right)^2.}
\end{equation}
$\beta_{j\delta}$ is the average matching rate of each feature within a cluster, which can be expressed as:
\begin{equation}
\label{eq:server_beta}
\beta_{j\delta} = 
\frac{1}{|C_j|} 
\sum_{x_i \in C_j} 
\frac{\Upsilon (x_{i\delta}^{(e)}, C_j)}{|C_j|},
\end{equation}
where $\Upsilon (x_{i\delta}^{(e)}, C_j)$ counts the number of objects in cluster $C_j$ that share the same value $x^{(e)}_{i\delta}$, defined as:
\begin{equation}
\label{eq:count}
\Upsilon (x_{i\delta}^{(e)}, C_j)=\sum_{x_t^{(e)} \in C_j} \neg (x_{t\delta}^{(e)} \oplus x_{i\delta}^{(e)}). 
\end{equation}
The feature-cluster weight $u_{j\delta}$ is calculated as follows:
\begin{equation}
\label{eq:server_feature_cluster_weight}
u_{j\delta} = \frac{\alpha_{j\delta} \beta_{j\delta}}{ {\textstyle \sum_{t=1}^{\Delta} \alpha_{jt} \beta_{jt}}}. 
\end{equation}
REMC encodes multi-granular structures into $\mathbf{X}^{(e)}$, and the final clustering is obtained by jointly optimizing $\mathbf{Q}$ and $\mathbf{U}$.

\begin{algorithm}[!t]
    \caption{\small{GOLD: Global Oriented Local Distribution Learning}}
    \label{alg:GOLD_paper}
	\begin{algorithmic}[1]	
		\REQUIRE Local datasets $\mathbf{X}^{(l)} (l=1,\dots,L)$, learning rate $\eta$.
		\ENSURE Global partition matrix $\mathbf{Q}$.
            \STATE \textit{// Phase 1: FCPL (client)}
            \FORALL{clients}
                \STATE Run CPL with $\mathbf{X}^{(l)}$ to obtain centroids $C^{(l)}$;
                \STATE Upload $C^{(l)}$ to server;
            \ENDFOR
            \STATE \textit{// Phase 2: MCPL (server)}
            \STATE Aggregate all $C^{(l)}$ into centroid matrix $\mathbf{X}$;
            \STATE Initialize $convergence = false$, $Q = \emptyset$, $K = \emptyset$;
            \WHILE{$convergence = false$}
                \STATE Run CPL on $\mathbf{X}$, obtain $\mathbf{Q}_{new}$, $k_{new}$, and $P_{new}$;
                \IF{$k_{old} = k_{new}$ and $|P_{new} - P_{old}| \le \epsilon $}
                    \STATE Set $convergence = true$;
                \ELSE
                    \STATE $Q \leftarrow Q \cup \{\mathbf{Q}_{new}\};$ 
                        $K \leftarrow K \cup \{k_{new}\};$ 
                         $k_{old} \leftarrow k_{new};$
                \ENDIF
            \ENDWHILE
            \STATE \textit{// Phase 3: REMC (server)}
            \STATE Encode $Q$ into $\mathbf{X}^{(e)}$ via Eq.~(\ref{eq:encode});
            \STATE Cluster $\mathbf{X}^{(e)}$ by alternating optimization to convergence;
            \RETURN $\mathbf{Q}$.
    	\end{algorithmic}
\end{algorithm}
%\FloatBarrier

\subsection{Overall Algorithm and Complexity Analysis}
This joint optimization employs an alternating strategy to update the object-cluster assignments $\mathbf{Q}$ and the feature-cluster weights $\mathbf{U}$. By fixing one variable while optimizing the other, each subproblem can be solved independently, facilitating efficient convergence to a stable solution. The process begins with initializing $\mathbf{Q}$ using Eq.(\ref{eq:server_q}), then alternates between the following two steps:
\begin{enumerate}
    \item Fix $\mathbf{Q} = \tilde{\mathbf{Q}}$, update $\tilde{\mathbf{U}}$ by Eqs~(\ref{eq:server_alpha})--(\ref{eq:server_feature_cluster_weight});
    \item Fix $\mathbf{U} = \tilde{\mathbf{U}}$, update $\tilde{\mathbf{Q}}$ by Eq.~(\ref{eq:server_q}).
\end{enumerate}

The learning process of GOLD can be summarized as Algorithm~\ref{alg:GOLD_paper}, which has three key phases: FCPL discovers fine-grained local subclusters at clients, MCPL recursively explores multi-granular global structures at the server, and REMC performs final clustering via alternating optimization. Additional details of the proposed GOLD method are provided in Appendix~\ref{ap:algorithm}.
The convergence is established since FCPL and MCPL inherit convergence from CPL, where each iteration monotonically reduces the objective through: 1) competitive cluster assignment, i.e., assigning objects to minimize weighted distance, and 2) penalty-based weight adjustment, i.e., eliminating redundant clusters. Both operations have been proven to prevent any increase in the objective~\cite{cheung2005maximum}. As for the REMC module, since it follows the optimization of conventional weighted-$k$-means clustering~\cite{huang2005automated}, the convergence is also rigorously guaranteed.

\begin{theorem}
\label{theorem:time_complexity}
    The time complexity of GOLD comprises client-side $\mathcal{O}(Ldn^{(l)}k_0)$ and server-side $\mathcal{O}(dnk_0)$, yielding an overall time complexity of $\mathcal{O}(dk_0(Ln^{(l)}+n))$. 
\end{theorem}

\begin{lemma}
\label{lemma:client_time}
    The aggregate time complexity of the client-side FCPL algorithm is $\mathcal{O}(Ldn^{(l)}k_0)$.
\end{lemma}

\begin{prf}
    To analyze the worst-case time complexity of the $l$-th client, let $k_0$ denote the initial number of candidate clusters and $M$ denote the maximum number of iterations for the competitive penalization learning process. During the computation of object-cluster similarities, $n^{(l)} \times k_0$ distance pairs must be calculated across $d$ features, yielding a time complexity of $\mathcal{O}(M d n^{(l)} k_0)$. Similarly, updating the feature-cluster importance matrix $\mathbf{H}$ requires processing all $n^{(l)}$ data objects, which incurs the same time complexity of $\mathcal{O}(M d n^{(l)} k_0)$. Since $M$ is relatively small compared to other variables, its contribution to overall time complexity is negligible. Considering $L$ clients, the aggregate time complexity across all clients is $\mathcal{O}(L d n^{(l)} k_0)$.
\qed
\end{prf}

\begin{lemma}
\label{lemma:server_time}
    The overall time complexity of the server-side algorithms, comprising MCPL and REMC, is $\mathcal{O}(dnk_0)$.
\end{lemma}

\begin{prf}
    Assume the exploration of multi-granular cluster distributions requires $\Delta$ iterations. In each iteration, the time complexity is $O(d n k_0)$, as detailed in the client-side time complexity analysis. Consequently, the overall time complexity for this process is $\mathcal{O}(\Delta d n k_0)$. For the clustering process in REMC, let $M$ denote the iterations required for convergence with $k^*$ clusters. In each iteration, both the update of feature-cluster weights $\mathbf{U}$ and the partitioning of $n$ data objects involve processing $\Delta$ features across $k^*$ clusters. The total time complexity is therefore $\mathcal{O}(M n \Delta k^*)$. Because $\Delta$, $M$, and $k^*$ are relatively small compared to other variables, their contributions to the overall time complexity are negligible. Thus, based on the previous analysis, the overall server-side time complexity is expressed as $\mathcal{O}(dnk_0)$.
\qed
\end{prf}

\begin{theorem}
\label{theorem:space_complexity}
    The space complexity of GOLD comprises client-side $\mathcal{O}(L n^{(l)} (d + k_0))$ and server-side $\mathcal{O}(n(d+k_0))$, yielding an overall space complexity of GOLD being $\mathcal{O}((d+k_0)(Ln^{(l)}+n))$.
\end{theorem}

\begin{prf}
Please refer to Appendix~\ref{ap:space}.
\qed
\end{prf}

The theoretical analysis of time complexity is verified by the empirical studies in Section~\ref{experiment:scalability}.

\section{Experiment}

To evaluate the proposed Global Oriented Local Distribution (GOLD), it is compared with eight state-of-the-art methods across ten datasets. The experimental settings are first described, followed by results and discussion.

\subsection{Experimental Settings}
\label{sec:settings}

\textbf{Six experiments} are designed as follows:
\begin{itemize}
    \item \textbf{Clustering Performance Comparison:} The proposed GOLD is compared with seven state-of-the-art counterparts to illustrate its superiority. A Wilcoxon signed-rank test is conducted on the performance of the compared counterparts to statistically demonstrate the superiority.

    \item\textbf{Impact of Non-ICD:} The degree of Non-ICD can be reflected by the degree $\lambda$ defined in Section~\ref{sec:problem}. To evaluate its impact on FC performance, the proposed GOLD is compared with the counterparts on Non-ICD datasets with different $\lambda$.
    
    \item \textbf{Ablation Studies:} 1) Component Ablation. To assess the effectiveness of GOLD, its performance is compared against ablated variants with specific components removed. 2) Granularity Ablation. To analyze the role of multi-granular cluster distributions in global aggregation, the contribution of individual granularity levels is evaluated by selectively controlling their participation. 

    \item \textbf{Scalability Evaluation:} To examine the robustness and scalability of GOLD across diverse federated settings, its performance is measured under varying numbers of clients and compared with state-of-the-art counterparts.
    To assess the computational efficiency of GOLD, its execution time is compared with that of SOTA counterparts under different numbers of objects $N$ and dimensions $d$.
    
    \item \textbf{Qualitative Results:} To validate effectiveness of GOLD, visualizations of outputs at different stages are shown.
    
    \item \textbf{Hyper-Parameter Evaluation:} To examine the sensitivity of GOLD to its two key hyperparameters, their effects are analyzed through a joint investigation of different parameter combinations, see Appendix~\ref{ap:parameter}.
\end{itemize}

\textbf{Eight counterparts} are compared, including four iterative and four one-shot FC methods. The iterative methods include FedSC~\cite{qiao2024federated}, AFCL~\cite{zhang2025asynchronous}, FFCM-avg1, and FFCM-avg2~\cite{stallmann2022towards}, while the one-shot methods include $k$-Fed~\cite{dennis2021heterogeneity}, OSFSC-SSC~\cite{xie2023fed}, OSFSC-TSC~\cite{xie2023fed}, and NN-FC~\cite{wang2024one}. For each baseline, the hyperparameter settings recommended in the original papers are followed. For the proposed GOLD method, the learning rate is set at $\eta = 0.05$ based on the Hyper-Parameter Evaluation in Appendix~\ref{ap:parameter} and the recommendation from the source literature of CPL~\cite{cheung2005maximum}. The initial number of clusters for launching the learning of FCPL and MCPL at the client and server is set at $k_0 = 0.5 n^{(l)}$ and $k_0 = 0.5 n$, respectively. For all compared methods, the global number of clusters $k^*$ searched at the server is set according to the benchmark labels of the datasets reported in Table~\ref{tbl:dataset}.

\textbf{Six Validity Indices} are employed, including four external indices, i.e., Purity, Adjusted Rand Index (ARI)~\cite{hubert1985comparing}, Normalized Mutual Information (NMI)~\cite{strehl2002cluster}, and Clustering Accuracy (ACC), and two internal indices, i.e., Silhouette Coefficient (SC)~\cite{Silhouettes1987} and Calinski-Harabasz Index (CH)~\cite{CH1974}. 
ARI effectively discriminates clustering performance by accounting for chance groupings.
NMI evaluates performance from an information-theoretic perspective.
ACC measures the proportion of correctly assigned objects.
SC captures clustering quality by balancing intra-cluster cohesion and inter-cluster separation based on pairwise distance.
CH assesses cluster validity by maximizing the ratio of inter-cluster variance to intra-cluster dispersion, promoting partition compactness.
For all indices, larger values indicate better clustering performance.

\begin{table}[!t]
    \centering
    \caption{Statistics of the 10 datasets. $n$, $d$, $k^*$, and $\lambda$ indicate the number of objects, the number of features, the true number of clusters, and the Non-ICD degree, respectively.}
    \label{tbl:dataset}
    \scalebox{0.95}{\begin{tabular}{c|c c|c c c c}
    \toprule
    \textbf{No.} & \textbf{Data} & \textbf{Abbrev.} & \textbf{$n$} & \textbf{$d$} & \textbf{$k^*$} & \textbf{$\lambda $}\\
    \midrule
    \text{1} & \text{Ecoli} & \textbf{EC} & \text{336} & \text{7} & \text{8} & \text{0.43} \\
    \rowcolor{gray!15}
    \text{2} & \text{User Knowledge Modeling} & \textbf{US} & \text{403} & \text{5} & \text{4} & \text{0.28}\\
    \text{3} & \text{Statlog(Vehicle Silhouettes)} & \textbf{VE} & \text{845} & \text{18} & \text{4} & \text{0.43}\\
    \rowcolor{gray!15}
    \text{4} & \text{HCV for Egyptian Patients} & \textbf{EP} & \text{1385} & \text{28} & \text{4} & \text{0.02}\\
    \text{5} & \text{Yeast} & \textbf{YE} & \text{1484} & \text{8} & \text{10} & \text{0.30}\\
    \rowcolor{gray!15}
    \text{6} & \text{Cardiotocography} & \textbf{CA} & \text{2126} & \text{21} & \text{10} & \text{0.53}\\
    \text{7} & \text{Statlog(Landsat Satellite)} & \textbf{LA} & \text{6435} & \text{36} & \text{6} & \text{0.10}\\
    \rowcolor{gray!15}
    \text{8} & \text{Wine Quality} & \textbf{WI} & \text{6497} & \text{11} & \text{7} & \text{0.27}\\
    \text{9} & \text{Pen-Based Digits} & \textbf{PE} & \text{10992} & \text{16} & \text{10} & \text{0.57}\\
    \rowcolor{gray!15}
    \text{10} & \text{Letter Recognition} & \textbf{LE} & \text{20000} & \text{16} & \text{26} & \text{0.52}\\
    \bottomrule
    \end{tabular}}
\end{table}

\textbf{Ten Data Sets} are employed for comprehensive evaluation, obtained from the UCI Machine Learning Repository\footnote{https://archive.ics.uci.edu/}.  
Detailed statistics are shown in Table~\ref{tbl:dataset}. These datasets provide diverse evaluation scenarios with sample sizes ranging from 336 to 20,000, feature dimensions from 5 to 36, and number of clusters from 4 to 26, spanning biological (Ecoli, Yeast), medical (HCV, Cardiotocography), and recognition (Pen-Based Digits, Letter Recognition) domains. In the experimental simulations, the number of clients is set to $L=8$ by default.

\begin{table}[!t]
    \centering
    \caption{Comparison of clustering algorithms for Non-ICD simulation. A \colorbox{green!15}{green cell} indicates that the algorithm satisfies the corresponding Non-ICD simulation requirement.}
    \label{tab:algorithm_comparison}
    \scalebox{0.87}{
    \begin{tabular}{>{\centering\arraybackslash}m{2.4cm}|>{\centering\arraybackslash}m{1.8cm}>{\centering\arraybackslash}m{2.55cm}>{\centering\arraybackslash}m{1.7cm}}
    \toprule
    \textbf{Algorithm} & \textbf{Granularity Controllability} & \textbf{Partitioning Randomness} & \textbf{Assignment Completeness} \\
    \midrule
    Conventional partitional (e.g., $k$-means) 
    & \cellcolor{green!15}Directly specified via $k$ 
    & \cellcolor{green!15}Fully random; Realistic simulation 
    & \cellcolor{green!15}All points assigned to subclusters \\
    \arrayrulecolor{white}\specialrule{4pt}{0pt}{0pt}\arrayrulecolor{black}
    Initialization-enhanced partitional (e.g., $k$-means++) 
    & \cellcolor{green!15}Directly specified via $k$ 
    & Distance-weighted; Stability over randomness 
    & \cellcolor{green!15}All points assigned to subclusters \\
    \arrayrulecolor{white}\specialrule{4pt}{0pt}{0pt}\arrayrulecolor{black}
    Density-based (e.g., DBSCAN) 
    & Density and hyperparameter-dependent 
    & Natural cluster shape-dependent; Stable 
    & Some points identified as noise \\
    \bottomrule
    \end{tabular}}
\end{table}

To simulate Non-ICD, each benchmark cluster is divided into non-overlapping subclusters at different granularities. Available partitioning methods include conventional partitional clustering (e.g., $k$-means), initialization-enhanced partitional clustering for more stable results (e.g., $k$-means++), and density-based clustering for arbitrary cluster shapes (e.g., DBSCAN). To faithfully simulate Non-ICD, the algorithm should have granularity controllability, partitioning randomness, and assignment completeness, rather than superior clustering performance. As summarized in Table~\ref{tab:algorithm_comparison}, conventional $k$-means is a good choice, as it directly controls the number of subclusters via $k$, and provides random partitioning for realistic Non-ICD distribution simulation. The obtained subclusters are randomly allocated to different clients to simulate Non-ICD, and the workflow is summarized as follows:

\textbf{Step 1: Initialization.} 
Given a global dataset $\mathbf{X}$ with cluster labels $\{1, 2, \dots, k\}$, each client is randomly assigned a subset of global clusters, where assigned clusters $k_{(l)} \in [1, k]$.

\textbf{Step 2: Cluster Selection for Each Client.} 
For the $l$-th client, randomly select $k_{(l)}$ clusters from the global dataset.

\textbf{Step 3: Local Sub-Partitioning.} 
For each selected global cluster assigned to the $l$-th client, denote the number of data points in that cluster as $n_{k}$. Randomly determine the number of subclusters $k_{\text{sub}}$, and apply $k$-means clustering with $k_{\text{sub}}$ to partition the cluster into smaller subclusters. The subclusters are then recorded for subsequent selection.

\textbf{Step 4: Sub-Cluster Selection.} 
For each cluster allocated to the $l$-th client, randomly select $N_{\text{select}}$ subclusters, where $N_{\text{select}} \in [1, k_{\text{sub}}]$. The data points belonging to the selected subclusters are aggregated to form the client's candidate pool.

\textbf{Step 5: Data Point Sampling.} 
From the candidate pool, randomly sample $N_l$ data points, where $N_l \in [0.25N, 0.75N]$ and $N$ denotes total data objects in the pool. The resulting subset $\mathbf{X}^{(l)}$ constitutes the dataset assigned to the $l$-th client.

The above procedure is executed independently for each client to appropriately simulate Non-ICD in real scenarios.

\begin{table*}[!t]
    \centering
    \caption{Global clustering performance evaluated by Purity and ARI. ``$\overline{AR}(ST)$'' row reports the average performance ranks and the significance test results, where a significant difference between the corresponding method and GOLD is indicated by ``+''.}
    \label{tbl:centralized_performance_paper}
    \scalebox{0.92}{\begin{tabular}{c|c|c c c c c c c c|c}
    \toprule
    \multirow{2}{*}{\textbf{Index}} & \multirow{2}{*}{\textbf{Data}} 
    & \textbf{$k$-Fed} & \textbf{FFCM-avg1} & \textbf{FFCM-avg2} & \textbf{OSFSC-SSC} & \textbf{OSFSC-TSC} & \textbf{FedSC} & \textbf{NN-FC} & \textbf{AFCL} & \textbf{GOLD} \\
    & & \textbf{(2021 \cite{dennis2021heterogeneity})} & \textbf{(2022 \cite{stallmann2022towards})} & \textbf{(2022 \cite{stallmann2022towards})} & \textbf{(2023 \cite{xie2023fed})} & \textbf{(2023 \cite{xie2023fed})} & \textbf{(2024 \cite{qiao2024federated})} & \textbf{(2024 \cite{wang2024one})} & \textbf{(2025 \cite{zhang2025asynchronous})} & \textbf{(ours)}\\ 
    \midrule
    \multirow{10}{*}{Purity} 
     & EC & 0.767±0.03 & 0.661±0.04 & 0.691±0.01 & 0.802±0.02 & \cellcolor[RGB]{255,243,218}0.810±0.01 & 0.708±0.09 & 0.735±0.03 & 0.471±0.02 & \cellcolor[RGB]{255,228,173}0.816±0.03 \\ 
     & US & 0.451±0.02 & \cellcolor[RGB]{255,243,218}0.477±0.02 & 0.467±0.01 & \cellcolor[RGB]{255,243,218}0.477±0.03 & 0.454±0.00 & 0.441±0.02 & 0.466±0.04 & 0.320±0.00 &  \cellcolor[RGB]{255,228,173}0.495±0.05 \\
     & VE & 0.371±0.01 & 0.379±0.02 & 0.382±0.01 & \cellcolor[RGB]{255,243,218}0.384±0.01 & 0.370±0.00 & 0.369±0.00 & 0.380±0.02 & 0.273±0.00 &  \cellcolor[RGB]{255,228,173}0.387±0.02 \\
     & EP & 0.274±0.00 & \cellcolor[RGB]{255,243,218}0.276±0.00 & 0.273±0.00 & \cellcolor[RGB]{255,243,218}0.276±0.00 & 0.270±0.01 & 0.268±0.00 &  \cellcolor[RGB]{255,228,173}0.277±0.00 & 0.261±0.00 &  \cellcolor[RGB]{255,228,173}0.277±0.01 \\
     & YE & 0.480±0.04 & 0.446±0.02 & 0.398±0.03 & \cellcolor[RGB]{255,243,218}0.493±0.01 & 0.486±0.01 & 0.426±0.05 & 0.422±0.03 & 0.313±0.00 &  \cellcolor[RGB]{255,228,173}0.498±0.04 \\
     & CA & 0.429±0.02 & 0.368±0.01 & 0.327±0.00 & 0.424±0.02 & \cellcolor[RGB]{255,243,218}0.430±0.01 & 0.345±0.02 & 0.364±0.04 & 0.272±0.00 &  \cellcolor[RGB]{255,228,173}0.436±0.03 \\
     & LA & 0.617±0.05 & 0.632±0.00 & 0.537±0.02 & 0.643±0.00 & 0.640±0.00 & 0.615±0.03 & \cellcolor[RGB]{255,243,218}0.657±0.05 & 0.323±0.02 &  \cellcolor[RGB]{255,228,173}0.658±0.07 \\
     & WI & 0.452±0.00 & 0.461±0.00 & 0.449±0.00 & 0.451±0.01 & \cellcolor[RGB]{255,243,218}0.463±0.00 & 0.439±0.00 & 0.446±0.01 & 0.437±0.00 &  \cellcolor[RGB]{255,228,173}0.463±0.01 \\
     & PE & 0.583±0.06 & 0.431±0.00 & 0.421±0.03 & 0.595±0.02 & 0.594±0.02 & 0.509±0.04 &  \cellcolor[RGB]{255,228,173}0.619±0.04 & 0.104±0.00 & \cellcolor[RGB]{255,243,218}0.608±0.06 \\
     & LE & 0.272±0.01 & \cellcolor[RGB]{255,243,218}0.191±0.01 & 0.064±0.00 & \cellcolor[RGB]{255,243,218}0.279±0.01 & 0.277±0.01 & 0.191±0.01 & 0.275±0.01 & 0.041±0.00 &  \cellcolor[RGB]{255,228,173}0.281±0.01 \\ 
    \midrule
    \multicolumn{2}{c|}{$\overline{AR}(ST)$}
    & 4.9(+) & 4.9(+) & 6.6(+) & 3.1(+) & 3.9(+) & 7.1(+) & 4.2(+) & 9.0(+) & 1.1 \\
    \midrule
    \multirow{10}{*}{ARI} 
     & EC & 0.632±0.05 & 0.387±0.08 & 0.435±0.05 & 0.603±0.06 &  \cellcolor[RGB]{255,228,173}0.662±0.02 & 0.451±0.16 & 0.575±0.08 & 0.044±0.02 & \cellcolor[RGB]{255,243,218}0.645±0.10 \\
     & US & 0.110±0.02 & \cellcolor[RGB]{255,243,218}0.135±0.01 & 0.126±0.01 & 0.123±0.03 & 0.116±0.00 & 0.124±0.01 & 0.103±0.03 & 0.000±0.00 &  \cellcolor[RGB]{255,228,173}0.160±0.05 \\
     & VE & \cellcolor[RGB]{255,243,218}0.078±0.01 & 0.073±0.02 & 0.074±0.00 & 0.077±0.01 & 0.077±0.00 & 0.076±0.00 & 0.075±0.01 & 0.010±0.00 &  \cellcolor[RGB]{255,228,173}0.079±0.01 \\
     & EP & \cellcolor[RGB]{255,243,218}0.001±0.00 & \cellcolor[RGB]{255,243,218}0.001±0.00 & 0.000±0.00 & 0.000±0.00 & 0.000±0.00 & 0.000±0.00 & \cellcolor[RGB]{255,243,218}0.001±0.00 & 0.000±0.00 &  \cellcolor[RGB]{255,228,173}0.002±0.02 \\
     & YE & 0.151±0.05 & 0.129±0.01 & 0.122±0.02 &  \cellcolor[RGB]{255,228,173}0.162±0.01 & 0.143±0.00 & 0.075±0.06 & 0.120±0.05 & 0.000±0.00 & \cellcolor[RGB]{255,243,218}0.160±0.04 \\
     & CA & 0.152±0.02 & 0.130±0.00 & 0.122±0.00 & \cellcolor[RGB]{255,243,218}0.154±0.01 & \cellcolor[RGB]{255,243,218}0.154±0.01 & 0.109±0.03 & 0.115±0.05 & 0.000±0.00 &  \cellcolor[RGB]{255,228,173}0.160±0.02 \\
     & LA & 0.373±0.04 & 0.432±0.00 & 0.300±0.01 & \cellcolor[RGB]{255,243,218}0.435±0.00 & 0.421±0.00 & 0.351±0.02 & 0.413±0.05 & 0.091±0.02 &  \cellcolor[RGB]{255,228,173}0.439±0.08 \\
     & WI & 0.023±0.01 & 0.021±0.00 & 0.018±0.00 & \cellcolor[RGB]{255,243,218}0.024±0.01 & \cellcolor[RGB]{255,243,218}0.024±0.01 & -0.005±0.01 & 0.005±0.01 & 0.000±0.00 &  \cellcolor[RGB]{255,228,173}0.029±0.01 \\
     & PE & 0.433±0.06 & 0.212±0.00 & 0.300±0.04 & \cellcolor[RGB]{255,243,218}0.439±0.02 & 0.432±0.02 & 0.382±0.04 & 0.420±0.04 & 0.000±0.00 &  \cellcolor[RGB]{255,228,173}0.441±0.05 \\
     & LE & 0.110±0.01 & 0.064±0.00 & 0.003±0.00 &  \cellcolor[RGB]{255,228,173}0.132±0.00 & 0.126±0.00 & 0.075±0.01 & 0.111±0.01 & 0.000±0.00 & \cellcolor[RGB]{255,243,218}0.129±0.01 \\  
    \midrule
    \multicolumn{2}{c|}{$\overline{AR}(ST)$}
    & 4.0(+) & 5.3(+) & 6.3(+) & 3.0(+) & 4.1(+) & 6.5(+) & 5.8(+) & 8.7(+) & 1.3 \\
    \bottomrule
    \end{tabular}}
\end{table*}

\begin{table*}[!t]
    \centering
    \caption{Federated clustering performance evaluated by Purity and ARI. ``$\overline{AR}(ST)$'' row reports the average performance ranks and the significance test results, where a significant difference between the corresponding method and GOLD is indicated by ``+''.}
    \label{tbl:federated_performance_paper}
    \scalebox{0.92}{
    \begin{tabular}{c|c|c c c c c c c c|c}
    \toprule
    \multirow{2}{*}{\textbf{Index}} & \multirow{2}{*}{\textbf{Data}} 
    & \textbf{$k$-Fed} & \textbf{FFCM-avg1} & \textbf{FFCM-avg2} 
    & \textbf{OSFSC-SSC} & \textbf{OSFSC-TSC} 
    & \textbf{FedSC} & \textbf{NN-FC} & \textbf{AFCL} & \textbf{GOLD} \\ 
    \textbf{} & \textbf{} 
    & \textbf{(2021 \cite{dennis2021heterogeneity})}
    & \textbf{(2022 \cite{stallmann2022towards})}
    & \textbf{(2022 \cite{stallmann2022towards})}
    & \textbf{(2023 \cite{xie2023fed})}
    & \textbf{(2023 \cite{xie2023fed})}
    & \textbf{(2024 \cite{qiao2024federated})}
    & \textbf{(2024 \cite{wang2024one})}
    & \textbf{(2025 \cite{zhang2025asynchronous})}
    & \textbf{(ours)}\\ 
    \midrule
    \multirow{10}{*}{Purity} 
    & EC & \cellcolor[RGB]{255,243,218}0.687±0.01 & 0.417±0.05 & 0.374±0.02 & 0.538±0.03 & 0.593±0.01 & 0.345±0.00 & 0.650±0.06 & 0.348±0.01 &  \cellcolor[RGB]{255,228,173}0.696±0.05 \\
    & US & 0.470±0.01 & 0.465±0.00 & 0.459±0.01 & 0.409±0.02 & 0.451±0.00 & 0.343±0.00 & \cellcolor[RGB]{255,243,218}0.473±0.04 & 0.337±0.00 &  \cellcolor[RGB]{255,228,173}0.529±0.05 \\
    & VE & 0.380±0.02 & 0.342±0.00 & 0.343±0.00 & 0.362±0.01 & \cellcolor[RGB]{255,243,218}0.405±0.00 & 0.347±0.01 & 0.368±0.02 & 0.298±0.00 &  \cellcolor[RGB]{255,228,173}0.406±0.02 \\
    & EP & 0.345±0.01 & 0.341±0.00 & 0.331±0.00 &  \cellcolor[RGB]{255,228,173}0.348±0.01 & 0.343±0.00 & 0.342±0.00 & 0.334±0.00 & 0.341±0.00 & \cellcolor[RGB]{255,243,218}0.346±0.00 \\
    & YE & 0.447±0.05 & 0.294±0.00 & 0.289±0.00 & 0.406±0.02 & \cellcolor[RGB]{255,243,218}0.480±0.01 & 0.283±0.00 & 0.358±0.04 & 0.281±0.01 &  \cellcolor[RGB]{255,228,173}0.493±0.06 \\
    & CA & 0.515±0.03 & 0.375±0.00 & 0.351±0.00 & 0.467±0.02 & \cellcolor[RGB]{255,243,218}0.527±0.01 & 0.441±0.05 & 0.418±0.02 & 0.289±0.00 &  \cellcolor[RGB]{255,228,173}0.542±0.04 \\
    & LA & 0.610±0.01 & 0.613±0.00 & 0.475±0.00 & 0.612±0.00 & \cellcolor[RGB]{255,243,218}0.713±0.00 & 0.622±0.01 & 0.669±0.05 & 0.283±0.11 &  \cellcolor[RGB]{255,228,173}0.752±0.01 \\
    & WI & 0.492±0.02 & \cellcolor[RGB]{255,243,218}0.520±0.00 & 0.464±0.00 & 0.485±0.01 &  \cellcolor[RGB]{255,228,173}0.525±0.00 & 0.467±0.01 & 0.467±0.00 & 0.464±0.00 & 0.514±0.02 \\
    & PE & 0.632±0.00 & 0.335±0.00 & 0.311±0.00 &  \cellcolor[RGB]{255,228,173}0.664±0.02 & 0.653±0.02 & 0.519±0.04 & 0.655±0.03 & 0.162±0.00 & \cellcolor[RGB]{255,243,218}0.662±0.06 \\
    & LE & 0.335±0.01 & 0.143±0.00 & 0.100±0.00 & 0.329±0.01 & 0.332±0.01 & 0.241±0.02 & \cellcolor[RGB]{255,243,218}0.338±0.02 & 0.089±0.00 &  \cellcolor[RGB]{255,228,173}0.346±0.02 \\
    \midrule
    \multicolumn{2}{c|}{$\overline{AR}(ST)$}
    & 3.6(+) & 5.8(+) & 7.5(+) & 4.3(+) & 3.1(+) & 6.3(+) & 4.3(+) & 8.7(+) & 1.4 \\
    \midrule
    \multirow{10}{*}{ARI} 
    & EC & \cellcolor[RGB]{255,243,218}0.420±0.06 & 0.060±0.04 & 0.048±0.01 & 0.163±0.04 & 0.225±0.01 & -0.001±0.01 & 0.415±0.12 & 0.013±0.01 &  \cellcolor[RGB]{255,228,173}0.452±0.09 \\
    & US & \cellcolor[RGB]{255,243,218}0.155±0.00 & 0.127±0.00 & 0.125±0.01 & 0.045±0.01 & 0.094±0.00 & 0.001±0.00 & 0.128±0.05 & 0.000±0.00 &  \cellcolor[RGB]{255,228,173}0.227±0.12 \\
    & VE & 0.079±0.00 & 0.048±0.00 & 0.049±0.00 & 0.037±0.00 & \cellcolor[RGB]{255,243,218}0.082±0.00 & 0.061±0.00 & 0.056±0.02 & 0.000±0.00 &  \cellcolor[RGB]{255,228,173}0.088±0.02 \\
    & EP & \cellcolor[RGB]{255,243,218}0.015±0.01 & 0.010±0.01 &  \cellcolor[RGB]{255,228,173}0.016±0.00 & 0.014±0.01 & 0.007±0.00 & 0.003±0.00 & 0.002±0.01 & 0.000±0.00 & 0.012±0.00 \\
    & YE & \cellcolor[RGB]{255,243,218}0.198±0.06 & 0.045±0.00 & 0.044±0.00 & 0.088±0.01 & 0.151±0.01 & 0.003±0.00 & 0.097±0.04 & 0.003±0.01 &  \cellcolor[RGB]{255,228,173}0.217±0.06 \\
    & CA & 0.193±0.02 & 0.064±0.00 & 0.061±0.00 & 0.107±0.02 & \cellcolor[RGB]{255,243,218}0.211±0.01 & 0.185±0.05 & 0.122±0.03 & 0.000±0.00 &  \cellcolor[RGB]{255,228,173}0.251±0.04 \\
    & LA & 0.383±0.02 & 0.372±0.00 & 0.219±0.00 & 0.315±0.01 & \cellcolor[RGB]{255,243,218}0.531±0.00 & 0.483±0.03 & 0.452±0.05 & 0.041±0.13 &  \cellcolor[RGB]{255,228,173}0.539±0.05 \\
    & WI & 0.031±0.03 & \cellcolor[RGB]{255,243,218}0.038±0.00 & -0.010±0.01 & 0.022±0.02 &  \cellcolor[RGB]{255,228,173}0.048±0.00 & 0.003±0.01 & -0.010±0.01 & 0.000±0.00 & \cellcolor[RGB]{255,228,173}0.048±0.03 \\
    & PE & 0.457±0.00 & 0.137±0.00 & 0.153±0.01 & 0.426±0.04 & 0.475±0.02 & 0.368±0.04 & \cellcolor[RGB]{255,243,218}0.511±0.03 & 0.000±0.00 &  \cellcolor[RGB]{255,228,173}0.516±0.07 \\
    & LE & 0.160±0.01 & 0.037±0.00 & 0.005±0.00 & 0.080±0.02 & \cellcolor[RGB]{255,243,218}0.167±0.01 & 0.129±0.01 & 0.144±0.01 & 0.000±0.00 &  \cellcolor[RGB]{255,228,173}0.174±0.02 \\
    \midrule
    \multicolumn{2}{c|}{$\overline{AR}(ST)$}
    & 3.0(+) & 5.9(+) & 6.5(+) & 5.7(+) & 3.1(+) & 6.0(+) & 4.7(+) & 8.7(+) & 1.3 \\
    \bottomrule
    \end{tabular}}
\end{table*}

\begin{figure}[!t]
    \centering
    \includegraphics[width=0.6\linewidth]{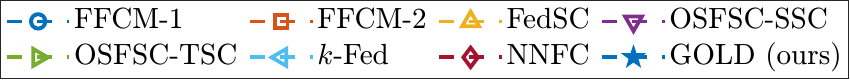} \\[1em]
    \includegraphics[width=0.49\linewidth]{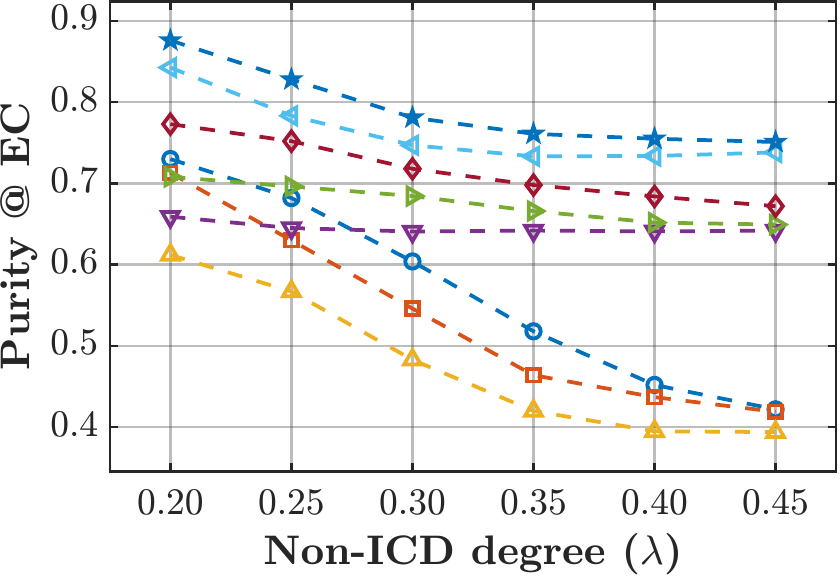}
    \includegraphics[width=0.49\linewidth]{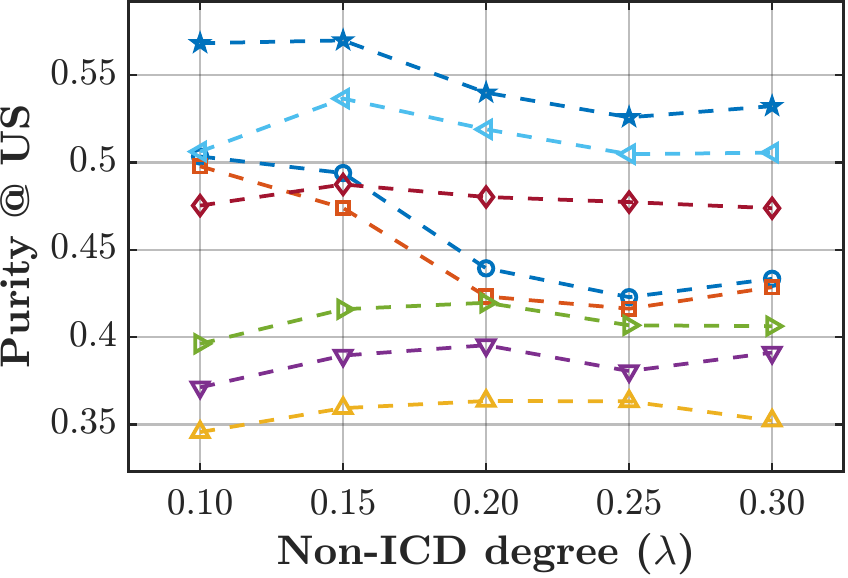}
    \includegraphics[width=0.49\linewidth]{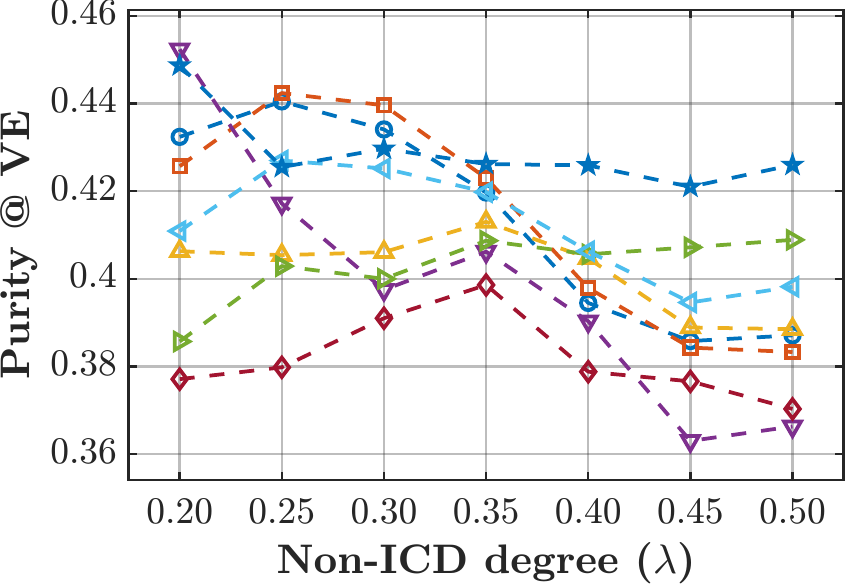}
    \includegraphics[width=0.49\linewidth]{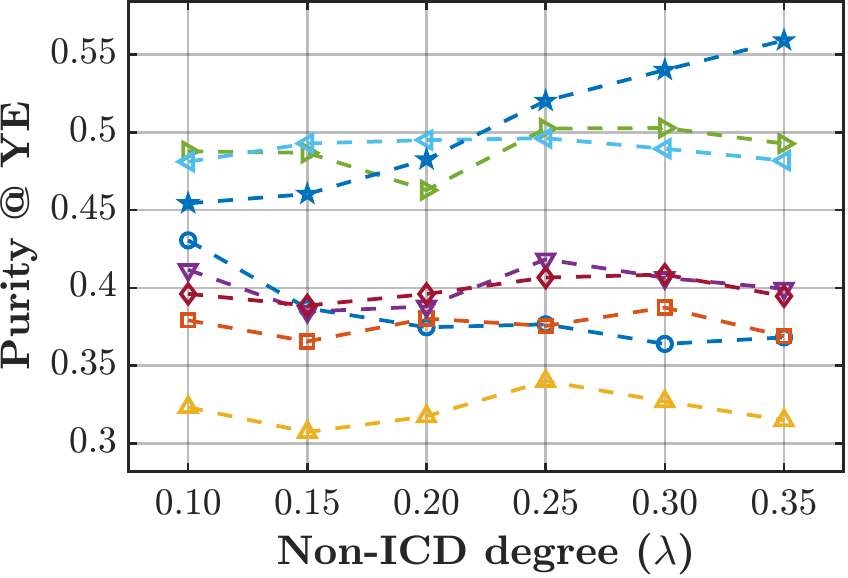}
    \caption{Federated clustering performance comparison under different Non-ICD degrees ($\lambda$). A smaller $\lambda$ indicates lower data heterogeneity across clients.}
    \label{fig:non-icd-papers}
\end{figure}

\begin{table}[h]
    \centering
    \caption{Ablation study of FCPL and MCPL components. \colorbox[RGB]{232,243,252}{Light blue}: Improvement over baseline; \colorbox[RGB]{185,219,247}{Dark blue}: Best performance.}
    \label{tbl:ablation_of_components}
    \scalebox{0.95}{
    \begin{tabular}{c|c|c|c c c c}
    \toprule
    \textbf{Data} & \textbf{FCPL} & \textbf{MCPL} & \textbf{Purity} & \textbf{ARI} & \textbf{NMI} & \textbf{ACC}\\
    \midrule
    \multirow{4}{*}{EC}
    & \XSolidBrush & \XSolidBrush & 0.7248 & 0.5381 & 0.6381 & 0.7125\\
    & \Checkmark & \XSolidBrush & \cellcolor[RGB]{232,243,252}0.7462 & \cellcolor[RGB]{232,243,252}0.6438 & 0.6244 & \cellcolor[RGB]{232,243,252}0.7339\\
    & \XSolidBrush & \Checkmark & \cellcolor[RGB]{232,243,252}0.7615 & \cellcolor[RGB]{232,243,252}0.5696 & \cellcolor[RGB]{232,243,252}0.6440 & 0.6881\\
    & \Checkmark & \Checkmark & \cellcolor[RGB]{185,219,247}0.7798 & \cellcolor[RGB]{185,219,247}0.6881 & \cellcolor[RGB]{185,219,247}0.6475 & \cellcolor[RGB]{185,219,247}0.7431\\
    \midrule
    \multirow{4}{*}{US}
    & \XSolidBrush & \XSolidBrush & 0.4702 & 0.0400 & 0.0796 & 0.3436\\
    & \Checkmark & \XSolidBrush & 0.3960 & \cellcolor[RGB]{232,243,252}0.0449 & \cellcolor[RGB]{232,243,252}0.0985 & \cellcolor[RGB]{232,243,252}0.3743\\
    & \XSolidBrush & \Checkmark & 0.4412 & \cellcolor[RGB]{232,243,252}0.0914 & \cellcolor[RGB]{232,243,252}0.1243 & \cellcolor[RGB]{232,243,252}0.4376\\
    & \Checkmark & \Checkmark & \cellcolor[RGB]{185,219,247}0.4882 & \cellcolor[RGB]{185,219,247}0.1033 & \cellcolor[RGB]{185,219,247}0.1965 & \cellcolor[RGB]{185,219,247}0.4882\\
    \midrule
    \multirow{4}{*}{VE}
    & \XSolidBrush & \XSolidBrush & 0.4585 & 0.0881 & 0.1203 & 0.3990\\
    & \Checkmark & \XSolidBrush & 0.4531 & \cellcolor[RGB]{232,243,252}0.0882 & \cellcolor[RGB]{232,243,252}0.1220 & \cellcolor[RGB]{232,243,252}0.4117\\
    & \XSolidBrush & \Checkmark & 0.4533 & \cellcolor[RGB]{232,243,252}0.0945 & \cellcolor[RGB]{232,243,252}0.1259 & \cellcolor[RGB]{232,243,252}0.4031\\
    & \Checkmark & \Checkmark & \cellcolor[RGB]{185,219,247}0.4738 & \cellcolor[RGB]{185,219,247}0.0958 & \cellcolor[RGB]{185,219,247}0.1336 & \cellcolor[RGB]{185,219,247}0.4262\\
    \midrule
    \multirow{4}{*}{EP}
    & \XSolidBrush & \XSolidBrush & 0.3038 & 0.0056 & 0.0125 & 0.3038\\
    & \Checkmark & \XSolidBrush & \cellcolor[RGB]{232,243,252}0.3149 & \cellcolor[RGB]{232,243,252}0.0166 & \cellcolor[RGB]{232,243,252}0.0176 & \cellcolor[RGB]{232,243,252}0.3093\\
    & \XSolidBrush & \Checkmark & \cellcolor[RGB]{232,243,252}0.3204 & \cellcolor[RGB]{232,243,252}0.0091 & \cellcolor[RGB]{232,243,252}0.0196 & 0.3022\\
    & \Checkmark & \Checkmark & \cellcolor[RGB]{185,219,247}0.3244 & \cellcolor[RGB]{185,219,247}0.0174 & \cellcolor[RGB]{185,219,247}0.0248 & \cellcolor[RGB]{185,219,247}0.3180\\
    \midrule
    \multirow{4}{*}{YE}
    & \XSolidBrush & \XSolidBrush & 0.4192 & 0.1541 & 0.3159 & 0.3937\\
    & \Checkmark & \XSolidBrush & \cellcolor[RGB]{232,243,252}0.4658 & \cellcolor[RGB]{232,243,252}0.2040 & \cellcolor[RGB]{232,243,252}0.3467 & \cellcolor[RGB]{232,243,252}0.4541\\
    & \XSolidBrush & \Checkmark & \cellcolor[RGB]{232,243,252}0.5189 & \cellcolor[RGB]{232,243,252}0.2164 & \cellcolor[RGB]{232,243,252}0.3374 & \cellcolor[RGB]{232,243,252}0.4367\\
    & \Checkmark & \Checkmark & \cellcolor[RGB]{185,219,247}0.5266 & \cellcolor[RGB]{185,219,247}0.2449 & \cellcolor[RGB]{185,219,247}0.3661 & \cellcolor[RGB]{185,219,247}0.4753\\
    \midrule
    \multirow{4}{*}{CA}
    & \XSolidBrush & \XSolidBrush & 0.5064 & 0.1934 & 0.3588 & 0.4047\\
    & \Checkmark & \XSolidBrush & \cellcolor[RGB]{232,243,252}0.5254 & \cellcolor[RGB]{232,243,252}0.2168 & \cellcolor[RGB]{232,243,252}0.3831 & \cellcolor[RGB]{232,243,252}0.4286\\
    & \XSolidBrush & \Checkmark & \cellcolor[RGB]{232,243,252}0.5647 & \cellcolor[RGB]{232,243,252}0.2313 & \cellcolor[RGB]{232,243,252}0.4065 & \cellcolor[RGB]{232,243,252}0.4132\\
    & \Checkmark & \Checkmark & \cellcolor[RGB]{185,219,247}0.5996 & \cellcolor[RGB]{185,219,247}0.2798 & \cellcolor[RGB]{185,219,247}0.4366 & \cellcolor[RGB]{185,219,247}0.4629\\
    \bottomrule
    \end{tabular}}
\end{table}

\begin{figure}[!t]
    \centering
    \includegraphics[width=0.89\linewidth]{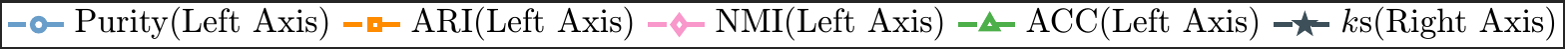} \\[1em]
    \includegraphics[width=0.47\linewidth]{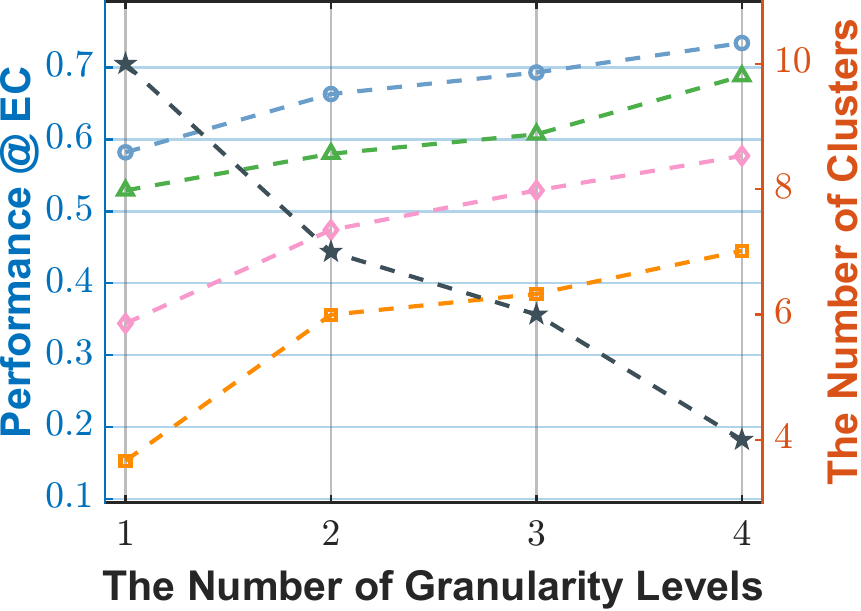}
    \includegraphics[width=0.47\linewidth]{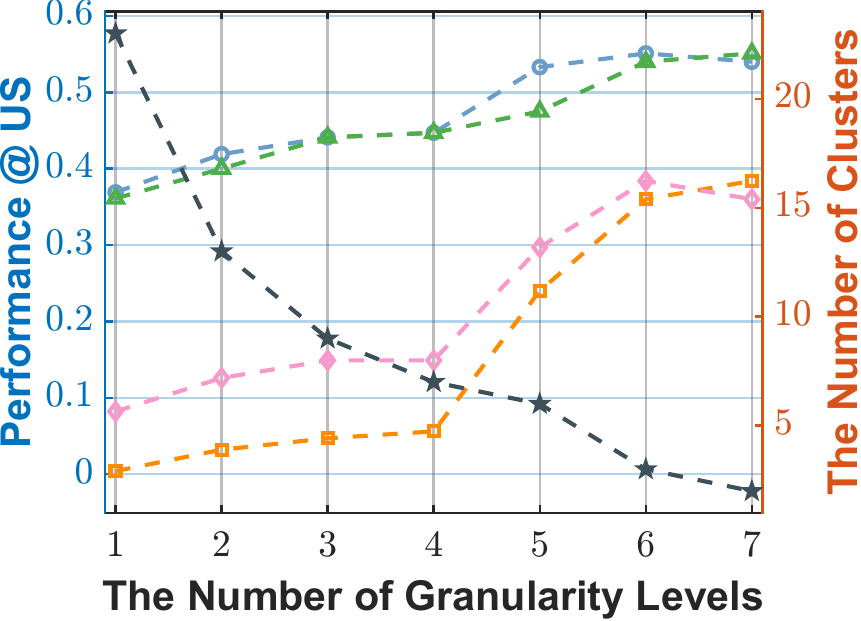}
    \includegraphics[width=0.47\linewidth]{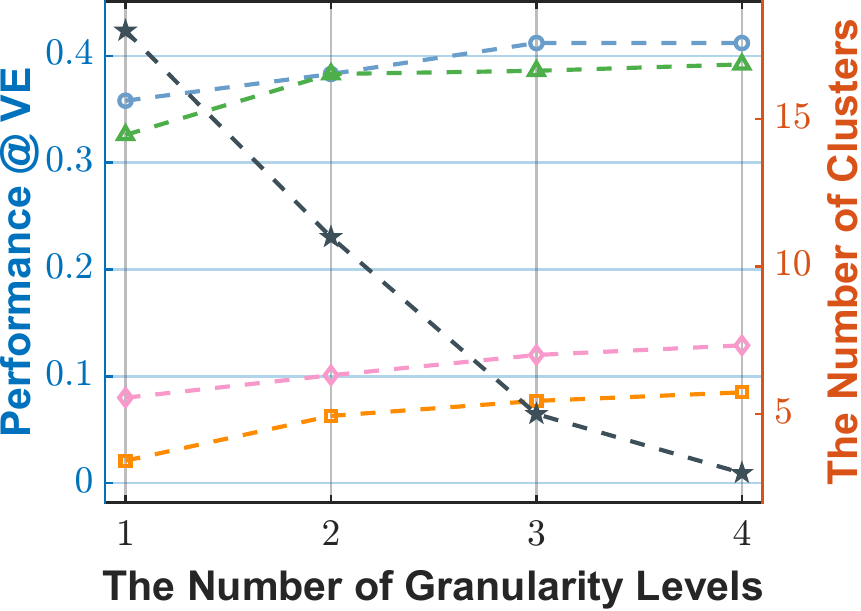}
    \includegraphics[width=0.47\linewidth]{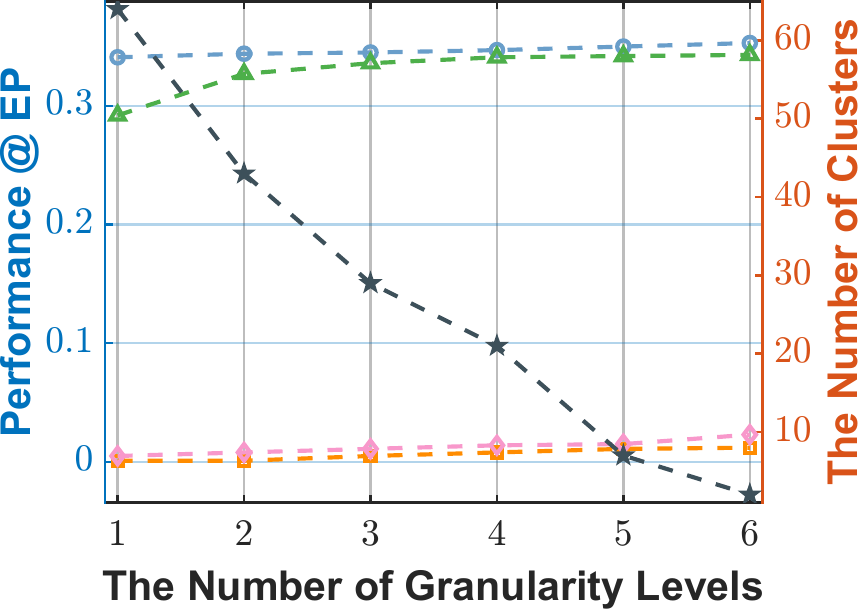}
    \includegraphics[width=0.47\linewidth]{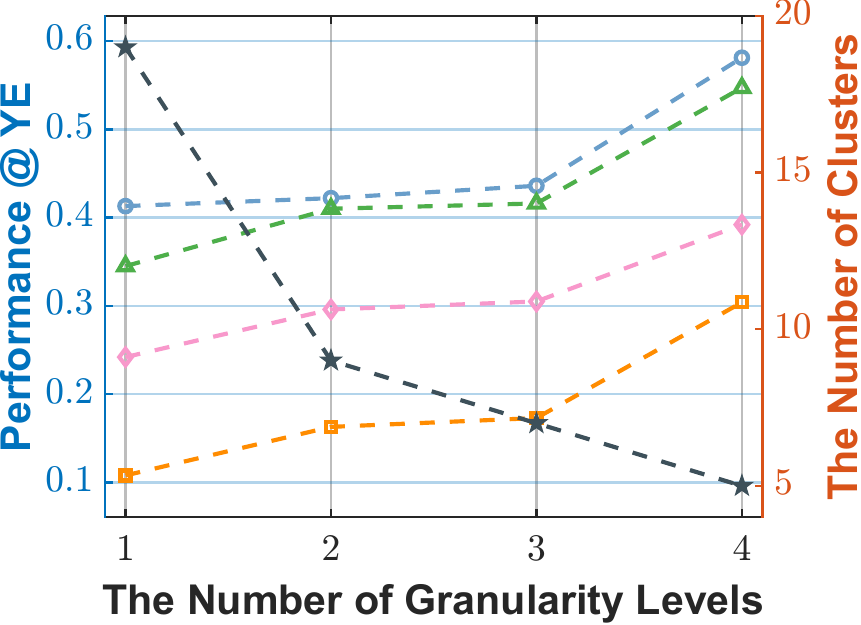}
    \includegraphics[width=0.47\linewidth]{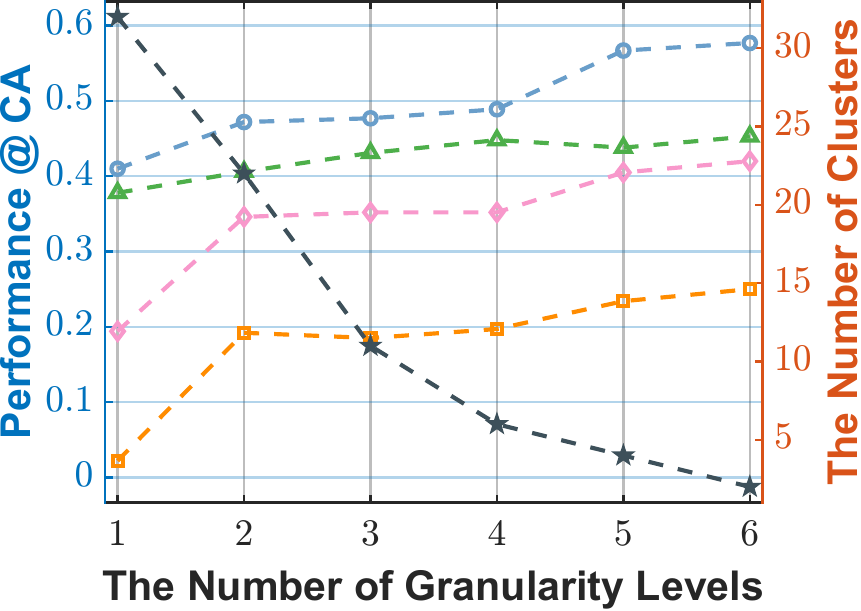}
    \caption{Accumulated granularity ablation of GOLD with progressively incorporated granularity levels. The left axis indicates the model performance at each accumulation stage, while the right axis shows the number of converged clusters corresponding to each granularity.}
    \label{fig:accumulated granularity ablation}
\end{figure}

\subsection{Clustering Performance Comparison}
\label{sec:clustering performance}
To experimentally evaluate the performance of the proposed GOLD, it is compared with state-of-the-art counterparts on ten real-world datasets. 
The compared methods are executed 10 times on each dataset, and the average results are reported. The best and second-best results are indicated in \colorbox[RGB]{255,228,173}{dark orange} and \colorbox[RGB]{255,243,218}{light orange}, respectively. The ``$\overline{AR}(ST)$'' row summarizes the average rank of each method across all datasets, and the symbol ``(+)'' denotes a significant difference based on the Wilcoxon signed-rank test at the 95\% confidence level.

To evaluate the interpretability and representation quality of the clusters learned by FC, the experiment is designed to inherit the output centroids learned by FC methods and launch $k$-means on the complete raw datasets. 
To ensure that the performance is closely tied to the capabilities of the compared FC methods, $k$-means is restricted to a single iteration, serving solely to assign data objects to their nearest centroid. Such a design enables a direct evaluation of how effectively the learned prototypes capture the underlying global structure.
Table~\ref{tbl:centralized_performance_paper} reports the Purity and ARI performance, while the results evaluated by the other four metrics, i.e., NMI, ACC, SC, and CH, are provided in Appendix~\ref{ap:performance}. According to the results, some key observations are provided below: 1) GOLD consistently attains the highest performance, demonstrating superior representation learning capability for exploring the global cluster distribution in FC tasks. GOLD ranks first on the majority of datasets, while it achieves the second-best results on the PE dataset for Purity and on the EC, YE, and LE datasets for ARI. However, the gap between it and the winner is marginal, still demonstrating its robustness across diverse datasets.
2) The comparative methods have only demonstrated satisfactory performance on specific datasets, without showing consistently competitive results across different scenarios. This reveals their vulnerability when dealing with Non-ICD data composed of incomplete and multi-granular subcluster distributions.
3) Average-rank and significance tests further corroborate the above findings, confirming that the improvements achieved by GOLD are statistically significant and robust.

To validate the consistency between the cluster assignments obtained by the FC and the ground-truth labels across clients, the overall federated clustering performance is measured by treating datasets from all clients as a whole.
Table~\ref{tbl:federated_performance_paper} reports the Purity and ARI performance, while the NMI and ACC performance are provided in Appendix~\ref{ap:performance}. Key observations include: 1) GOLD consistently outperforms its counterparts, highlighting its superiority in FC accuracy. Specifically, GOLD achieves the best on nearly all datasets w.r.t. the Purity index, except for the EP and PE datasets, where it still ranks second. For the cases where GOLD does not perform the best, i.e., the Purity index on the WI and the ARI index on the EP, the gap between GOLD and the best-performing counterpart is tiny, demonstrating its robustness and effectiveness across diverse scenarios.
2) For datasets with small $\lambda$, e.g., the EP dataset ($\lambda=0.02$), the performance gap between GOLD and the other methods is narrowed. This indicates that $\lambda$ serves as a reliable measure for quantifying the Non-ICD degree of a dataset. This also illustrates the robustness and competitiveness of GOLD even when the Non-ICD effect is extremely mild, thanks to its thorough multi-granularity information representation mechanism.
3) The significance tests confirm that GOLD demonstrates statistically significant performance over its counterparts.

\begin{figure}[!t]
    \centering
    \includegraphics[width=0.89\linewidth]{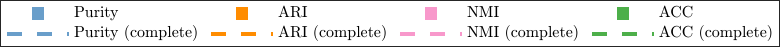} \\[1em]
    \includegraphics[width=0.47\linewidth]{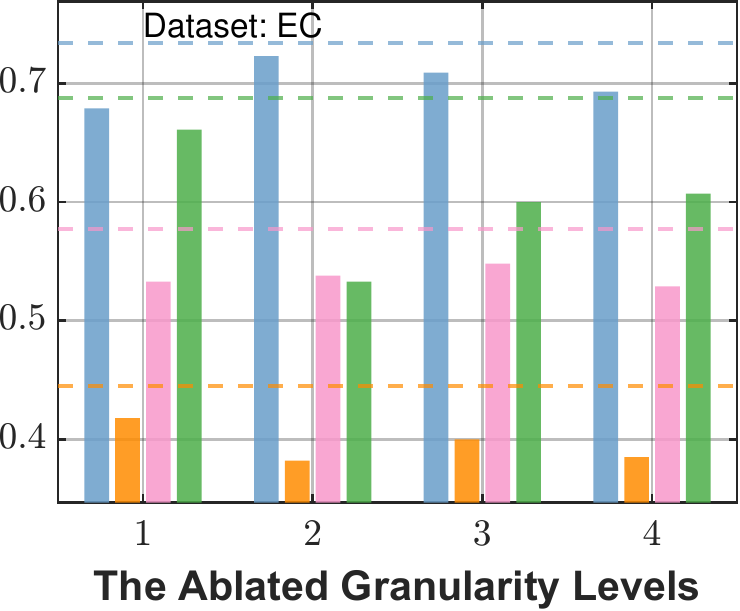}
    \includegraphics[width=0.47\linewidth]{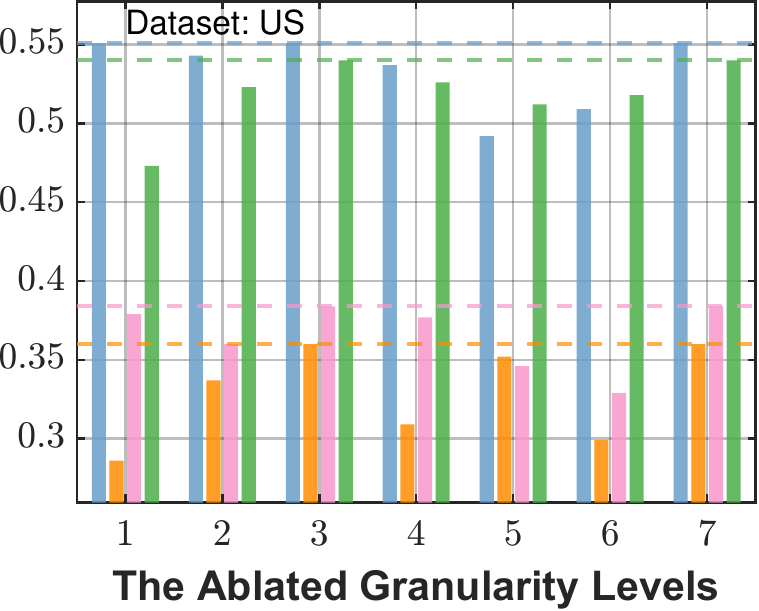}
    \includegraphics[width=0.47\linewidth]{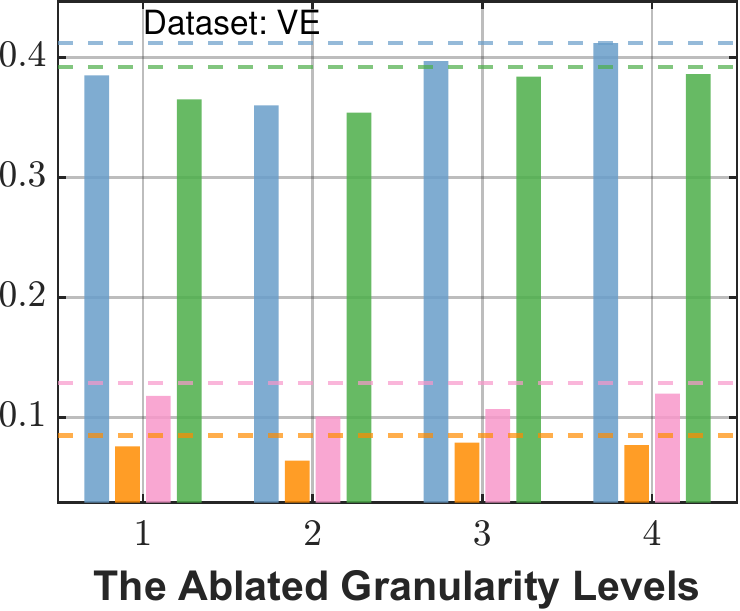}
    \includegraphics[width=0.47\linewidth]{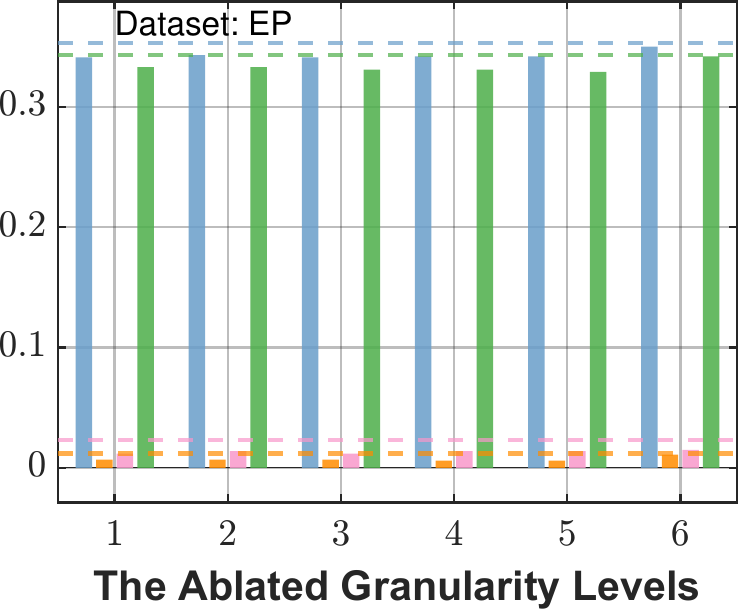}
    \includegraphics[width=0.47\linewidth]{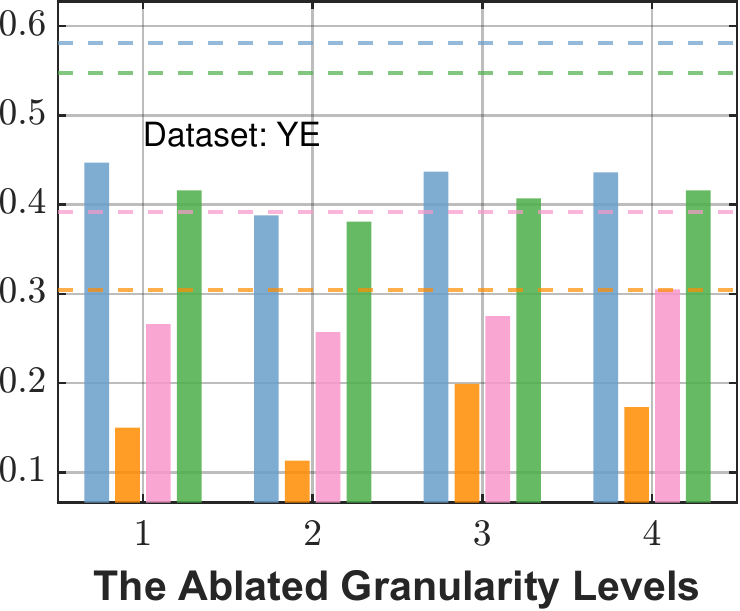}
    \includegraphics[width=0.47\linewidth]{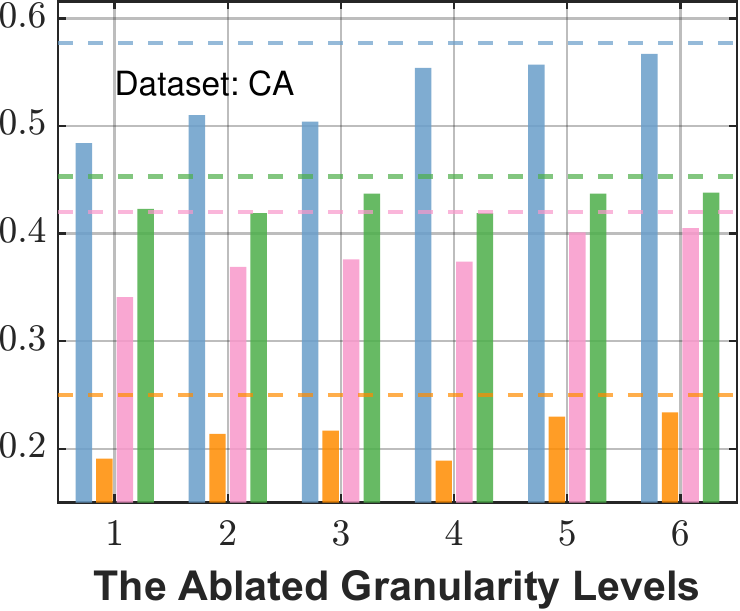}
    \caption{Single-granular representation ablation of GOLD, evaluated by 4 indices. The dashed lines represent the performance of the complete GOLD, while the bars show the performance after ablating each granularity level.}
    \label{fig:single granularity ablation}
\end{figure}

\subsection{Evaluation of the Impact of Non-ICD}

To evaluate the impact of federated clustering by different levels of data heterogeneity, GOLD is compared with its counterparts by varying the degree of Non-ICD, i.e., $\lambda$. As shown in Fig.~\ref{fig:non-icd-papers}, the clustering performance is evaluated using the Purity index on four datasets, with ARI index results provided in Appendix~\ref{ap:non_icd}. It can be seen that on datasets such as EC, US, and VE, baseline methods suffer significant performance drops as $\lambda$ increases, whereas GOLD exhibits remarkable robustness. On dataset YE, where the overall Non-ICD impact is milder, GOLD still outperforms others, suggesting its superior adaptability in FC. In short, GOLD performs well on extremely Non-ICD data and demonstrates its superior adaptability to different degrees of Non-ICD.

\subsection{Ablation Studies}

\begin{figure}[!t]
	\centering
	\includegraphics[width=0.9\linewidth]{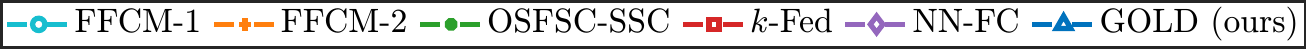} \\[1em]
	% EC
	\includegraphics[width=0.49\linewidth]{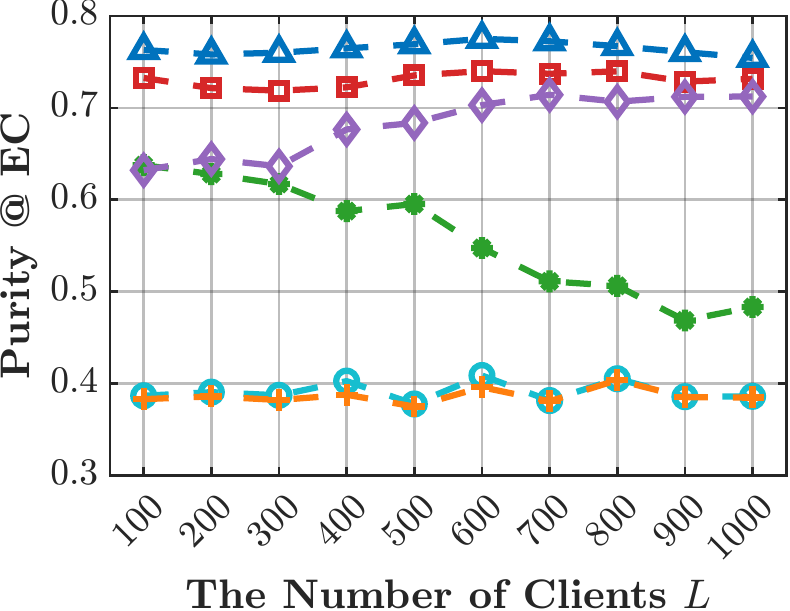}
	% US
	\includegraphics[width=0.49\linewidth]{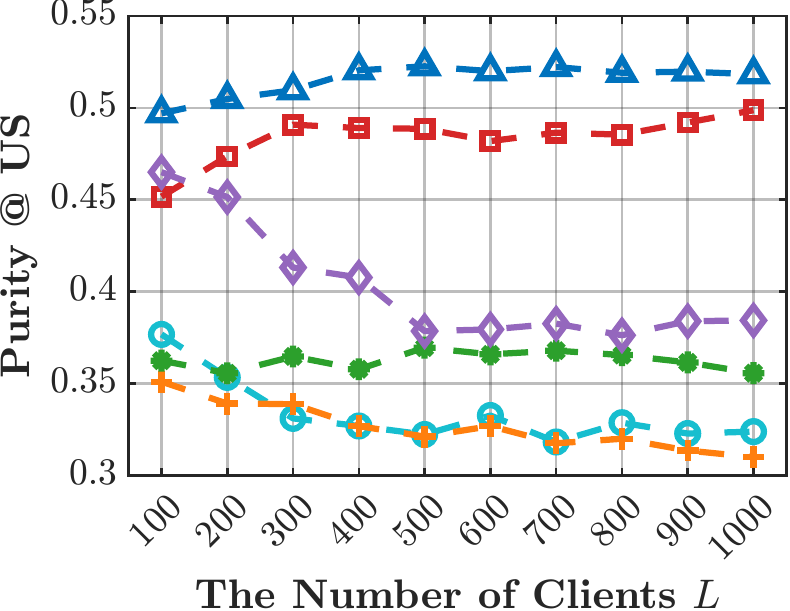}
	% VE
	\includegraphics[width=0.49\linewidth]{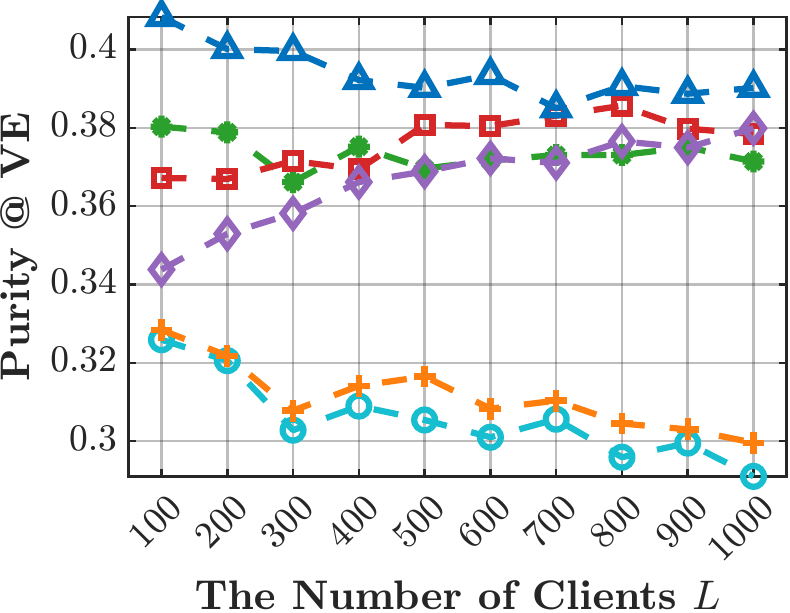}
	\includegraphics[width=0.49\linewidth]{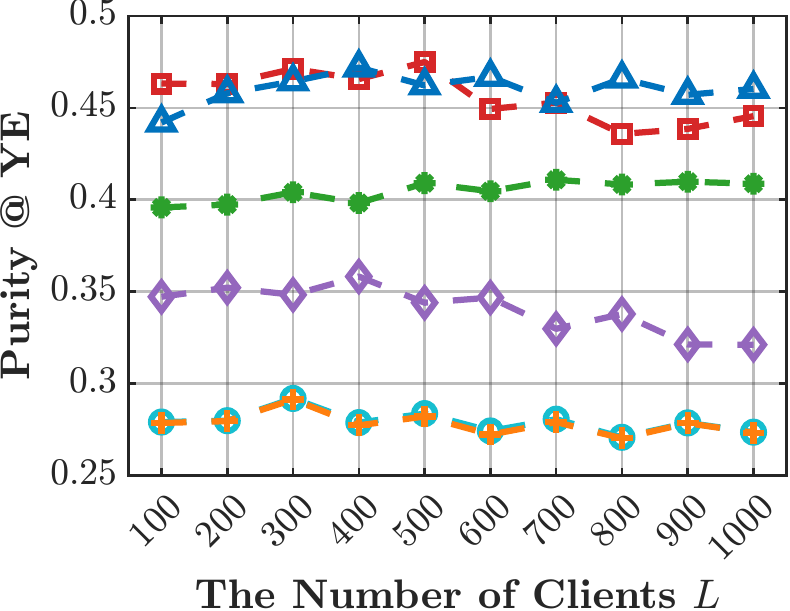}
	
	\caption{Performance comparison under different numbers of clients.}
	\label{fig:ablation_of_client_paper}
\end{figure}

\begin{figure}[!t]
	\centering
	\subfigure[Execution time w.r.t. $N$]{\includegraphics[width=0.49\linewidth]{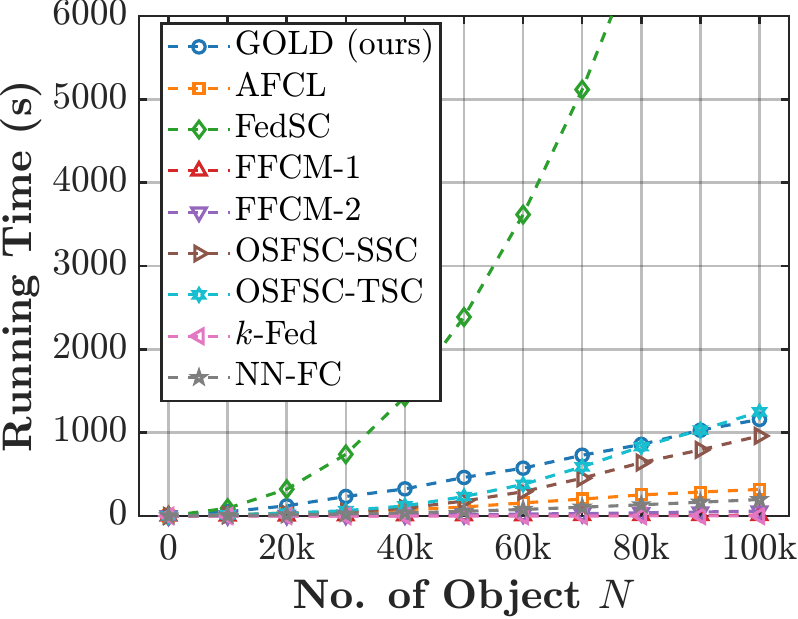}}
	\subfigure[Execution time w.r.t. $d$]{\includegraphics[width=0.47\linewidth]{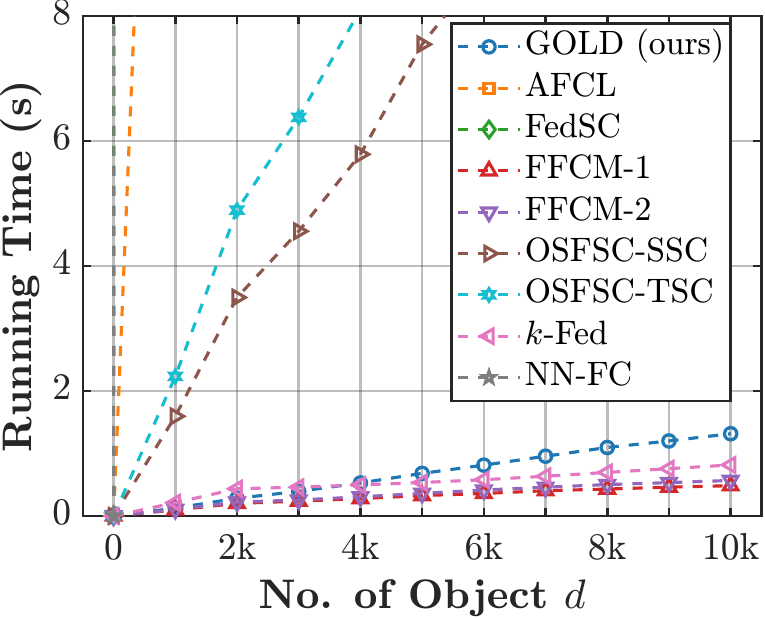}}
	\caption{Comparison of execution time of GOLD and its counterparts on (a) dataset SD with increasing $N$ and (b) dataset SD with increasing $d$.}
	\label{fig:efficiency}
\end{figure}

\textbf{Component Ablation:} 
% 介绍实验
Table~\ref{tbl:ablation_of_components} evaluates the contribution of GOLD's two core components: FCPL for clients and MCPL for the server. The symbol ``\Checkmark'' indicates that the corresponding component is enabled, while ``\XSolidBrush'' denotes its replacement with a conventional $k$-means algorithm. 
% 分析结果 
The results demonstrate two key insights: 1) Variants incorporating either FCPL or MCPL consistently outperform the baseline (neither component enabled) in 41 out of 48 comparisons, as shown by the light blue shading on the results. This confirms that each component contributes positively to FC performance. 2) The variant with both components activated achieves the best performance across all validity indices (dark blue shading), indicating that FCPL and MCPL effectively complement each other and exhibit quite good synergistic effect. These findings validate the necessity of the two key learning components of GOLD in FC tasks.

\begin{figure*}
    \centering
    \includegraphics[width=0.98\linewidth]{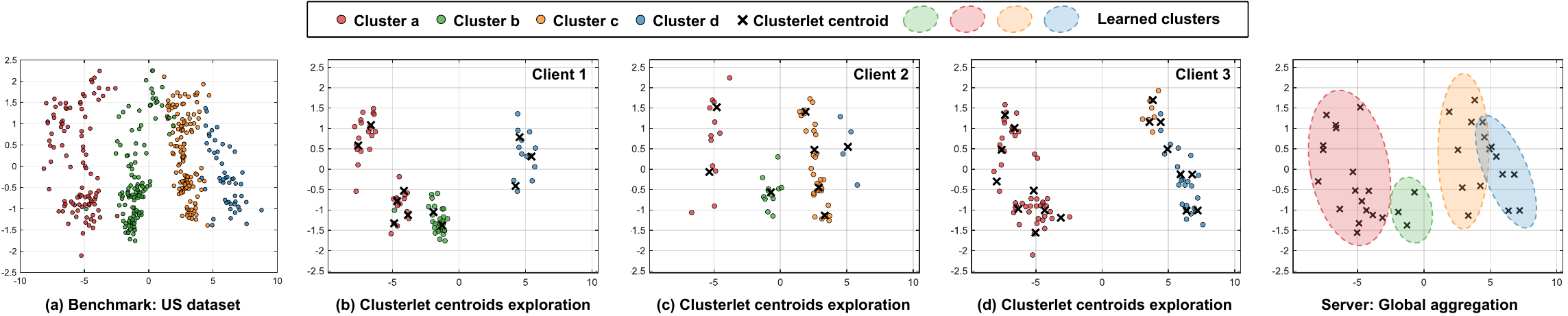}
    \caption{Federated clustering process of GOLD in a real-world scenario. (a) visualizes the US dataset. (b) - (d) illustrate the process of cluster exploration on three Non-ICD clients. (e) visualizes the global cluster distribution obtained by the server.}
    \label{fig:visualization}
\end{figure*}

\textbf{Accumulated Granularity Ablation:}
To quantify the contribution of different granularity levels, GOLD is compared with its granularity-restricted variants by varying the number of granularity levels involved in clustering, from using only the finest-grained level to incorporating all levels. In Fig.~\ref{fig:accumulated granularity ablation}, the right axis shows the converged number of clusters at each granularity level, while the left axis presents the performance across four evaluation metrics. It can be observed that the granularity converges progressively during training, revealing a strong positive correlation between the utilization of multiple granularity levels and clustering accuracy. This trend substantiates the effectiveness of MCPL in exploring the proper number of clusters under different granularities, and also illustrates that the diversity of granularity introduces richer information into data representations to enhance the clustering performance.

\textbf{Single Granularity Ablation:}
To further examine the role of individual granularity levels, each single granularity is ablated in the clustering process of GOLD, from the finest to the coarsest. In Fig.~\ref{fig:single granularity ablation}, the dashed lines indicate the performance of the complete GOLD, while the bars show the results after ablating each single granularity. By comparing the dashed lines and the bars in the same color, it can be found that the performance degradation occurs whenever any single granularity level is removed. This indicates that each level of granularity provides essential information to the clustering process. 
It can also be observed that different datasets exhibit varying levels of contribution from different granularities. This indicates that the datasets have distinct multi-granular Non-ICD patterns as described by Definition~\ref{def:non-icd}. Given GOLD's consistently dominant performance in prior comparisons, the results here provide specific evidence of GOLD's capability in extracting and utilizing multi-granular distribution information.

\subsection{Scalability Evaluation}
\label{experiment:scalability}

To comprehensively evaluate GOLD under realistic federated clustering scenarios, both its robustness across varying numbers of clients and its computational efficiency on large-scale datasets are evaluated. 

The scalability of GOLD is first evaluated by comparing
its performance with the counterparts under different numbers of clients $L$ ranging from 100 to 1000 (with a step size of 100). Fig.~\ref{fig:ablation_of_client_paper} presents the results on four representative datasets on the Purity index. More comprehensive results are provided in Appendix~\ref{ap:scalability}. 
It can be observed that GOLD outperforms its counterparts in general.
Although some counterparts achieve comparable performance on certain datasets, GOLD demonstrates overall superiority and more stable performance.
In addition, while the performance of both GOLD and its counterparts declines as the number of clients $L$ increases, GOLD consistently maintains superior performance and stability over other counterparts, which demonstrates the robustness and scalability of GOLD on federated settings.

The execution time of GOLD is compared with the counterparts on large-scale datasets, and a synthetic dataset SD is generated as an isotropic Gaussian mixture with 5 clusters. Two groups of comparisons are conducted on the SD dataset by varying the total number of its data objects $N$ and feature dimensionality $d$. 
In Fig.~\ref{fig:efficiency}(a), the SD dataset is evaluated with $d$ fixed at 10, while $N$ increases from 10k to 100k in increments of 10k. 
It can be observed that the execution time of GOLD is similar to the state-of-the-art FC methods, and grows slowly w.r.t. $N$. This confirms the linear time complexity of GOLD as analyzed in Theorem~\ref{theorem:time_complexity}, and indicates its scalability to the scale of data samples.
In Fig.~\ref{fig:efficiency}(b), the SD dataset is evaluated with $N$ fixed at 500, while $d$ increases from 1k to 10k in steps of 1k. The experimental results demonstrate the efficiency of GOLD compared to advanced methods. Specifically, in contrast to the state-of-the-art FedSC, which exhibits polynomial complexity, GOLD scales linearly w.r.t. both $N$ and $d$. In comparison with the conventional $k$-Fed, although GOLD incurs a marginally higher computational overhead, it demonstrates the crucial linear scalability. In summary, the multi-granular distribution exploration mechanisms of GOLD do not incur much extra computation overhead compared to the advanced methods, and GOLD has a considerably satisfactory scalability in processing large-scale and high-dimensional datasets. The above observations collectively verify the time complexity analyzed by Theorem~\ref{theorem:time_complexity}.

\subsection{Qualitative Results}
To provide an intuitive validation of GOLD’s effectiveness, its outputs are presented at both the client and server sides. Fig. \ref{fig:visualization}(a) depicts the ground-truth distribution of the US dataset, Figs. \ref{fig:visualization}(b)-(d) illustrate the local exploration results of client-side clusters, and Fig.~\ref{fig:visualization}(e) visualizes the globally aggregated clusters on the server side.  
%% 观察结果
It can be observed that even though the three clients possess Non-ICD data with varying granularity levels, GOLD can still automatically partition each local dataset into an appropriate number of compact, well-separated small clusters. Furthermore, the aggregated results on the server side reveal that GOLD can effectively align and merge these heterogeneous local clusters, yielding a coherent and representative global clustering structure.

\section{Conclusion}

This paper generalizes Non-IID FC into a more challenging but realistic Non-ICD scenario, where local clusters are highly dispersed across clients. The incomplete clusters within each single client bring performance bottlenecks to most existing approaches proposed for tackling the Non-IID FC tasks. To bridge the gap between FC and the more practical Non-ICD scenario, a novel one-shot FC framework called GOLD is proposed. It is capable of handling the challenging granularity misalignment and non-adjacent subclusters issues in Non-ICD data via efficient and secure one-way communication from clients to the server. On clients, GOLD discovers compact micro-clusters and captures potentially non-adjacent subclusters belonging to the same global cluster. On the server, GOLD enables local clusters to aggregate at multiple granularity levels and then fuses them into a coherent global distribution. The one-shot protocol minimizes communication overhead and privacy leakage to fulfill FC requirements, while the designed multi-granular exploration improves cluster representation quality to compensate for the information deficiency brought by the one-shot communication. Comprehensive experiments across diverse settings show consistent gains over existing FC methods. The next avenue of this work includes the FC under streaming and continual IoT scenarios, FC of heterogeneous modalities beyond tabular data, and dedicated higher-level privacy protection in FC.

\bibliographystyle{IEEEtran}
\bibliography{References}

@inproceedings{dennis2021heterogeneity,
  author    = {Dennis, D. K. and Li, T. and Smith, V.},
  title     = {Heterogeneity for the win: One-shot federated clustering},
  booktitle = {Proceedings of the 38th International Conference on Machine Learning},
  pages     = {2611--2620},
  year      = {2021}
}

@inproceedings{xie2023fed,
  author    = {Xie, S. and Wu, Y. and Liao, K. and others},
  title     = {{Fed-SC}: One-shot federated subspace clustering over high-dimensional data},
  booktitle = {Proceedings of the 39th IEEE International Conference on Data Engineering},
  pages     = {2905--2918},
  year      = {2023}
}

@inproceedings{chung2022federated,
  author    = {Chung, J. and Lee, K. and Ramchandran, K.},
  title     = {Federated unsupervised clustering with generative models},
  booktitle = {Proceedings of the AAAI Workshop on Trustable, Verifiable and Auditable Federated Learning},
  year      = {2022}
}

@inproceedings{hernandez2021federated,
  author    = {Hern{\'a}ndez-Pereira, E. and Fontenla-Romero, O. and Guijarro-Berdin{\~n}as, B. and others},
  title     = {Federated learning approach for spectral clustering},
  booktitle = {Proceedings of the 29th European Symposium on Artificial Neural Networks, Computational Intelligence and Machine Learning},
  year      = {2021}
}

@inproceedings{qiao2024federated,
  author    = {Qiao, D. and Ding, C. and Fan, J.},
  title     = {Federated spectral clustering via secure similarity reconstruction},
  booktitle = {Proceedings of the 38th International Conference on Neural Information Processing Systems},
  pages     = {58520--58555},
  year      = {2024}
}

@inproceedings{li2022federated,
  author    = {Li, Q. and Diao, Y. and Chen, Q. and others},
  title     = {Federated learning on non-{IID} data silos: An experimental study},
  booktitle = {Proceedings of the 38th IEEE International Conference on Data Engineering},
  pages     = {965--978},
  year      = {2022}
}

@inproceedings{ghosh2020efficient,
  author    = {Ghosh, A. and Chung, J. and Yin, D. and others},
  title     = {An efficient framework for clustered federated learning},
  booktitle = {Proceedings of the 24th International Conference on Neural Information Processing Systems},
  pages     = {19586--19597},
  year      = {2020}
}

@inproceedings{briggs2020federated,
  author    = {Briggs, C. and Fan, Z. and Andras, P.},
  title     = {Federated learning with hierarchical clustering of local updates to improve training on non-{IID} data},
  booktitle = {Proceedings of the 20th International Joint Conference on Neural Networks},
  pages     = {1--9},
  year      = {2020}
}

@inproceedings{qin2023fedapen,
  author    = {Qin, Z. and Deng, S. and Zhao, M. and others},
  title     = {{FedAPEN}: Personalized cross-silo federated learning with adaptability to statistical heterogeneity},
  booktitle = {Proceedings of the 29th ACM SIGKDD International Conference on Knowledge Discovery and Data Mining},
  pages     = {1954--1964},
  year      = {2023}
}

@inproceedings{jiang2023anomaly,
  author    = {Jiang, M. and Han, S. and Huang, H.},
  title     = {Anomaly detection with score distribution discrimination},
  booktitle = {Proceedings of the 29th ACM SIGKDD International Conference on Knowledge Discovery and Data Mining},
  pages     = {984--996},
  year      = {2023}
}

@inproceedings{yuan2023spatio,
  author    = {Yuan, Y. and Ding, J. and Shao, C. and others},
  title     = {Spatio-temporal diffusion point processes},
  booktitle = {Proceedings of the 29th ACM SIGKDD International Conference on Knowledge Discovery and Data Mining},
  pages     = {3173--3184},
  year      = {2023}
}

@inproceedings{akash2019inter,
  author    = {Akash, P. S. and Kadir, M. E. and Ali, A. A. and others},
  title     = {Inter-node Hellinger distance based decision tree},
  booktitle = {Proceedings of the 28th International Joint Conference on Artificial Intelligence},
  pages     = {1967--1973},
  year      = {2019}
}

@inproceedings{kumar2020federated,
  author    = {Kumar, H. H. and Karthik, V. R. and Nair, M. K.},
  title     = {Federated k-means clustering: A novel edge {AI}-based approach for privacy preservation},
  booktitle = {Proceedings of the 9th IEEE International Conference on Cloud Computing in Emerging Markets},
  pages     = {52--56},
  year      = {2020}
}

@inproceedings{cai2024robust,
  author    = {Cai, S. and Zhang, Y. and Luo, X. and others},
  title     = {Robust categorical data clustering guided by multi-granular competitive learning},
  booktitle = {Proceedings of the 44th IEEE International Conference on Distributed Computing Systems},
  pages     = {288--299},
  year      = {2024}
}

@inproceedings{zhang2025asynchronous,
  author    = {Zhang, Y. and Zhang, Y. and Lu, Y. and others},
  title     = {Asynchronous federated clustering with unknown number of clusters},
  booktitle = {Proceedings of the 39th AAAI Conference on Artificial Intelligence},
  pages     = {22695--22703},
  year      = {2025}
}

@inproceedings{sutter2020multimodal,
  author    = {Sutter, T. and Daunhawer, I. and Vogt, J.},
  title     = {Multimodal generative learning utilizing Jensen-Shannon-divergence},
  booktitle = {Proceedings of the 24th International Conference on Neural Information Processing Systems},
  pages     = {6100--6110},
  year      = {2020}
}

@inproceedings{heer2021fast,
  author    = {Heer, J.},
  title     = {Fast \& accurate {Gaussian} kernel density estimation},
  booktitle = {Proceedings of the 32nd IEEE Visualization Conference},
  pages     = {11--15},
  year      = {2021}
}

@inproceedings{0003SPRM23,
  author    = {C. Pan and J. Sima and S. Prakash and others},
  title     = {Machine Unlearning of Federated Clusters},
  booktitle = {Proceedings of the 11th International Conference on Learning Representations},
  year      = {2023}
}

@article{wang2020federated,
  author  = {Wang, H. and Li, A. and Shen, B. and others},
  title   = {Federated multi-view spectral clustering},
  journal = {IEEE Access},
  volume  = {8},
  pages   = {202249--202259},
  year    = {2020}
}

@article{pedrycz2021federated,
  author  = {Pedrycz, W.},
  title   = {Federated {FCM}: Clustering under privacy requirements},
  journal = {IEEE Transactions on Fuzzy Systems},
  volume  = {30},
  pages   = {3384--3388},
  year    = {2021}
}

@article{li2021survey,
  author  = {Li, Q. and Wen, Z. and Wu, Z. and others},
  title   = {A survey on federated learning systems: Vision, hype and reality for data privacy and protection},
  journal = {IEEE Transactions on Knowledge and Data Engineering},
  volume  = {35},
  pages   = {3347--3366},
  year    = {2021}
}

@article{yin2021comprehensive,
  author  = {Yin, X. and Zhu, Y. and Hu, J.},
  title   = {A comprehensive survey of privacy-preserving federated learning: A taxonomy, review, and future directions},
  journal = {ACM Computing Surveys},
  volume  = {54},
  pages   = {1--36},
  year    = {2021}
}

@article{yang2019federated,
  author  = {Yang, Q. and Liu, Y. and Chen, T. and others},
  title   = {Federated machine learning: Concept and applications},
  journal = {ACM Transactions on Intelligent Systems and Technology},
  volume  = {10},
  pages   = {1--19},
  year    = {2019}
}

@article{ye2023adaptive,
  author  = {Ye, Z. and Zhang, X. and Chen, X. and others},
  title   = {Adaptive clustering-based personalized federated learning framework for next {POI} recommendation with location noise},
  journal = {IEEE Transactions on Knowledge and Data Engineering},
  volume  = {35},
  pages   = {1--15},
  year    = {2023}
}

@article{chen2022feddual,
  author  = {Chen, Q. and Wang, Z. and Wang, H. and others},
  title   = {{FedDual}: Pair-wise gossip helps federated learning in large decentralized networks},
  journal = {IEEE Transactions on Information Forensics and Security},
  volume  = {18},
  pages   = {335--350},
  year    = {2022}
}

@article{yang2024federated,
  author  = {Yang, X. and Yu, H. and Gao, X. and others},
  title   = {Federated continual learning via knowledge fusion: A survey},
  journal = {IEEE Transactions on Knowledge and Data Engineering},
  volume  = {36},
  pages   = {1--18},
  year    = {2024}
}

@article{liu2023fedforgery,
  author  = {Liu, D. and Dang, Z. and Peng, C. and others},
  title   = {{FedForgery}: Generalized face forgery detection with residual federated learning},
  journal = {IEEE Transactions on Information Forensics and Security},
  volume  = {18},
  pages   = {1--15},
  year    = {2023}
}

@article{yang2022k,
  author  = {Yang, M. and Tjuawinata, I. and Lam, K.-Y.},
  title   = {K-means clustering with local d-privacy for privacy-preserving data analysis},
  journal = {IEEE Transactions on Information Forensics and Security},
  volume  = {17},
  pages   = {2524--2537},
  year    = {2022}
}

@article{chen2024lightweight,
  author  = {Chen, Z. and Yu, S. and Chen, F. and others},
  title   = {Lightweight privacy-preserving cross-cluster federated learning with heterogeneous data},
  journal = {IEEE Transactions on Information Forensics and Security},
  volume  = {19},
  pages   = {1--15},
  year    = {2024}
}

@article{acar2018survey,
  author  = {Acar, A. and Aksu, H. and Uluagac, A. S. and others},
  title   = {A survey on homomorphic encryption schemes: Theory and implementation},
  journal = {ACM Computing Surveys},
  volume  = {51},
  pages   = {1--35},
  year    = {2018}
}

@article{wei2020federated,
  author  = {Wei, K. and Li, J. and Ding, M. and others},
  title   = {Federated learning with differential privacy: Algorithms and performance analysis},
  journal = {IEEE Transactions on Information Forensics and Security},
  volume  = {15},
  pages   = {3454--3469},
  year    = {2020}
}

@article{sengar2008detecting,
  author  = {Sengar, H. and Wang, H. and Wijesekera, D. and others},
  title   = {Detecting {VoIP} floods using the Hellinger distance},
  journal = {IEEE Transactions on Parallel and Distributed Systems},
  volume  = {19},
  pages   = {794--805},
  year    = {2008}
}

@article{zhou2022pflf,
  author  = {Zhou, H. and Yang, G. and Dai, H. and others},
  title   = {{PFLF}: Privacy-preserving federated learning framework for edge computing},
  journal = {IEEE Transactions on Information Forensics and Security},
  volume  = {17},
  pages   = {1905--1918},
  year    = {2022}
}

@article{zhang2024prototype,
  author  = {Zhang, C. and Xie, Y. and Chen, T. and others},
  title   = {Prototype similarity distillation for communication-efficient federated unsupervised representation learning},
  journal = {IEEE Transactions on Knowledge and Data Engineering},
  volume  = {36},
  pages   = {1--15},
  year    = {2024}
}

@article{le2023privacy,
  author  = {Le, J. and Zhang, D. and Lei, X. and others},
  title   = {Privacy-preserving federated learning with malicious clients and honest-but-curious servers},
  journal = {IEEE Transactions on Information Forensics and Security},
  volume  = {18},
  pages   = {1--15},
  year    = {2023}
}

@article{wang2024one,
  author  = {Wang, Y. and Pang, W. and Pedrycz, W.},
  title   = {One-shot federated clustering based on stable distance relationships},
  journal = {IEEE Transactions on Industrial Informatics},
  volume  = {20},
  pages   = {1--12},
  year    = {2024}
}

@article{hubert1985comparing,
  author  = {Hubert, L. and Arabie, P.},
  title   = {Comparing partitions},
  journal = {Journal of Classification},
  volume  = {2},
  pages   = {193--218},
  year    = {1985}
}

@article{strehl2002cluster,
  author  = {Strehl, A. and Ghosh, J.},
  title   = {Cluster ensembles: A knowledge reuse framework for combining multiple partitions},
  journal = {Journal of Machine Learning Research},
  volume  = {3},
  pages   = {583--617},
  year    = {2002}
}

@article{Silhouettes1987,
  author  = {Rousseeuw, P. J.},
  title   = {Silhouettes: A graphical aid to the interpretation and validation of cluster analysis},
  journal = {Journal of Computational and Applied Mathematics},
  volume  = {20},
  pages   = {53--65},
  year    = {1987}
}

@article{CH1974,
  author  = {Cali{\'n}ski, T. and Harabasz, J.},
  title   = {A dendrite method for cluster analysis},
  journal = {Communications in Statistics: Theory and Methods},
  volume  = {3},
  pages   = {1--27},
  year    = {1974}
}

@article{zhangcais2024,
  author  = {Zhang, Y. and Zou, R. and Zhang, Y. and others},
  title   = {Adaptive micro-partition and hierarchical merging for accurate mixed data clustering},
  journal = {Complex and Intelligent Systems},
  volume  = {11},
  pages   = {1--14},
  year    = {2025}
}

@article{zhang2025learningSOM,
  title={Learning self-growth maps for fast and accurate imbalanced streaming data clustering},
  author={Zhang, Yiqun and Feng, Sen and Wang, Pengkai and Tan, Zexi and Luo, Xiaopeng and Ji, Yuzhu and Zou, Rong and Cheung, Yiu-Ming},
  journal={IEEE Transactions on Neural Networks and Learning Systems},
  year={2025},
  volume={36},
  number={9},
  pages={16049-16061},
  publisher={IEEE}
}

@article{hu2025significance,
  author  = {Hu, L. and Jiang, M. and Liu, X. and others},
  title   = {Significance-based decision tree for interpretable categorical data clustering},
  journal = {Information Sciences},
  volume  = {690},
  pages   = {121588},
  year    = {2025}
}

@article{li2021federated,
  author  = {Li, C. and Li, G. and Varshney, P. K.},
  title   = {Federated learning with soft clustering},
  journal = {IEEE Internet of Things Journal},
  volume  = {9},
  pages   = {7773--7782},
  year    = {2021}
}

@article{ma2024feduc,
  author  = {Ma, Q. and Xu, Y. and Xu, H. and others},
  title   = {{FedUC}: A unified clustering approach for hierarchical federated learning},
  journal = {IEEE Transactions on Mobile Computing},
  volume  = {23},
  pages   = {9737--9756},
  year    = {2024}
}

@article{li2023differentially,
  author  = {Li, Y. and Wang, S. and Chi, C.-Y. and others},
  title   = {Differentially private federated clustering over non-{IID} data},
  journal = {IEEE Internet of Things Journal},
  volume  = {11},
  pages   = {6705--6721},
  year    = {2023}
}

@article{ngo2023federated,
  author  = {Ngo, H. and Fang, H. and Rumbut, J. and others},
  title   = {Federated fuzzy clustering for decentralized incomplete longitudinal behavioral data},
  journal = {IEEE Internet of Things Journal},
  volume  = {11},
  pages   = {14657--14670},
  year    = {2023}
}

@article{zou2025sdenk,
  title={SDENK: Unbiased Subspace Density-k-Clustering},
  author={Zou, Rong and Zhang, Yunfan and Zhao, Mingjie and Tan, Zexi and Zhang, Yiqun and Cheung, Yiu-ming},
  journal={Neurocomputing},
  pages={131225},
  year={2025},
  publisher={Elsevier}
}

@article{zhang2025federated,
  title={Federated hierarchical clustering with automatic selection of optimal cluster numbers},
  author={Zhang, Yue and Qiu, Chuanlong and Liao, Xinfa and Zhang, Yiqun},
  journal={Information Sciences},
  pages={122957},
  year={2025},
  publisher={Elsevier}
}

@article{cheung2005maximum,
  author  = {Cheung, Y.-M.},
  title   = {Maximum weighted likelihood via rival penalized {EM} for density mixture clustering with automatic model selection},
  journal = {IEEE Transactions on Knowledge and Data Engineering},
  volume  = {17},
  pages   = {750--761},
  year    = {2005}
}

@article{wang2025one,
author={Y. Wang and W. Pang and D. Wang and others},
  title={One-shot secure federated K-means clustering based on density cores},
  journal={IEEE Transactions on Neural Networks and Learning Systems},
  volume={36},
  pages={14131--14143},
year={2025}
}

@article{huang2005automated,
author={J. Z. Huang and M. K. Ng and H. Rong and Z. Li},
title={Automated variable weighting in k-means type clustering},
journal={IEEE Transactions on Pattern Analysis and Machine Intelligence},
volume={27},
pages={657--668},
year={2005}
}

@article{stallmann2022towards,
  author  = {Stallmann, M. and Wilbik, A.},
  title   = {Towards federated clustering: A federated fuzzy $c$-means algorithm ({FFCM})},
  journal = {arXiv:2201.07316},
  year    = {2022}
}

@article{wang2020federatedArxiv,
  author  = {Wang, S. and Chang, T.-H.},
  title   = {Federated clustering via matrix factorization models: From model averaging to gradient sharing},
  journal = {arXiv:2002.04930},
  year    = {2020}
}

@article{kim2021comparing,
  author  = {T. Kim and J. Oh and N. Kim and others},
  title   = {Comparing Kullback-Leibler Divergence and Mean Squared Error Loss in Knowledge Distillation},
  journal = {arXiv:2105.08919},
  year    = {2021}
}

\begin{IEEEbiography}[{\includegraphics[width=1in,height=1.3in,keepaspectratio]{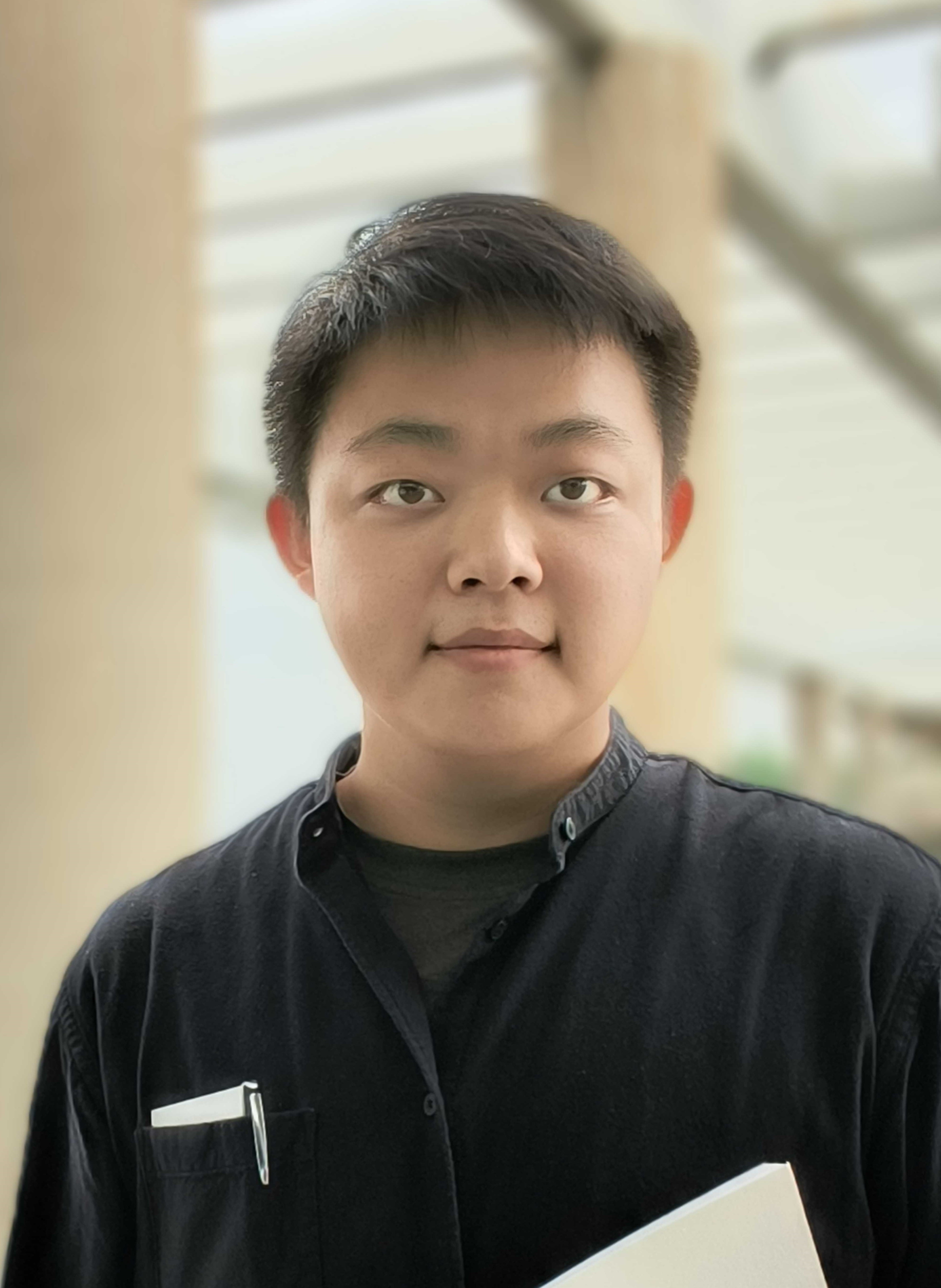}}]{Yiqun Zhang}
(Senior Member, IEEE) received the B.Eng. degree from South China University of Technology, Guangzhou, China, in 2013, and the M.S. and Ph.D. degrees from Hong Kong Baptist University, Hong Kong SAR, China, in 2014 and 2019, respectively. 
He is currently an Associate Professor with the School of Computer Science and Technology, Guangdong University of Technology, Guangzhou, China. 
His research works have been published in reputable journals and conferences, including TPAMI, SIGMOD, and SIGKDD, to name a few. 
His current research interests include machine learning and data science. 
Dr. Zhang serves as an Associate Editor for the \textit{IEEE Transactions on Emerging Topics in Computational Intelligence}. 
\end{IEEEbiography}

\begin{IEEEbiography}
[{\includegraphics[width=1in,height=1.5in,keepaspectratio]{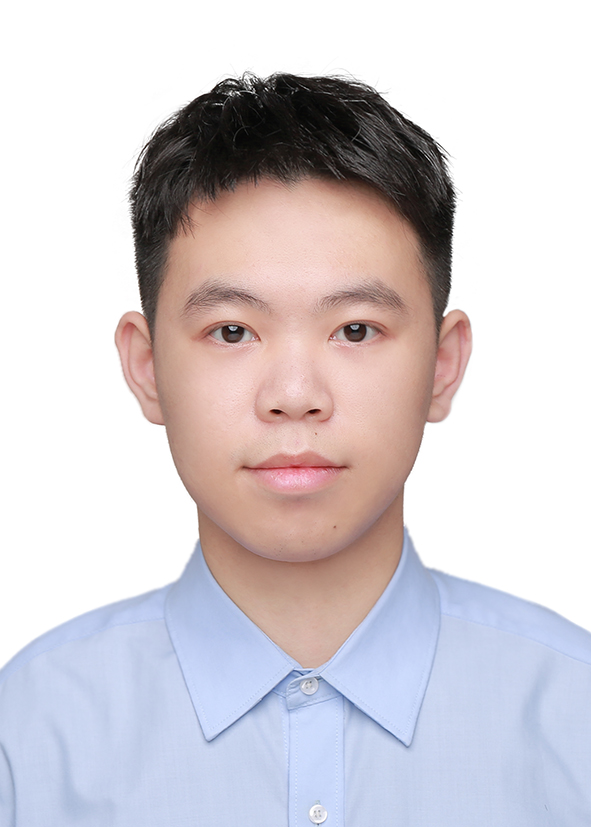}}]{Shenghong Cai}
received the B.Eng. degree from Guangdong University of Technology, Guangzhou, China, in 2025. He is now a research assistant with the Department of Computer Science, Beijing Normal-Hong Kong Baptist University. He has published his research works in reputable conferences like ICDCS.
His current research interests include unsupervised machine learning, deep learning, and federated clustering.
\end{IEEEbiography}

\begin{IEEEbiography}[{\includegraphics[width=1.05in,height=1.5in,keepaspectratio]{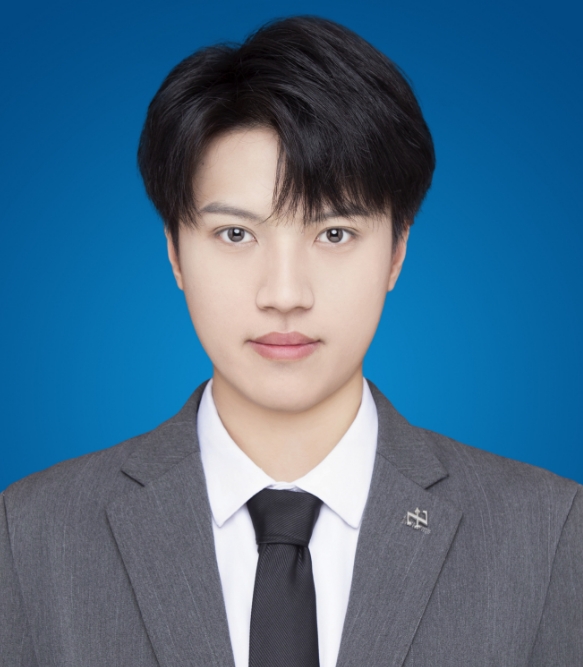}}]{Zihua Yang}
is currently pursuing a bachelor’s degree with the School of Computer Science and Technology, Guangdong University of Technology, Guangzhou, China. 
His current research interests include unsupervised machine learning and clustering of data with complex distributions.
\end{IEEEbiography}

\begin{IEEEbiography}[{\includegraphics[width=1in,height=1.5in,keepaspectratio]{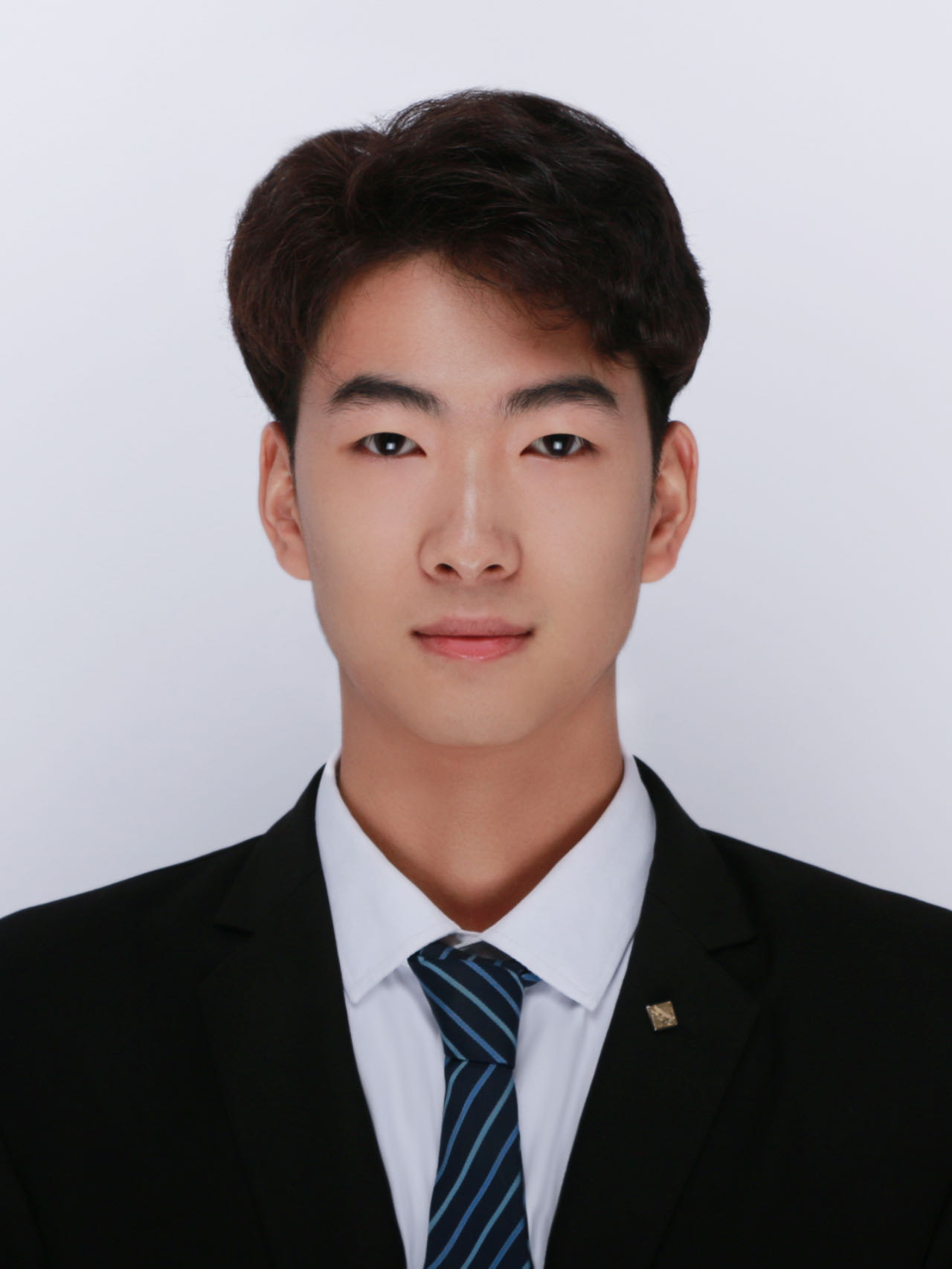}}]{Sen Feng}
received the B.Eng. degree from Henan University of Technology, Zhengzhou, China, in 2023. He is currently pursuing a master’s degree with the School of Computer Science and Technology, Guangdong University of Technology, Guangzhou, China. 
He has published a series of research works in reputable journals and conferences, including TNNLS, ECAI, and ICASSP. His current research interests include unsupervised machine learning, ensemble clustering, and streaming data clustering.
\end{IEEEbiography}

\begin{IEEEbiography}[{\includegraphics[width=1in,height=1.3in,keepaspectratio]{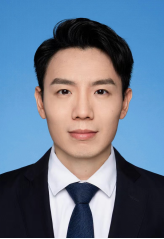}}]{Yuzhu Ji}
received a B.S. degree in computer science from the PLA Information Engineering University, Zhengzhou, China, in 2012, and the M.S. and Ph.D. degrees from the Department of Computer Science, Harbin Institute of Technology, Shenzhen, China, in 2015 and 2019. He is currently an Associate Professor at the School of Computer Science and Technology, Guangdong University of Technology, Guangzhou, China. 
His current research interests include salient object detection, image segmentation, and graph clustering.
\end{IEEEbiography}

\begin{IEEEbiography}[{\includegraphics[width=1in,height=1.5in,keepaspectratio]{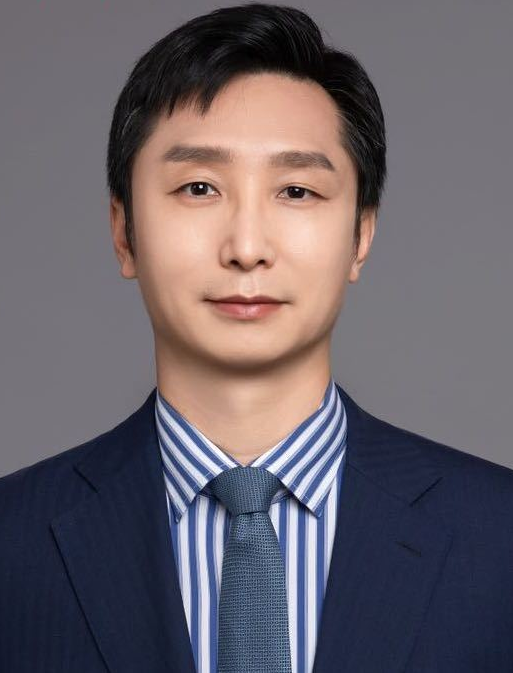}}]{Haijun Zhang} (Senior Member, IEEE) received the B.Eng. and master’s degrees from Northeastern University, Shenyang, China, in 2004 and 2007, respectively, and the Ph.D. degree from the Department of Electronic Engineering, City University of Hong Kong, Hong Kong, in 2010. He was a Post-Doctoral Research Fellow with the Department of Electrical and Computer Engineering, University of Windsor, Windsor, ON, Canada, from 2010 to 2011. Since 2012, he has been with the Harbin Institute of Technology, where he is currently a Professor of computer science. 
His current research interests include data mining, machine learning, fashion intelligence, and service computing. Prof. Zhang serves as an Associate Editor for the \textit{IEEE Transactions on Consumer Electronics}.
\end{IEEEbiography}

\clearpage
\appendices

\setcounter{section}{0}
\setcounter{figure}{0}
\setcounter{equation}{0}
\setcounter{theorem}{0}
\setcounter{lemma}{0}
\setcounter{table}{0}
\setcounter{remark}{0}
\setcounter{definition}{0}
\setcounter{algorithm}{0}

\renewcommand{\thesection}{\Roman{section}}
\renewcommand{\thesectiondis}{\Roman{section}}

\renewcommand{\thesubsection}{\Roman{section}.\Alph{subsection}}
\renewcommand{\thesubsectiondis}{\Alph{subsection}}

\renewcommand{\theequation}{A.\arabic{equation}}
\renewcommand{\thetheorem}{A.\arabic{theorem}}
\renewcommand{\thelemma}{A.\arabic{lemma}}
\renewcommand{\thetable}{A.\arabic{table}}
\renewcommand{\thefigure}{A.\arabic{figure}}
\renewcommand{\theremark}{A.\arabic{remark}}
\renewcommand{\thedefinition}{A.\arabic{definition}}
\renewcommand{\thealgorithm}{A.\arabic{algorithm}}

\section{Supplementary Details of the Proposed Method}
\label{ap:1}
This section presents additional theoretical analysis to further claim the underlying principles of the proposed method. Frequently used notations are summarized in Table~\ref{tbl:notation}.

\subsection{Execution Details of the Proposed Algorithm}
\label{ap:algorithm}

Consider $L$ clients, indexed by $l = \{ 1, \dots, L \}$, each holding a private local dataset $\mathbf{X}^{(l)} \in \mathbb{R}^{n^{(l)} \times d}$, where $n^{(l)}$ denotes the number of local data objects and $d$ is the feature dimensionality. 
In Competitive Penalized Learning (CPL), each dataset $\mathbf{X}^{(l)}$ is initialized with a set of candidate clusters, $C^{(l)} = \{C^{(l)}_j \mid 1 \le j \le k_0 \}$, where $k_0$ data objects are randomly selected as the initial cluster centroids.
All entries of the feature-cluster weight matrix $\mathbf{H}$ are initialized uniformly as $h_{jm} = \frac{1}{d}$, and all cluster weights are initialized as $w_j = \frac{1}{k_0}$, for $1 \le j \le k_0$ and $1 \le m \le d$.
Subsequently, each data object $x_{i}^{(l)} \in \mathbf{X}^{(l)}$ is assigned to its winning cluster according to Eq.~(\ref{eq:v}). The weights of the winning cluster $w_v$ and its nearest rival $w_r$ are then updated following Eqs.~(\ref{eq:v_weight})-(\ref{eq:client_weight}).
After all data objects have been assigned, the feature-cluster weight matrix $\mathbf{H}$ is updated using Eqs.~(\ref{eq:alpha_client})-(\ref{eq:client_feature_cluster_weight}).
It is noteworthy that when the weight of a cluster approaches zero, no data objects will be assigned to this cluster during subsequent learning epochs, and the corresponding feature-cluster weight becomes meaningless.
This competitive penalized learning process is repeated until the object-cluster affiliation matrix $\mathbf{Q}^{(l)}$ converges.
The complete clustering procedure is formally summarized in Algorithm~\ref{alg:CPL}.

To perform GOLD, each client executes Fine-grained Competitive Penalized Learning (FCPL) by initializing with a substantially large set of candidate clusters, $C^{(l)} = \{C^{(l)}_j \mid 1 \le j \le k_0 \}$, where $k_0$ is set to a sufficiently large value to activate CPL (Algorithm~\ref{alg:CPL}). After convergence, each client transmits its learned cluster centroids, denoted as $C^{(l)} = \{\mathbf{c}^{(l)}_j \mid 1 \le j \le k^{(l)} \}$, to the server.
The server then uncovers the multi-granular cluster distributions by executing Multi-granular Competitive Penalized Learning (MCPL), which recursively applies Algorithm~\ref{alg:CPL}. At each stage, the number of clusters $k$ inherited from the previously converged granularity is used to reinitialize the learning process.
This recursive exploration continues until the cluster number is stablilizes between consecutive iterations, i.e., $k_{new} = k_{old}$, and the objective function value converges, satisfying $\mid P_{new} - P_{old} \mid \le \epsilon$, where $\epsilon$ is a small threshold.
The resulting multi-granular object-cluster affiliation matrices $Q$ capture clustering distributions across different levels of granularity, ranging from fine to coarse.
To obtain the target clustering with $k^*$ clusters, these matrices are embedded into a data-enhanced representation $\mathbf{X}^{(e)}$ via Eq.~(\ref{eq:encode}).
For optimization of the objective function in Eq.~(\ref{eq:affiliation}), each entry of the feature-cluster weight matrix $\mathbf{U}$ is initialized uniformly as $u_{j\delta} = \frac{1}{\Delta}$, followed by a random initialization of $k^*$ clusters. 
The initial object-cluster assignment matrix $\mathbf{Q}$ is then computed using Eq.~(\ref{eq:server_q}). 
Subsequently, $\mathbf{U}$ is updated according to Eqs.~(\ref{eq:server_alpha})–(\ref{eq:server_feature_cluster_weight}), while $\mathbf{Q}$ is iteratively updated via Eq.~(\ref{eq:server_q}) until convergence. 
The overall federated clustering procedure under the GOLD framework is formally summarized in Algorithm~\ref{alg:GOLD}.

\begin{table}[t]
    \caption{Summary of notations.}
    \centering
    \begin{tabular} {c|l}
        \toprule
        \textbf{Notations} & \textbf{Explanations} \\
        \midrule
        $\mathbf{X}^{(l)}$ & Dataset of $l$-th client \\
        $C^{(l)}$ & Clusters set of $l$-th client \\
        $\mathbf{Q}^{(l)}$ & Object-cluster affiliation matrix w.r.t. $l$-th client \\
        $\mathbf{X}$ & Dataset of the server \\
        $\mathbf{Q}_{\delta}$ & Object-cluster affiliation matrix w.r.t. $\delta$-th granularity level \\
        $\mathbf{X}^{(e)}$ & Data-enhanced representation of the server \\
        $C$ & Clusters set of the server \\
        $\mathbf{Q}$ & Object-cluster affiliation matrix w.r.t. the server \\
        \bottomrule
    \end{tabular}
    \label{tbl:notation}
\end{table}

\begin{algorithm}[!t]
    \caption{CPL: Competitive Penalized Learning}
    \label{alg:CPL}
	\begin{algorithmic}[1]
		\REQUIRE Dataset $\mathbf{X}^{(l)}$, learning rate $\eta$, initialized $k_0$.
		\ENSURE Cluster centroids $C^{(l)}$.
        \STATE Initialize $convergence = false$, and randomly select $k_0$ data objects as the initial cluster centroids;
		\WHILE{$convergence = false$}
		  \FOR{$i = 1$ to $n^{(l)}$}
		  \STATE Determine the winning cluster $C_v^{(l)}$ and the rival cluster $C_r^{(l)}$ based on Eq.~(\ref{eq:v}) and Eq.~(\ref{eq:r});
            \STATE Assign $x_i^{(l)}$ to winning cluster $C_v^{(l)}$;
            \STATE Update cluster weights $W$ by Eqs.~(\ref{eq:v_weight})–(\ref{eq:client_weight}) and update winning time by Eq.~(\ref{eq:g});
		  \ENDFOR
            \IF{$\mathbf{Q}^{(l)}_{new} = \mathbf{Q}^{(l)}_{old}$}
		  \STATE Set $convergence = true$;
            \ELSE
		  \STATE Update $\mathbf{H}$ by Eqs.~(\ref{eq:alpha_client})-(\ref{eq:client_feature_cluster_weight});
            \ENDIF
		\ENDWHILE
	\end{algorithmic}
\end{algorithm}

\subsection{Proof of Space Complexity Analysis}
\label{ap:space}

\begin{theorem}
\label{theorem:space_complexity}
    The space complexity of GOLD comprises client-side $\mathcal{O}(L n^{(l)} (d + k_0))$ and server-side $\mathcal{O}(n(d+k_0))$, yielding an overall space complexity of GOLD being $\mathcal{O}((d+k_0)(Ln^{(l)}+n))$.
\end{theorem}

\begin{lemma}
\label{lemma:client_space}
The aggregate space complexity of the client-side FCPL algorithm is $\mathcal{O}(L n^{(l)} (d + k_0))$.
\end{lemma}

\begin{prf}
For a client's dataset $\mathbf{X}^{(l)}$ containing $n^{(l)}$ data objects and $d$ features, the space complexity for the input and output data is $(n^{(l)} d)$ and $\mathcal{O}(n^{(l)})$, respectively. During the computation of object-cluster similarities, $k_0$ is used as the initial number of clusters. Since $n^{(l)} \times k_0$ distance pairs must be computed, the corresponding space complexity is $\mathcal{O}(n^{(l)} k_0)$. Similarly, the feature-cluster importance matrix $\mathbf{H}$ requires $\mathcal{O}(k_0 d)$ space in total. Considering $L$ clients in total, the overall space complexity across all clients can be expressed as $\mathcal{O}(L n^{(l)} (d + k_0))$.
\qed
\end{prf}

\begin{lemma}
\label{lemma:server_space}
The overall space complexity of the server-side algorithms, comprising MCPL and REMC, is $\mathcal{O}(n(d+k_0))$.
\end{lemma}

\begin{prf}
Consider a server-side dataset $\mathbf{X}$, which comprises $n$ objects, each represented by $d$ features. The space complexity for storing the input data is $(n d)$, while that for the final output is $\mathcal{O}(n)$. Suppose the clustering process involves $\Delta$ iterations to explore multi-granular distributions. Each iteration requires temporary storage of size $\mathcal{O}(n(d + k_0))$, as also observed in the client-side analysis. Additionally, the data-enhanced representation matrix $\mathbf{X}^{(e)}$ has dimensions $n \times \Delta$, incurring a complexity of $\mathcal{O}(n \Delta)$. For computing object-cluster similarities, one maintains a similarity matrix of size $n \times k^*$ and a feature-cluster weight matrix of size $\Delta \times k^*$, requiring $\mathcal{O}(n k^*)$ and $\mathcal{O}(k^* \Delta)$ memory, respectively. However, as $\Delta$ and $k^*$ are typically small compared to other variables, these terms are asymptotically negligible. Therefore, the overall server-side space complexity simplifies to $\mathcal{O}(n(d + k_0))$.
\qed
\end{prf}

\begin{algorithm}[!t]
    \caption{\small{GOLD: Global Oriented Local Distribution learning}}
    \label{alg:GOLD}
	\begin{algorithmic}[1]	
		\REQUIRE Local Dataset $X = \{ \mathbf{X}^{(l)} \mid 1 \le l \le L \}$ and learning rate $\eta$.
		\ENSURE Partition matrix $\mathbf{Q}$.
            \STATE \textit{// Phase 1: FCPL}
            \FORALL{clients in $X$}
                \STATE Run Algorithm~\ref{alg:CPL} with $\mathbf{X}^{(l)}$;
                \STATE Transfer cluster centroids set $C^{(l)}$ to the server;
            \ENDFOR
            \STATE \textit{// Phase 2: MCPL}
            \STATE Initialize $convergence = false$;
            \WHILE{$convergence = false$}
                \STATE Run Algorithm~\ref{alg:CPL} with cluster centroids matrix $\mathbf{X}$;
                \IF{$k_{old} = k_{new}$ and $|P_{new}-P_{old}| \le \epsilon$}
                    \STATE Set $convergence = true$;
                \ELSE
                    \STATE $Q \leftarrow Q \cup \{\mathbf{Q}_{new}\};$ 
                        $K \leftarrow K \cup \{k_{new}\};$ 
                         $k_{old} \leftarrow k_{new};$
                \ENDIF
            \ENDWHILE
            \STATE \textit{// Phase 3: REMC}
            \FOR{$\delta = 1$ to $\Delta$}
                \STATE Transform $\mathbf{Q_\delta}$ into $\delta$-features of $\mathbf{X}^{(e)}$ by Eq.~(\ref{eq:encode});
            \ENDFOR
            \STATE Initialize $convergence = false$;
            \STATE Update $\tilde{\mathbf{Q}}$ according to Eq.~(\ref{eq:server_q});
            \WHILE{$convergence = false$}
                \STATE Set $\mathbf{Q} = \tilde{\mathbf{Q}}$, update $\tilde{\mathbf{U}}$ by Eqs.~(\ref{eq:server_alpha})-(\ref{eq:server_feature_cluster_weight});
                \STATE Set $\mathbf{U} = \tilde{\mathbf{U}}$, update $\tilde{\mathbf{Q}}$ by Eq.~(\ref{eq:server_q}).
                \IF{$\mathbf{Q} = \tilde{\mathbf{Q}}$}
                    \STATE Set $convergence = true$;
                \ENDIF
            \ENDWHILE
    	\end{algorithmic}
\end{algorithm}

\begin{table*}[!t]
    \centering
    \renewcommand{\arraystretch}{1.2}
    \caption{Global clustering performance evaluated by NMI and ACC. ``$\overline{AR}(ST)$'' row reports the average performance ranks and the significance test results, where a significant difference between the corresponding method and GOLD is indicated by ``+''.}
    \label{tbl:centralized_performance_appendix_external}
    \scalebox{0.95}{
    \begin{tabular}{c|c|c c c c c c c c|c}
    \toprule
    \textbf{Index} & \textbf{Data} & \textbf{$k$-Fed} & \textbf{FFCM-avg1} & \textbf{FFCM-avg2} & \textbf{OSFSC-SSC} & \textbf{OSFSC-TSC} & \textbf{FedSC} & \textbf{NN-FC} & \textbf{AFCL} & \textbf{GOLD}\\
    \text{} & \text{} & (2021~\cite{dennis2021heterogeneity}) & (2022~\cite{stallmann2022towards}) & (2022~\cite{stallmann2022towards}) & (2023~\cite{xie2023fed}) & (2023~\cite{xie2023fed}) & (2024~\cite{qiao2024federated}) & (2024~\cite{wang2024one}) & (2025~\cite{zhang2025asynchronous}) & (ours)\\ 
    \midrule
    \multirow{10}{*}{NMI} 
    & EC & 0.590±0.02       & 0.485±0.04          & 0.511±0.01       & 0.595±0.03          & \cellcolor[RGB]{255,243,218}0.623±0.01 & 0.469±0.10       & 0.583±0.05       & 0.078±0.03 & \cellcolor[RGB]{255,228,173}0.647±0.04 \\
    & US & 0.171±0.02       & \cellcolor[RGB]{255,243,218}0.199±0.02    & \cellcolor[RGB]{255,243,218}0.199±0.01 & 0.183±0.04          & 0.175±0.00       & 0.176±0.02       & 0.170±0.04       & 0.000±0.00 & \cellcolor[RGB]{255,228,173}0.219±0.06 \\
    & VE & 0.103±0.02       & 0.080±0.02          & 0.095±0.01       & 0.103±0.01          & 0.108±0.00       & 0.105±0.00       & \cellcolor[RGB]{255,243,218}0.109±0.03 & 0.013±0.00 & \cellcolor[RGB]{255,228,173}0.112±0.02 \\
    & EP & 0.002±0.00       & \cellcolor[RGB]{255,243,218}0.003±0.00    & \cellcolor[RGB]{255,243,218}0.003±0.00 & \cellcolor[RGB]{255,243,218}0.003±0.00    & \cellcolor[RGB]{255,243,218}0.003±0.00 & \cellcolor[RGB]{255,243,218}0.003±0.00       & \cellcolor[RGB]{255,243,218}0.003±0.00 & 0.000±0.00 & \cellcolor[RGB]{255,228,173}0.004±0.00 \\
    & YE & \cellcolor[RGB]{255,243,218}0.263±0.03 & 0.235±0.01          & 0.216±0.03       & 0.258±0.01          & 0.260±0.00       & 0.189±0.06       & 0.219±0.04       & 0.005±0.00 & \cellcolor[RGB]{255,228,173}0.267±0.03 \\
    & CA & 0.323±0.02       & 0.280±0.01          & 0.221±0.00       & \cellcolor[RGB]{255,243,218}0.330±0.01    & 0.329±0.00       & 0.236±0.04       & 0.250±0.06       & 0.000±0.00 & \cellcolor[RGB]{255,228,173}0.331±0.02 \\
    & LA & 0.488±0.03       & \cellcolor[RGB]{255,228,173}0.535±0.00 & 0.418±0.01       & 0.511±0.00          & 0.491±0.00       & 0.463±0.02       & \cellcolor[RGB]{255,243,218}0.524±0.04 & 0.128±0.04 & \cellcolor[RGB]{255,243,218}0.524±0.06    \\
    & WI & 0.052±0.01       & \cellcolor[RGB]{255,228,173}0.056±0.00 & 0.033±0.01       & 0.053±0.01          & \cellcolor[RGB]{255,243,218}0.054±0.00 & 0.012±0.01       & 0.021±0.01       & 0.000±0.00 & \cellcolor[RGB]{255,228,173}0.056±0.01 \\
    & PE & \cellcolor[RGB]{255,243,218}0.612±0.04 & 0.490±0.00          & 0.492±0.04       & \cellcolor[RGB]{255,228,173}0.613±0.01 & 0.607±0.02       & 0.562±0.03       & \cellcolor[RGB]{255,243,218}0.612±0.03 & 0.000±0.00 & 0.609±0.04          \\
    & LE & \cellcolor[RGB]{255,243,218}0.348±0.01 & 0.251±0.01          & 0.033±0.00       & 0.342±0.01          & 0.344±0.01       & 0.252±0.02       & 0.339±0.01       & 0.000±0.00 & \cellcolor[RGB]{255,228,173}0.350±0.01 \\
    \midrule
    \multicolumn{2}{c|}{$\overline{AR}(ST)$}
    & 4.6(+) & 4.6(+) & 6.3(+) & 3.6(+) & 3.9(+) & 6.3(+) & 5.1(+) & 9.0(+) & 1.4 \\
    \midrule
    \multirow{10}{*}{ACC} 
    & EC & 0.665±0.05       & 0.617±0.06          & 0.645±0.04          & \cellcolor[RGB]{255,243,218}0.722±0.05 & 0.719±0.01 & 0.603±0.07 & 0.686±0.04       & 0.442±0.00 & \cellcolor[RGB]{255,228,173}0.732±0.08 \\
    & US & 0.414±0.03       & \cellcolor[RGB]{255,243,218}0.443±0.02    & 0.435±0.02          & 0.417±0.04       & 0.378±0.00 & 0.411±0.02 & 0.414±0.03       & 0.320±0.00 & \cellcolor[RGB]{255,228,173}0.450±0.04 \\
    & VE & 0.360±0.01       & 0.369±0.02          & \cellcolor[RGB]{255,228,173}0.380±0.01 & 0.363±0.01       & 0.370±0.00 & 0.357±0.01 & 0.371±0.02       & 0.273±0.00 & \cellcolor[RGB]{255,243,218}0.379±0.01    \\
    & EP & 0.269±0.00       & \cellcolor[RGB]{255,243,218}0.273±0.00    & 0.270±0.00          & \cellcolor[RGB]{255,243,218}0.273±0.01 & 0.271±0.01 & 0.266±0.00 & \cellcolor[RGB]{255,243,218}0.273±0.00 & 0.261±0.00 & \cellcolor[RGB]{255,228,173}0.274±0.01 \\
    & YE & 0.396±0.05       & 0.400±0.02          & 0.391±0.03          & \cellcolor[RGB]{255,243,218}0.403±0.03 & 0.387±0.01 & 0.341±0.04 & 0.391±0.04       & 0.312±0.00 & \cellcolor[RGB]{255,228,173}0.405±0.05 \\
    & CA & \cellcolor[RGB]{255,243,218}0.366±0.02 & 0.311±0.01          & 0.324±0.00          & 0.360±0.01       & 0.346±0.01 & 0.340±0.02 & 0.335±0.04       & 0.272±0.00 & \cellcolor[RGB]{255,228,173}0.368±0.03 \\
    & LA & 0.566±0.05       & 0.617±0.00          & 0.501±0.02          & 0.613±0.00       & 0.613±0.00 & 0.585±0.04 & \cellcolor[RGB]{255,243,218}0.625±0.05 & 0.323±0.02 & \cellcolor[RGB]{255,228,173}0.626±0.07 \\
    & WI & 0.278±0.01       & \cellcolor[RGB]{255,228,173}0.318±0.00 & 0.268±0.02          & 0.246±0.01       & 0.249±0.00 & 0.222±0.02 & \cellcolor[RGB]{255,243,218}0.304±0.02 & 0.237±0.00 & 0.296±0.03          \\
    & PE & \cellcolor[RGB]{255,243,218}0.575±0.06 & 0.378±0.01          & 0.415±0.04          & 0.558±0.03       & 0.534±0.02 & 0.493±0.05 & 0.562±0.05       & 0.104±0.00 & \cellcolor[RGB]{255,228,173}0.579±0.06 \\
    & LE & 0.240±0.01       & 0.181±0.01          & 0.058±0.00          & 0.235±0.01       & 0.250±0.01 & 0.182±0.01 & \cellcolor[RGB]{255,243,218}0.254±0.01 & 0.041±0.00 & \cellcolor[RGB]{255,228,173}0.257±0.01 \\
    \midrule
    \multicolumn{2}{c|}{$\overline{AR}(ST)$}
    & 4.7(+) & 4.6(+) & 5.6(+) & 4.0(+) & 5.0(+) & 7.1(+) & 3.8(+) & 8.9(+) & 1.3 \\
    \bottomrule
    \end{tabular}
    }
\end{table*}

\begin{table*}[!t]
    \centering
    \renewcommand{\arraystretch}{1.2}
    \caption{Global clustering performance evaluated by SC and CH. ``$\overline{AR}(ST)$'' row reports the average performance ranks and the significance test results, where a significant difference between the corresponding method and GOLD is indicated by ``+''.}
    \label{tbl:centralized_performance_appendix_internal}
    \scalebox{0.95}{
    \begin{tabular}{c|c|c c c c c c c c|c}
    \toprule
    \textbf{Index} & \textbf{Data} & \textbf{$k$-Fed} & \textbf{FFCM-avg1} & \textbf{FFCM-avg2} & \textbf{OSFSC-SSC} & \textbf{OSFSC-TSC} & \textbf{FedSC} & \textbf{NN-FC} & \textbf{AFCL} & \textbf{GOLD}\\
    \text{} & \text{} & (2021~\cite{dennis2021heterogeneity}) & (2022~\cite{stallmann2022towards}) & (2022~\cite{stallmann2022towards}) & (2023~\cite{xie2023fed}) & (2023~\cite{xie2023fed}) & (2024~\cite{qiao2024federated}) & (2024~\cite{wang2024one}) & (2025~\cite{zhang2025asynchronous}) & (ours)\\ 
    \midrule
    \multirow{10}{*}{SC} 
    & EC & 0.264±0.03          & 0.171±0.05 & 0.181±0.02       & 0.255±0.04 & \cellcolor[RGB]{255,243,218}0.284±0.01 & 0.141±0.06          & 0.278±0.04       & 0.273±0.04 & \cellcolor[RGB]{255,228,173}0.288±0.05 \\
    & US & \cellcolor[RGB]{255,228,173}0.148±0.01 & 0.125±0.02 & 0.134±0.01       & 0.135±0.01 & 0.145±0.00       & 0.073±0.03          & 0.135±0.02       & 0.000±0.00 & \cellcolor[RGB]{255,243,218} 0.147±0.01    \\
    & VE & 0.222±0.03          & 0.202±0.01 & 0.207±0.01       & 0.225±0.01 & 0.106±0.00       & 0.205±0.00          & \cellcolor[RGB]{255,243,218}0.229±0.04 & 0.024±0.01 & \cellcolor[RGB]{255,228,173}0.241±0.03 \\
    & EP & 0.020±0.00          & 0.025±0.00 & 0.022±0.01       & 0.023±0.00 & 0.022±0.00       & 0.027±0.03          & \cellcolor[RGB]{255,243,218}0.017±0.01 & 0.000±0.00 & \cellcolor[RGB]{255,228,173}0.028±0.01 \\
    & YE & \cellcolor[RGB]{255,243,218}0.166±0.03 & 0.063±0.01 & 0.133±0.03       & 0.147±0.02 & 0.153±0.01       & 0.127±0.16          & 0.161±0.04       & 0.101±0.05 & \cellcolor[RGB]{255,228,173}0.167±0.03 \\
    & CA & 0.110±0.01          & 0.081±0.01 & 0.129±0.00       & 0.090±0.01 & 0.103±0.00       & \cellcolor[RGB]{255,228,173}0.139±0.05 & 0.121±0.02       & 0.000±0.00 & \cellcolor[RGB]{255,243,218} 0.132±0.02    \\
    & LA & 0.294±0.01          & 0.284±0.00 & \cellcolor[RGB]{255,243,218}0.312±0.05 & 0.309±0.00 & 0.308±0.00       & 0.301±0.02          & 0.306±0.03       & 0.131±0.04 & \cellcolor[RGB]{255,228,173}0.316±0.01 \\
    & WI & \cellcolor[RGB]{255,243,218}0.159±0.01    & 0.117±0.01 & 0.105±0.02       & 0.127±0.02 & 0.154±0.00       & 0.142±0.12          & 0.150±0.02       & 0.000±0.00 & \cellcolor[RGB]{255,228,173}0.162±0.02 \\
    & PE & 0.229±0.02          & 0.112±0.00 & 0.194±0.02       & 0.229±0.01 & 0.221±0.01       & 0.225±0.02          & \cellcolor[RGB]{255,243,218}0.230±0.03 & 0.000±0.00 & \cellcolor[RGB]{255,228,173}0.234±0.03 \\
    & LE & \cellcolor[RGB]{255,243,218}0.112±0.01    & 0.021±0.01 & 0.084±0.00       & 0.107±0.01 & 0.111±0.00       & \cellcolor[RGB]{255,228,173}0.117±0.01 & 0.109±0.00       & 0.000±0.00 & \cellcolor[RGB]{255,228,173}0.117±0.01 \\
    \midrule
    \multicolumn{2}{c|}{$\overline{AR}(ST)$}
    & 4.0(+) & 7.3(+) & 5.6(+) & 4.7(+) & 4.6(+) & 5.0(+) & 4.1(+) & 8.4(+) & 1.2 \\
    \midrule
    \multirow{10}{*}{CH} 
    & EC & 4.359±1.14       & 4.483±2.32 & 4.439±2.23       & \cellcolor[RGB]{255,243,218}4.754±2.30 & 4.739±1.59       & 3.620±2.59 & 4.696±2.83 & 4.232±5.55  & \cellcolor[RGB]{255,228,173}4.882±3.24 \\
    & US & \cellcolor[RGB]{255,243,218}4.213±1.38 & 4.010±1.76 & 3.938±1.80       & 4.208±1.51       & 4.199±0.10       & 3.840±2.21 & 4.135±1.93 & 0.000±0.00  & \cellcolor[RGB]{255,228,173}4.263±1.94 \\
    & VE & \cellcolor[RGB]{255,243,218}5.820±3.26 & 5.724±1.80 & 5.730±2.46       & 5.778±3.29       & 5.437±0.40       & 5.812±0.31 & 5.775±4.13 & 3.543±9.33  & \cellcolor[RGB]{255,228,173}5.828±3.92 \\
    & EP & \cellcolor[RGB]{255,228,173}3.750±0.50       & 3.653±0.75 & 3.635±1.06 & 3.555±1.38       & 3.501±1.63       & 3.079±2.12 & 3.306±1.74 & 0.000±0.00  & \cellcolor[RGB]{255,243,218}3.704±1.14 \\
    & YE & 5.467±3.04       & 4.955±2.56 & 4.800±2.25       & \cellcolor[RGB]{255,243,218}5.497±2.46 & 5.468±2.00       & 4.591±3.13 & 5.211±3.14 & 3.355±8.37  & \cellcolor[RGB]{255,228,173}5.511±4.04 \\
    & CA & \cellcolor[RGB]{255,243,218}5.397±2.67 & 5.154±1.56 & 4.380±0.00       & 5.325±2.63       & 5.393±0.77       & 4.653±2.47 & 5.082±3.07 & 0.000±0.00  & \cellcolor[RGB]{255,228,173}5.404±3.05 \\
    & LA & 8.189±5.28       & 8.053±3.43 & 7.828±4.97       & 8.211±0.78       & \cellcolor[RGB]{255,243,218}8.218±0.38 & 8.000±5.00 & 8.118±5.49 & 7.465±15.88 & \cellcolor[RGB]{255,228,173}8.219±6.41 \\
    & WI & \cellcolor[RGB]{255,243,218}6.917±4.22       & 6.661±2.83 & 6.231±4.74       & 6.852±4.83       & 6.853±1.72       & 5.455±4.77 & 6.234±4.21 & 0.000±0.00  & \cellcolor[RGB]{255,228,173}6.967±4.67 \\
    & PE & 7.534±4.86       & 6.948±3.10 & 7.046±4.83       & \cellcolor[RGB]{255,243,218}7.564±2.72 & 7.452±3.40       & 7.344±5.08 & 7.531±4.99 & 0.000±0.00  & \cellcolor[RGB]{255,228,173}7.574±5.30 \\
    & LE & 6.981±3.18       & 6.479±3.16 & 5.373±0.00       & 7.011±2.71       & \cellcolor[RGB]{255,243,218}0.000±2.53 & 6.630±3.44 & 6.927±3.24 & 0.000±0.00  & \cellcolor[RGB]{255,228,173}7.018±4.01 \\
    \midrule
    \multicolumn{2}{c|}{$\overline{AR}(ST)$}
    & 3.0(+) & 5.7(+) & 6.7(+) & 3.1(+) & 4.6(+) & 6.9(+) & 5.1(+) & 8.8(+) & 1.1 \\
    \bottomrule
    \end{tabular}
    }
\end{table*}

\begin{table*}[!t]
    \centering
    \renewcommand{\arraystretch}{1.2}
    \caption{Federated clustering performance evaluated by NMi and ACC. ``$\overline{AR}(ST)$'' row reports the average performance ranks and the significance test results, where a significant difference between the corresponding method and GOLD is indicated by ``+''.}
    \label{tbl:federated_performance_appendix}
    \scalebox{0.95}{
    \begin{tabular}{c|c|c c c c c c c c|c}
    \toprule
    \textbf{Index} & \textbf{Data} & \textbf{$k$-Fed} & \textbf{FFCM-avg1} & \textbf{FFCM-avg2} & \textbf{OSFSC-SSC} & \textbf{OSFSC-TSC} & \textbf{FedSC} & \textbf{NN-FC} & \textbf{AFCL} & \textbf{GOLD}\\
    \text{} & \text{} & (2021~\cite{dennis2021heterogeneity}) & (2022~\cite{stallmann2022towards}) & (2022~\cite{stallmann2022towards}) & (2023~\cite{xie2023fed}) & (2023~\cite{xie2023fed}) & (2024~\cite{qiao2024federated}) & (2024~\cite{wang2024one}) & (2025~\cite{zhang2025asynchronous}) & (ours)\\ 
    \midrule
    \multirow{10}{*}{NMI} 
    & EC & \cellcolor[RGB]{255,228,173}0.569±0.03 & 0.187±0.03 & 0.159±0.02 & 0.269±0.03       & 0.346±0.00       & 0.034±0.01 & 0.520±0.08 & 0.053±0.04 & \cellcolor[RGB]{255,243,218}0.565±0.07    \\
    & US & \cellcolor[RGB]{255,243,218}0.217±0.02 & 0.148±0.01 & 0.143±0.01 & 0.081±0.01       & 0.098±0.00       & 0.013±0.00 & 0.209±0.06 & 0.000±0.00 & \cellcolor[RGB]{255,228,173}0.301±0.13 \\
    & VE & \cellcolor[RGB]{255,243,218}0.131±0.02 & 0.078±0.00 & 0.079±0.00 & 0.056±0.00       & 0.117±0.00       & 0.090±0.01 & 0.105±0.03 & 0.000±0.00 & \cellcolor[RGB]{255,228,173}0.145±0.02 \\
    & EP & \cellcolor[RGB]{255,243,218}0.012±0.01 & 0.011±0.00 & \cellcolor[RGB]{255,228,173}0.014±0.00 & 0.009±0.01       & 0.011±0.00       & 0.007±0.00 & 0.009±0.00 & 0.000±0.00 & \cellcolor[RGB]{255,228,173}0.014±0.00 \\
    & YE & \cellcolor[RGB]{255,243,218}0.329±0.04 & 0.087±0.00 & 0.079±0.00 & 0.166±0.01       & 0.248±0.01       & 0.019±0.01 & 0.224±0.04 & 0.013±0.03 & \cellcolor[RGB]{255,228,173}0.337±0.05 \\
    & CA & \cellcolor[RGB]{255,243,218}0.372±0.02 & 0.169±0.00 & 0.144±0.00 & 0.266±0.02       & 0.330±0.01       & 0.292±0.04 & 0.260±0.03 & 0.000±0.00 & \cellcolor[RGB]{255,228,173}0.390±0.03 \\
    & LA & 0.512±0.02 & 0.394±0.00 & 0.252±0.00 & 0.358±0.00       & \cellcolor[RGB]{255,243,218}0.568±0.00 & 0.565±0.03 & 0.557±0.04 & 0.052±0.16 & \cellcolor[RGB]{255,228,173}0.576±0.03 \\
    & WI & 0.070±0.00 & 0.067±0.00 & 0.030±0.00 & \cellcolor[RGB]{255,243,218}0.074±0.01 & 0.071±0.00       & 0.012±0.02 & 0.028±0.02 & 0.000±0.00 & \cellcolor[RGB]{255,228,173}0.079±0.02 \\
    & PE & \cellcolor[RGB]{255,243,218}0.645±0.00 & 0.276±0.00 & 0.269±0.01 & 0.610±0.01       & 0.598±0.01       & 0.517±0.03 & 0.635±0.02 & 0.000±0.00 & \cellcolor[RGB]{255,228,173}0.662±0.03 \\
    & LE & \cellcolor[RGB]{255,228,173}0.428±0.01 & 0.139±0.01 & 0.017±0.00 & 0.318±0.01       & \cellcolor[RGB]{255,243,218}0.410±0.00 & 0.316±0.02 & 0.387±0.02 & 0.000±0.00 & \cellcolor[RGB]{255,228,173}0.428±0.02 \\
    \midrule
    \multicolumn{2}{c|}{$\overline{AR}(ST)$}
    & 2.4(+) & 5.9(+) & 6.4(+) & 5.4(+) & 3.7(+) & 6.5(+) & 4.5(+) & 8.9(+) & 1.2 \\
    \midrule
    \multirow{10}{*}{ACC} 
    & EC & 0.535±0.02       & 0.359±0.04 & 0.370±0.01       & 0.408±0.04 & 0.510±0.01 & 0.332±0.00       & \cellcolor[RGB]{255,243,218}0.612±0.08    & 0.348±0.01          & \cellcolor[RGB]{255,228,173}0.644±0.05 \\
    & US & 0.447±0.00       & 0.456±0.00 & \cellcolor[RGB]{255,243,218}0.457±0.00 & 0.364±0.02 & 0.438±0.00 & 0.341±0.00       & 0.433±0.05          & 0.337±0.00          & \cellcolor[RGB]{255,228,173}0.488±0.07 \\
    & VE & 0.368±0.01       & 0.322±0.01 & 0.320±0.00       & 0.355±0.02 & \cellcolor[RGB]{255,228,173}0.382±0.00 & 0.345±0.01       & 0.363±0.02          & 0.298±0.00          & \cellcolor[RGB]{255,243,218}0.379±0.03    \\
    & EP & 0.330±0.02       & 0.331±0.01 & 0.331±0.01       & 0.314±0.02 & 0.295±0.01 & 0.314±0.01       & 0.311±0.02          & \cellcolor[RGB]{255,228,173}0.341±0.00 & \cellcolor[RGB]{255,243,218} 0.334±0.01    \\
    & YE & \cellcolor[RGB]{255,243,218}0.432±0.06 & 0.291±0.00 & 0.287±0.00       & 0.315±0.01 & 0.384±0.01 & 0.278±0.00       & 0.342±0.04          & 0.281±0.01          & \cellcolor[RGB]{255,228,173}0.442±0.07 \\
    & CA & \cellcolor[RGB]{255,243,218}0.452±0.03 & 0.327±0.00 & 0.351±0.00       & 0.313±0.02 & 0.418±0.01 & 0.438±0.05       & 0.397±0.02          & 0.289±0.00          & \cellcolor[RGB]{255,228,173}0.473±0.03 \\
    & LA & 0.534±0.03       & 0.595±0.00 & 0.475±0.00       & 0.552±0.01 & \cellcolor[RGB]{255,243,218}0.689±0.00    & 0.622±0.01       & 0.646±0.06          & 0.283±0.11          & \cellcolor[RGB]{255,228,173}0.727±0.04 \\
    & WI & 0.323±0.01       & 0.360±0.00 & 0.391±0.01       & 0.384±0.01 & 0.270±0.00          & \cellcolor[RGB]{255,243,218}0.442±0.04 & 0.385±0.03          & \cellcolor[RGB]{255,228,173}0.464±0.00 & 0.345±0.03          \\
    & PE & 0.593±0.00       & 0.325±0.00 & 0.311±0.00       & 0.592±0.03 & 0.604±0.02          & 0.515±0.04       & \cellcolor[RGB]{255,228,173}0.649±0.04 & 0.162±0.00          & \cellcolor[RGB]{255,243,218}0.622±0.07    \\
    & LE & 0.288±0.02       & 0.141±0.00 & 0.100±0.00       & 0.249±0.01 & 0.235±0.01          & 0.238±0.02       & \cellcolor[RGB]{255,243,218}0.290±0.01    & 0.089±0.00          & \cellcolor[RGB]{255,228,173}0.292±0.02 \\
    \midrule
    \multicolumn{2}{c|}{$\overline{AR}$}
    & 4.1(+) & 5.8(+) & 6.0(+) & 5.7(+) & 4.6(+) & 5.8(+) & 3.9(+) & 7.2(+) & 1.9 \\
    \bottomrule
    \end{tabular}
    }
\end{table*}

\section{Complementary Evaluation Results}
This section presents the detailed experimental settings and complete experimental results, providing a comprehensive validation of the proposed method and its effectiveness.

\subsection{Clustering Performance Comparison}
\label{ap:performance}
To experimentally evaluate the performance of the proposed GOLD, it is compared with state-of-the-art counterparts on ten real-world datasets. The compared methods are executed 10 times on each dataset, and the average results are reported. The best and second-best results are indicated in \colorbox[RGB]{255,228,173}{dark orange} and \colorbox[RGB]{255,243,218}{light orange}, respectively. The ``$\overline{AR}(ST)$'' row summarizes the average rank of each method across all datasets, and the symbol ``(+)'' denotes a significant difference based on the Wilcoxon signed-rank test at the 95\% confidence level.

To evaluate the interpretability and representation quality of the clusters learned by FC, the experiment is designed to inherit the output centroids learned by FC methods and launch $k$-means on the complete raw datasets. 
To ensure that the performance is closely tied to the capabilities of the compared FC methods, $k$-means is restricted to a single iteration, serving solely to assign data objects to their nearest centroid. Such a design enables a direct evaluation of how effectively the learned prototypes capture the underlying global structure. 
Table~\ref{tbl:centralized_performance_appendix_external} reports two external indices, i.e., NMI and ACC, and Table~\ref{tbl:centralized_performance_appendix_internal} reports two internal indices, i.e., SC and CH. According to the results, some key observations are provided below: 1) GOLD consistently attains the highest performance, demonstrating superior representation learning capability for exploring the global cluster distribution in FC tasks. GOLD ranks first majority of datasets, while it achieves the second-best results on the LA dataset for NMI, on the VE dataset for ACC, on the US and CA datasets for SC, and on the EP dataset for CH. However, the gap between it and the winner is marginal, still demonstrating its robustness across diverse datasets.
2) The comparative methods have only demonstrated satisfactory performance on specific datasets, without showing consistently competitive results across different scenarios. This reveals their vulnerability when dealing with Non-ICD data composed of incomplete and multi-granular subcluster distributions.
3) Average-rank and significance tests further corroborate the above findings, confirming that the improvements achieved by GOLD are statistically significant and robust.

To validate the consistency between the cluster assignments obtained by the FC and the ground-truth labels across clients, the overall federated clustering performance is measured by treating datasets from all clients as a whole. 
% The clustering performance of GOLD is compared with existing FC approaches under the Non-ICD scenario. 
Table~\ref{tbl:federated_performance_appendix} reports the NMI and ARI performance. Key observations include: 1) GOLD consistently outperforms its counterparts, highlighting its superiority in FC accuracy. Specifically, GOLD achieves the best on nearly all datasets w.r.t. the NMI index, except for the EC dataset, where it still ranks second. For the cases where GOLD does not perform the best, i.e., the ACC index on the VE, EP, WI, and PE, the gap between GOLD and the best-performing counterpart is tiny, demonstrating its robustness and effectiveness across diverse scenarios.
2) For datasets with small $\lambda$, e.g., the EP dataset ($\lambda=0.02$), the performance gap between GOLD and the other methods is narrowed. This indicates that $\lambda$ serves as a reliable measure for quantifying the Non-ICD degree of a dataset. This also illustrates the robustness and competitiveness of GOLD even when the Non-ICD effect is extremely mild, thanks to its thorough multi-granularity information representation mechanism.
3) The significance tests confirm that GOLD demonstrates statistically significant performance over its counterparts.

\begin{figure}
    \centering
    \includegraphics[width=0.6\linewidth]{Figures/Non_ICD/legend_non_icd.pdf} \\[1em]
    \includegraphics[width=0.49\linewidth]{Figures/Non_ICD/EC_Purity_icd.pdf}
    \includegraphics[width=0.49\linewidth]{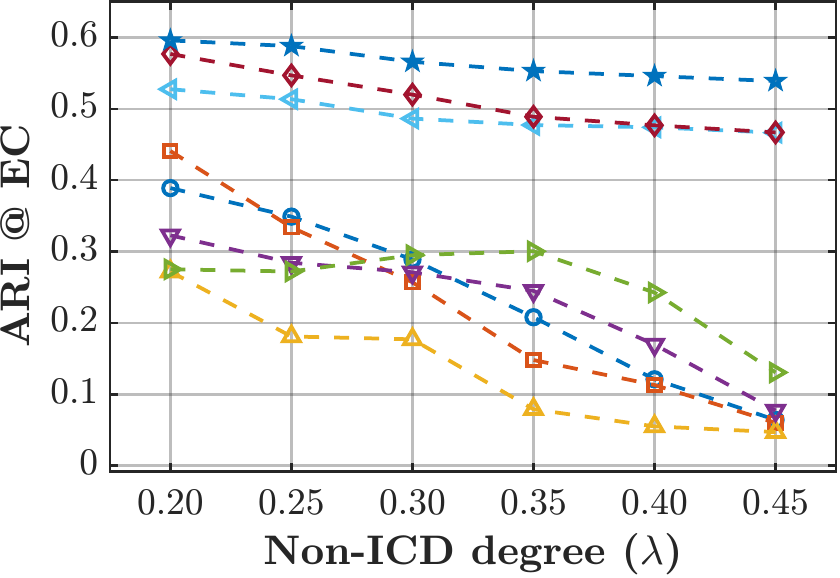}
    \includegraphics[width=0.49\linewidth]{Figures/Non_ICD/US_Purity_icd.pdf}
    \includegraphics[width=0.49\linewidth]{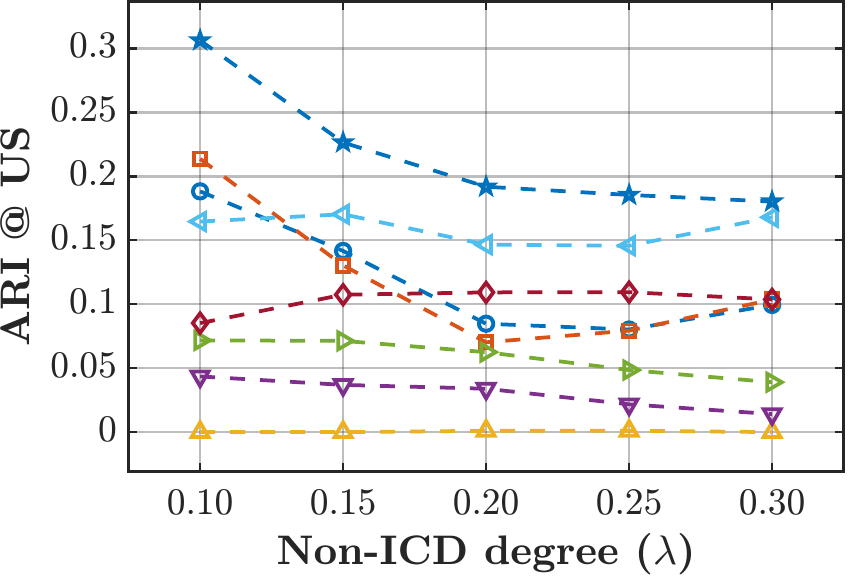}
    \includegraphics[width=0.49\linewidth]{Figures/Non_ICD/VE_Purity_icd.pdf}
    \includegraphics[width=0.49\linewidth]{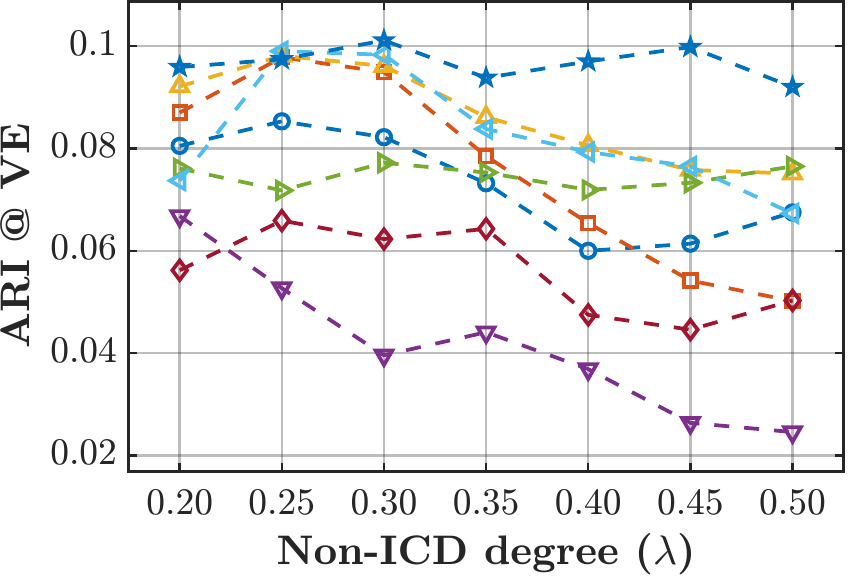}
    \includegraphics[width=0.49\linewidth]{Figures/Non_ICD/YE_Purity_icd.pdf}
    \includegraphics[width=0.49\linewidth]{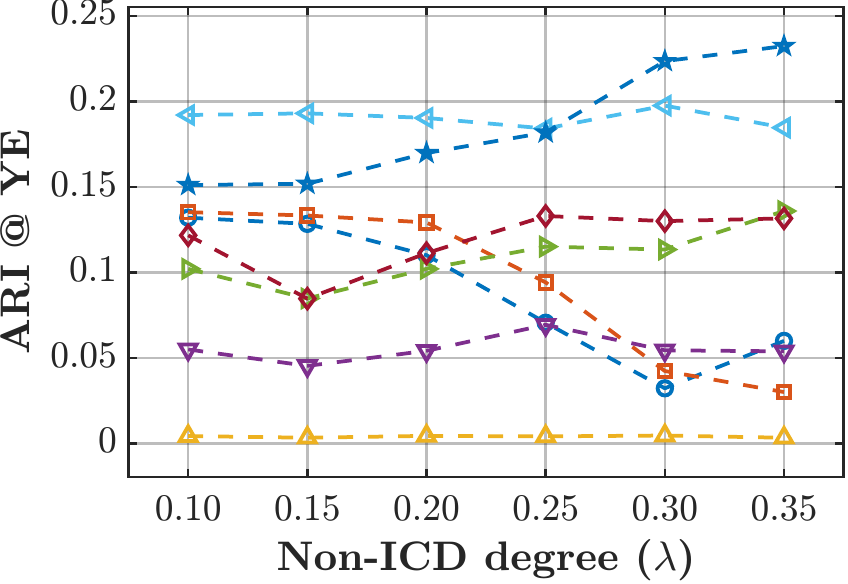}
    \includegraphics[width=0.49\linewidth]{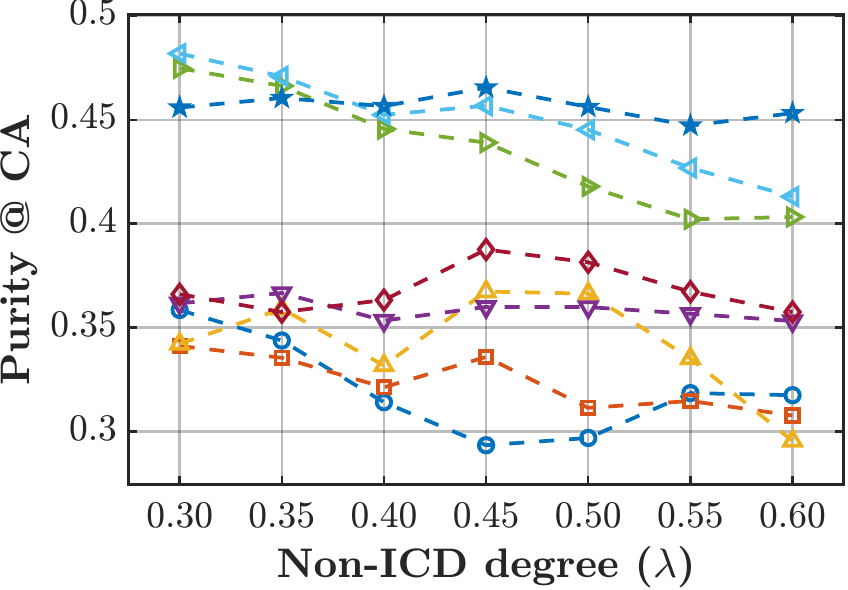}
    \includegraphics[width=0.49\linewidth]{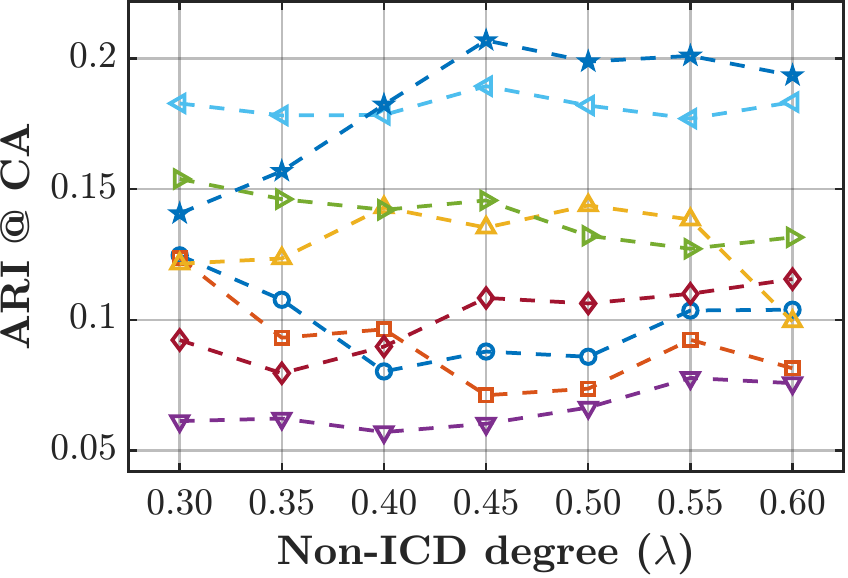}
    \caption{Federated clustering performance comparison under different Non-ICD degrees ($\lambda$). A smaller $\lambda$ indicates lower data heterogeneity across clients.}
    \label{fig:non-icd}
\end{figure}

\subsection{Evaluation of the Impact of Non-ICD}
\label{ap:non_icd}

To evaluate the impact of federated clustering by different levels of data heterogeneity, GOLD is compared with its counterparts by varying the degree of Non-ICD, i.e., $\lambda$. As shown in Fig.~\ref{fig:non-icd}, the clustering performance is evaluated using the Purity and ARI index on five datasets. It can be seen that on datasets such as EC, US, and VE, baseline methods suffer significant performance drops as $\lambda$ increases, whereas GOLD exhibits remarkable robustness. On dataset YE and CA, where the overall Non-ICD impact is milder, GOLD still outperforms others, suggesting its superior adaptability in FC. In short, GOLD performs well on extremely Non-ICD data and demonstrates its superior adaptability to different degrees of Non-ICD.

\subsection{Scalability Evaluation w.r.t. Performance}
\label{ap:scalability}
\begin{figure*}
    \centering
    \includegraphics[width=0.4\linewidth]{Figures/Ablation_of_Client/legend.pdf} \\[1em]
    % EC
    \includegraphics[width=0.24\linewidth]{Figures/Ablation_of_Client/EC_Purity.pdf}
    \includegraphics[width=0.24\linewidth]{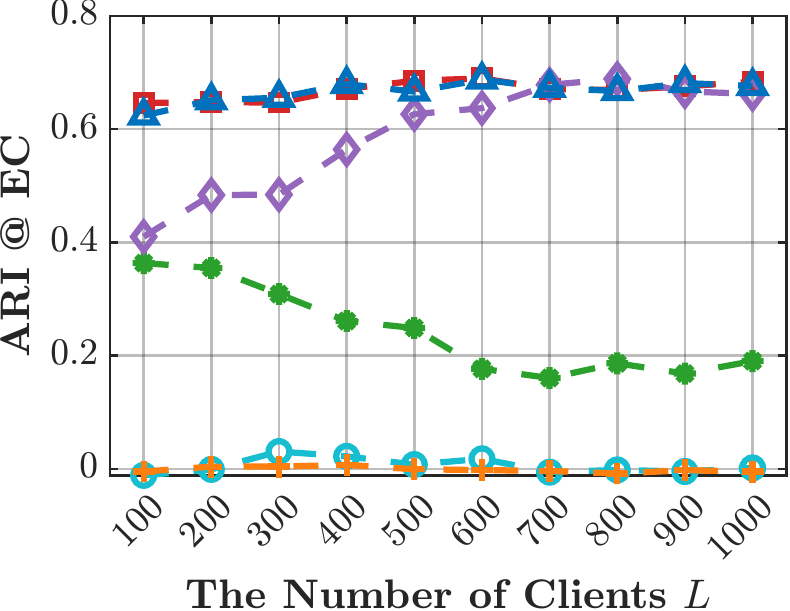}
    \includegraphics[width=0.24\linewidth]{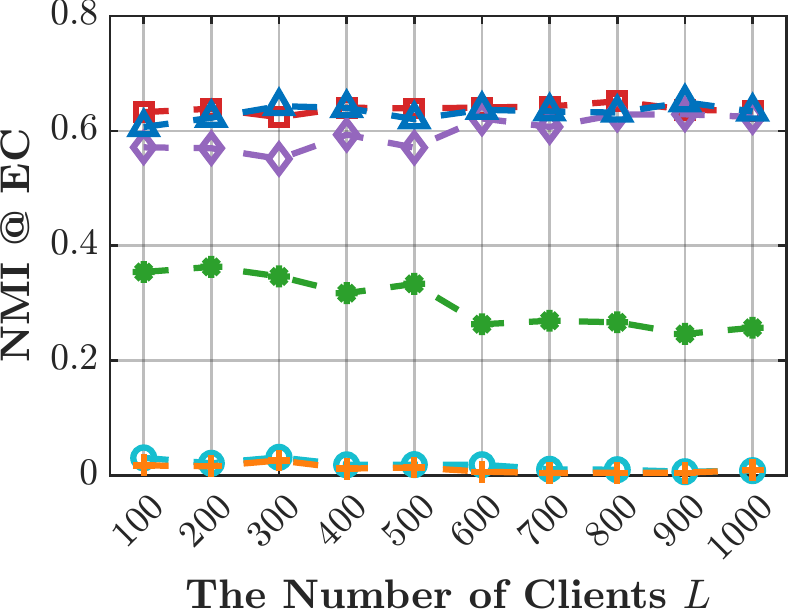}
    \includegraphics[width=0.24\linewidth]{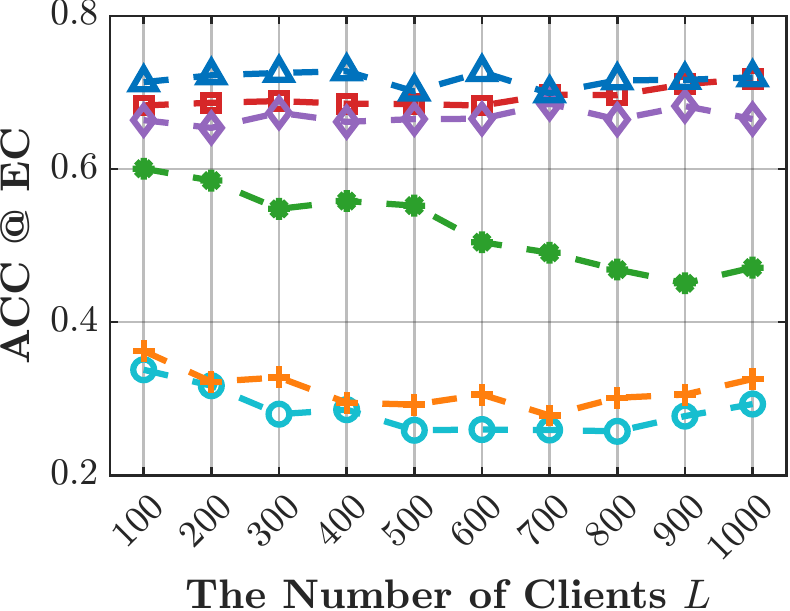}
    % US
    \includegraphics[width=0.24\linewidth]{Figures/Ablation_of_Client/US_Purity.pdf}
    \includegraphics[width=0.24\linewidth]{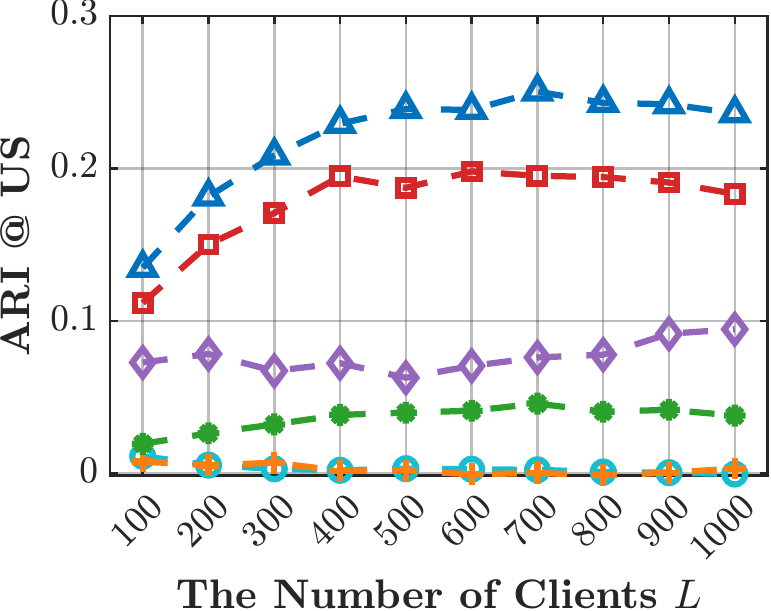}
    \includegraphics[width=0.24\linewidth]{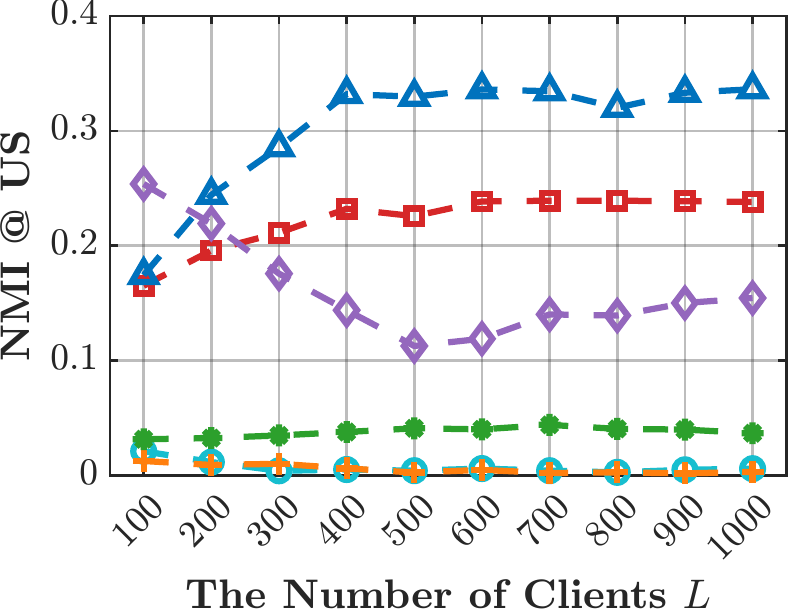}
    \includegraphics[width=0.24\linewidth]{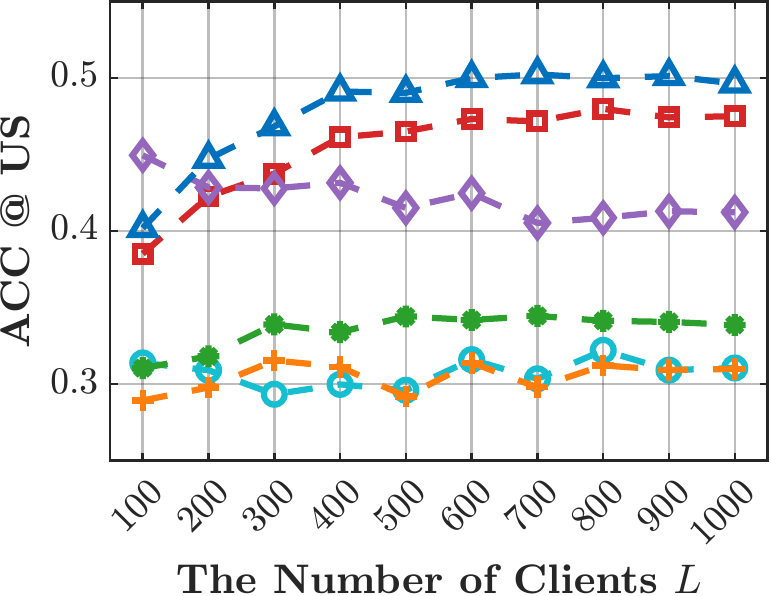}
    % VE
    \includegraphics[width=0.24\linewidth]{Figures/Ablation_of_Client/VE_Purity.pdf}
    \includegraphics[width=0.24\linewidth]{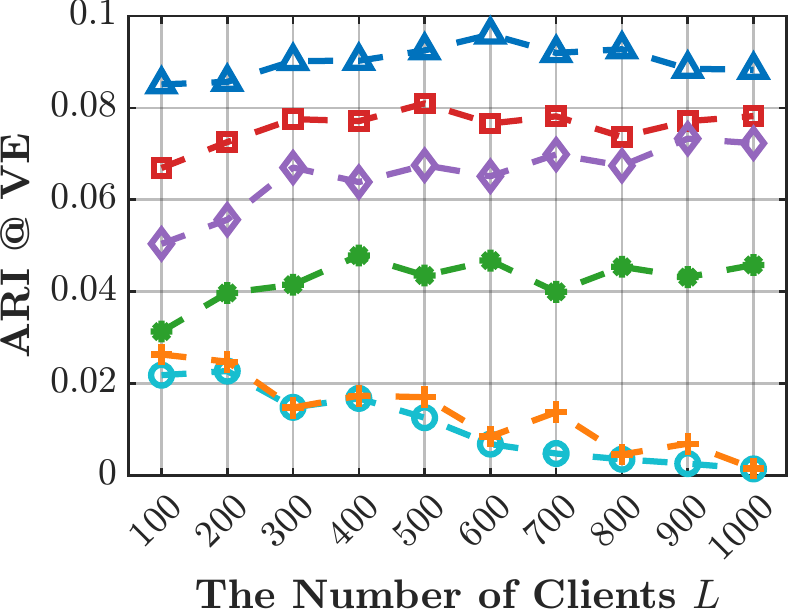}
    \includegraphics[width=0.24\linewidth]{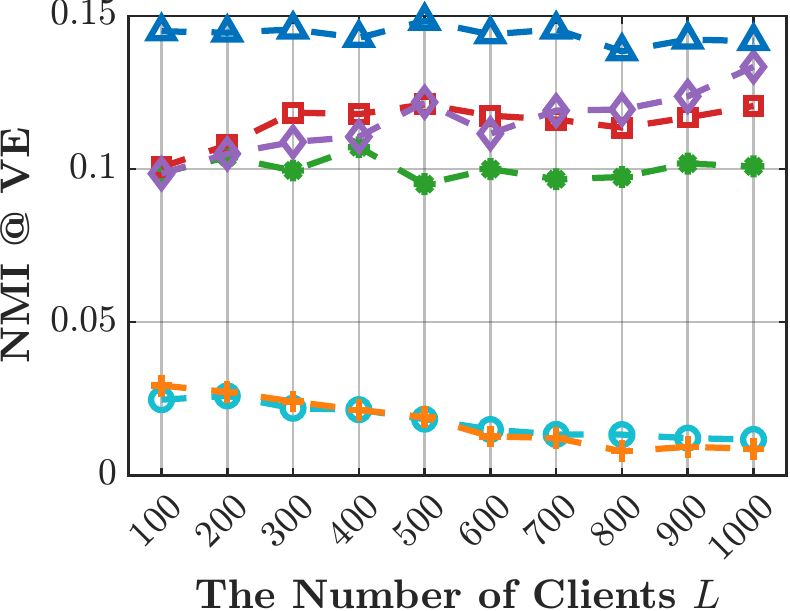}
    \includegraphics[width=0.24\linewidth]{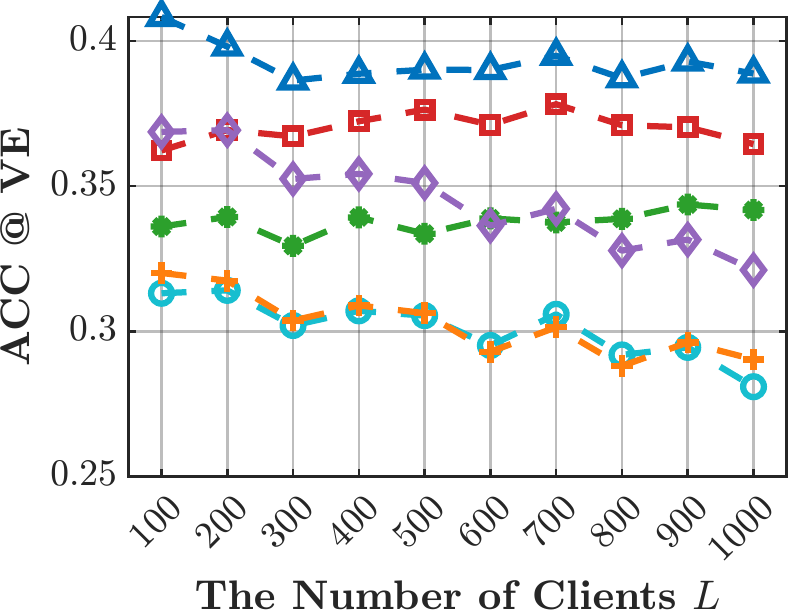}
    % EP
    \includegraphics[width=0.24\linewidth]{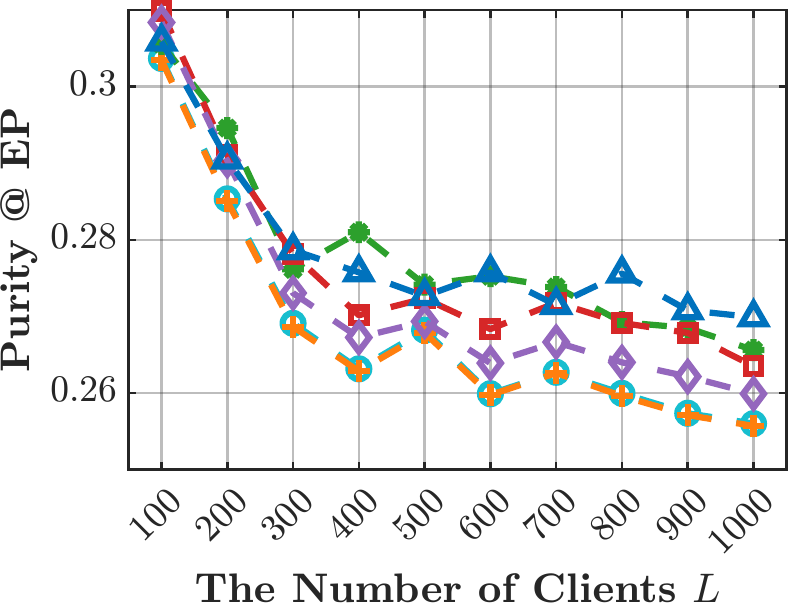}
    \includegraphics[width=0.24\linewidth]{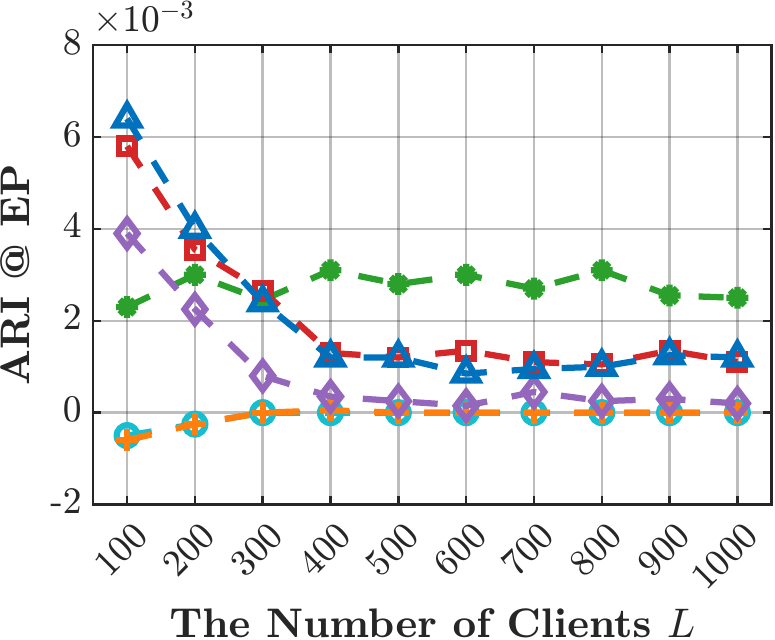}
    \includegraphics[width=0.24\linewidth]{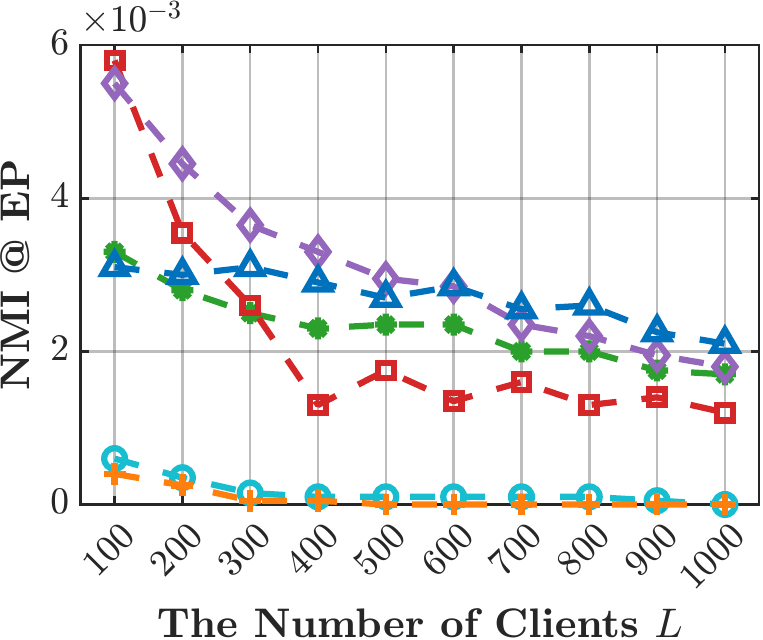}
    \includegraphics[width=0.24\linewidth]{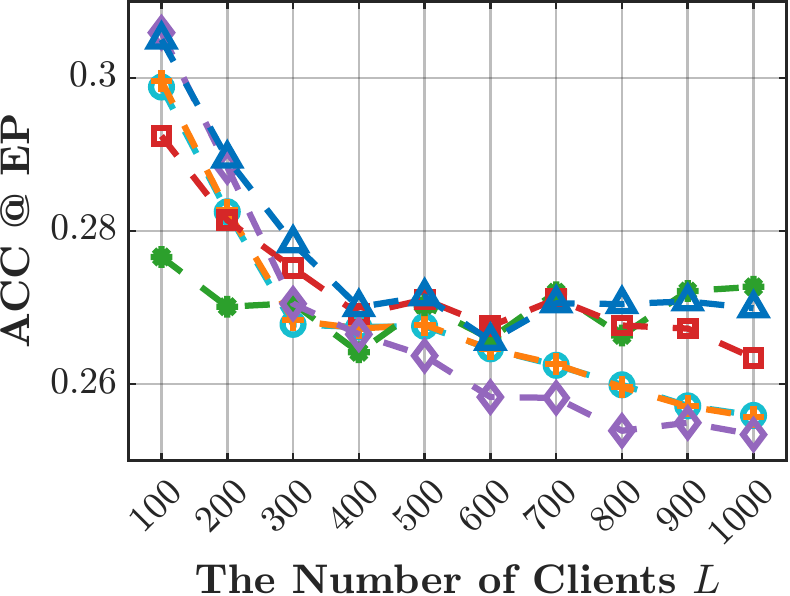}
    % YE
    \includegraphics[width=0.24\linewidth]{Figures/Ablation_of_Client/YE_Purity.pdf}
    \includegraphics[width=0.24\linewidth]{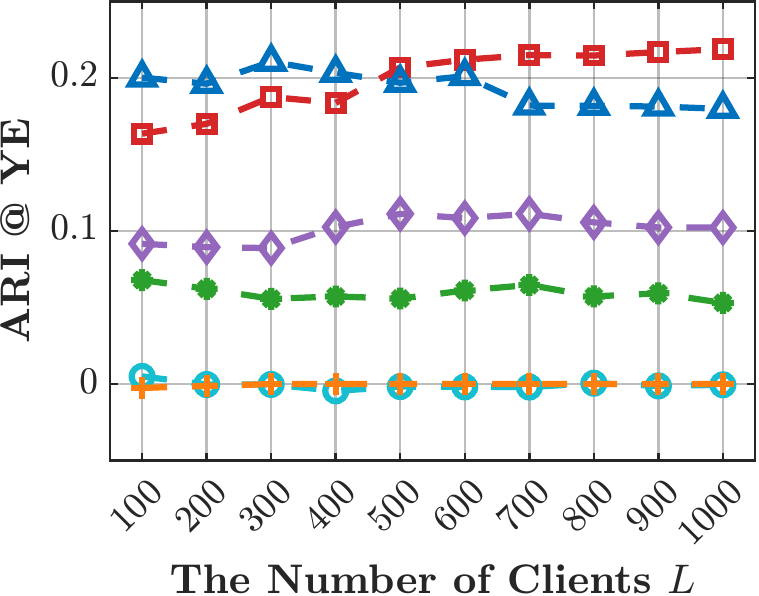}
    \includegraphics[width=0.24\linewidth]{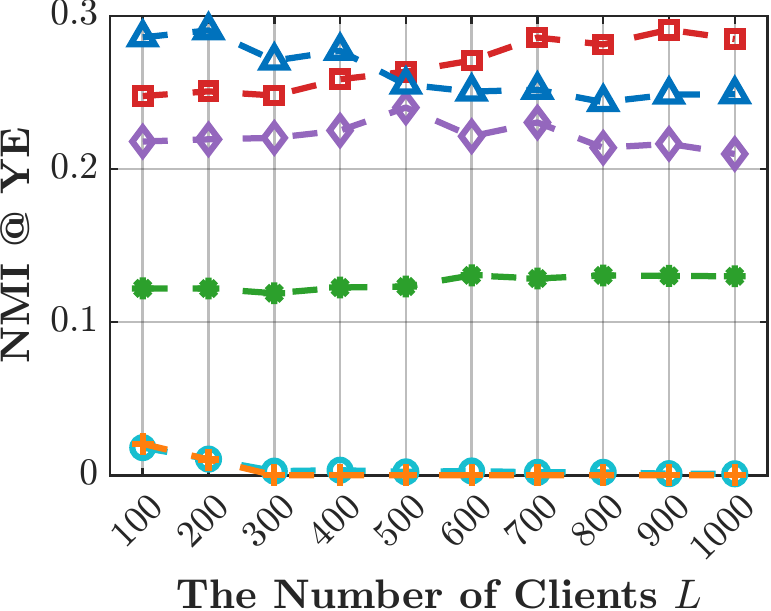}
    \includegraphics[width=0.24\linewidth]{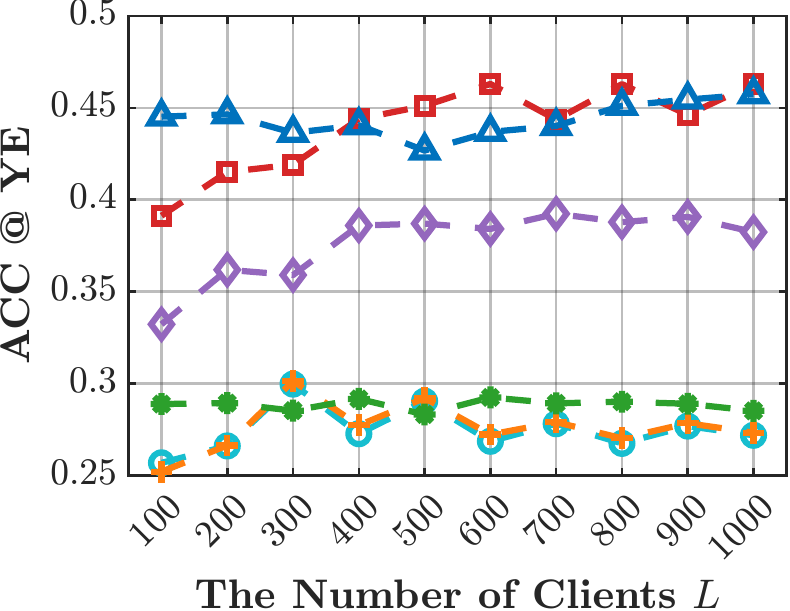}
    % CA
    \includegraphics[width=0.24\linewidth]{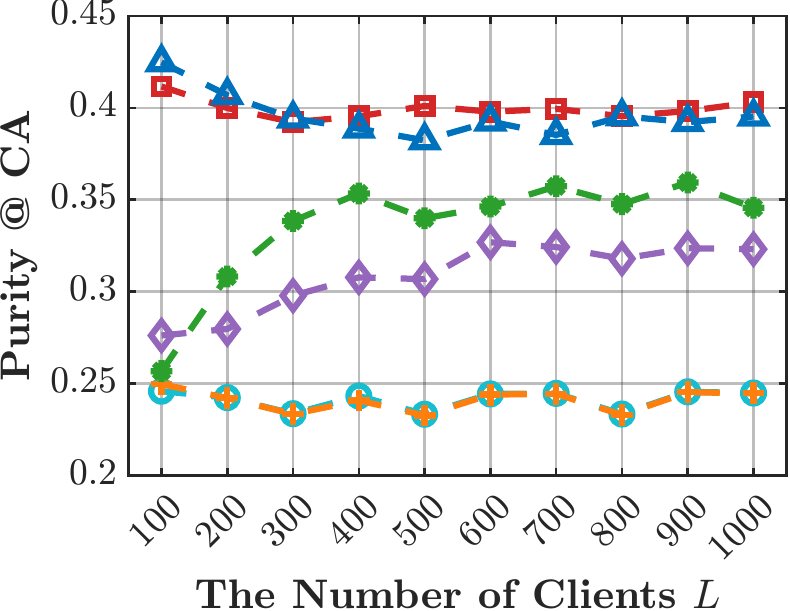}
    \includegraphics[width=0.24\linewidth]{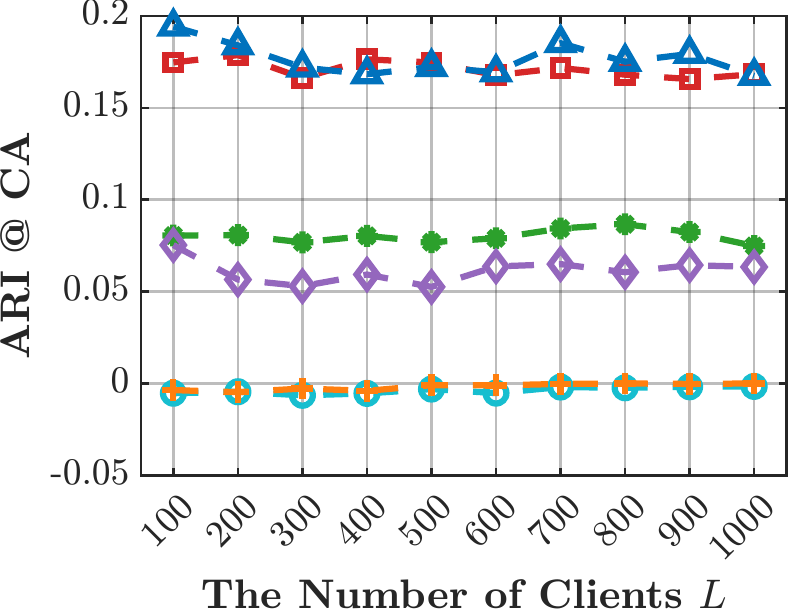}
    \includegraphics[width=0.24\linewidth]{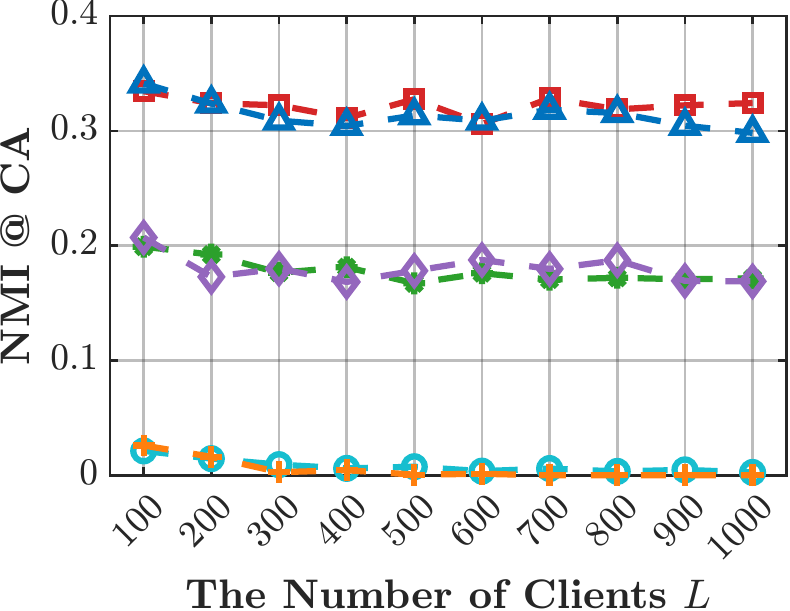}
    \includegraphics[width=0.24\linewidth]{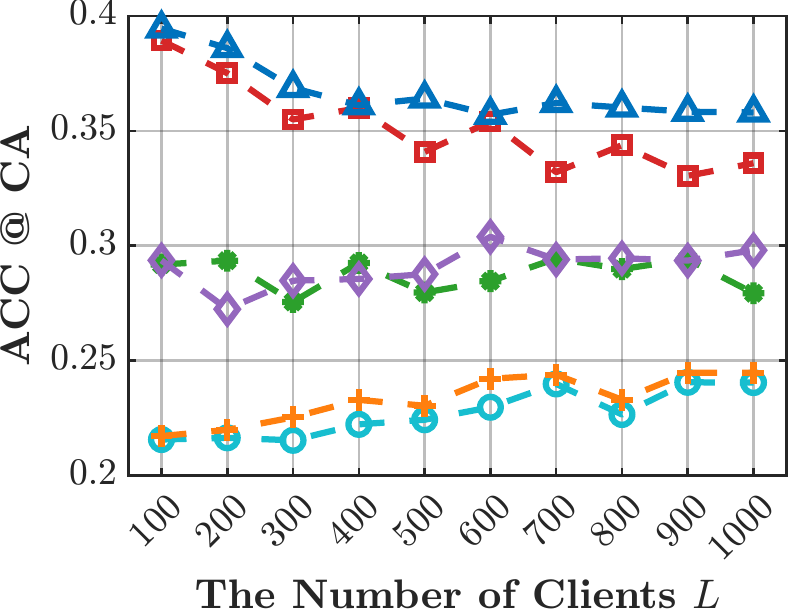}
    \caption{Client-number ablation study comparing GOLD with its counterparts.}
    \label{fig:ablation_of_cleint_appendix}
\end{figure*}

To systematically assess the scalability of GOLD under cross-client scenarios, its performance is compared with state-of-the-art counterparts by setting the number of clients $L$ to values ranging from $100$ to $1000$ (with a step size of $100$). As shown in Fig. \ref{fig:ablation_of_cleint_appendix}, across six datasets and four validity indices, GOLD traces the upper envelope of the curves and exhibits notably smoother trajectories as $L$ increases. While a few competitors occasionally approach GOLD at small or moderate scales on specific datasets, their performance tends to fluctuate as the federation grows and data heterogeneity intensifies, whereas GOLD remains stable and consistently higher. This behavior is particularly evident in settings where client expansion amplifies statistical shift, conditions under which the proposed multi-granular competitive penalization and prototype consolidation mitigate over-fragmentation and suppress noise accumulation during aggregation. Even on the most challenging datasets, where all methods experience degradation with larger $L$, GOLD shows a slower decline and smaller variance, reflecting better resilience to increasing federation size. Overall, these observations indicate that GOLD not only achieves strong average accuracy but also maintains robustness to scale, delivering superior and reliable clustering quality in federated environments.

\subsection{Hyper-Parameter Evaluation}
\label{ap:parameter}
\begin{figure*}
    \centering
    % EC
    \includegraphics[width=0.24\linewidth]{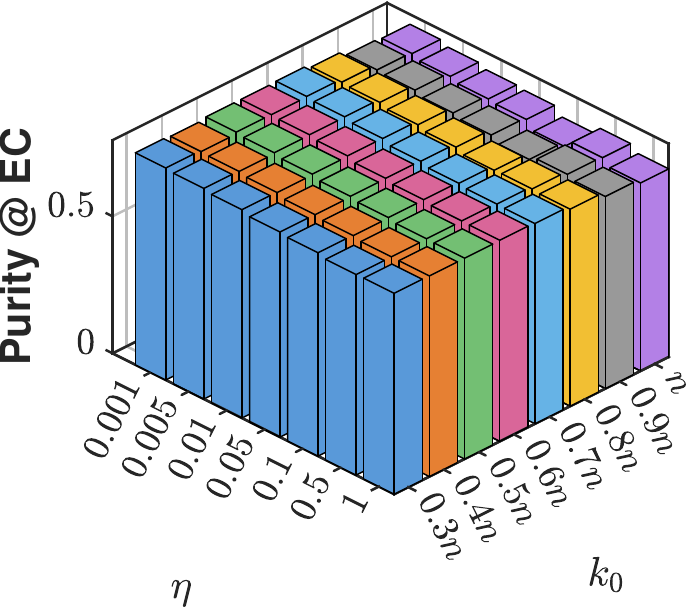}
    \includegraphics[width=0.24\linewidth]{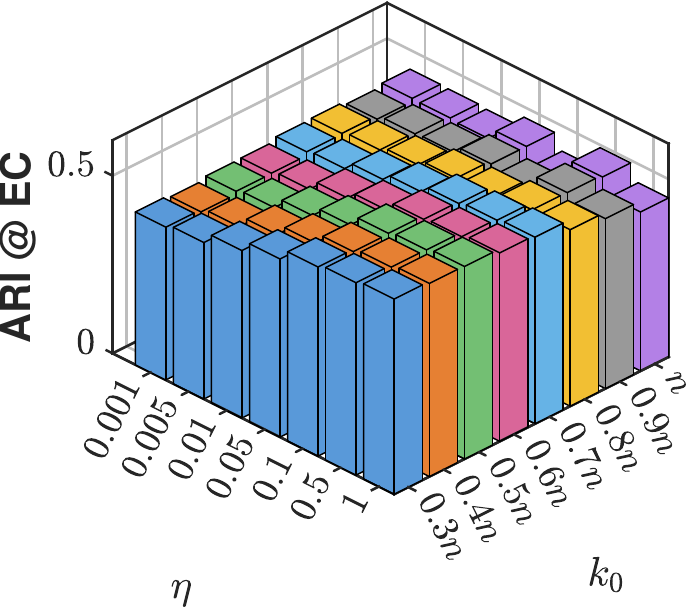}
    \includegraphics[width=0.24\linewidth]{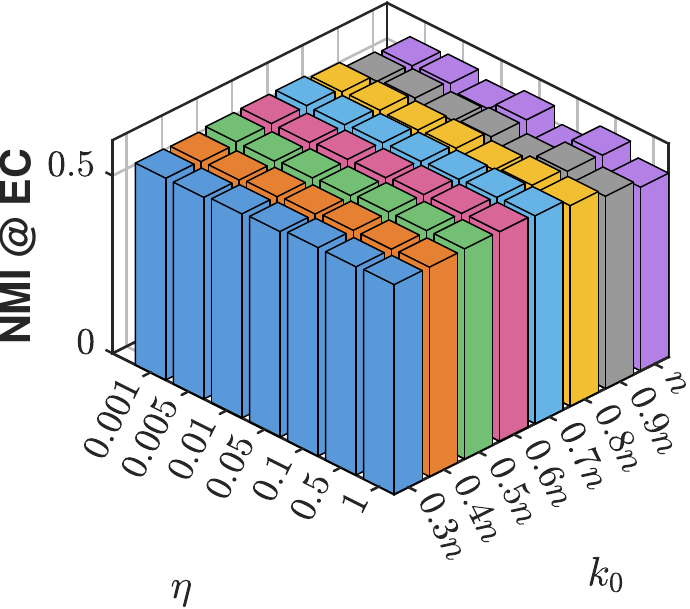}
    \includegraphics[width=0.24\linewidth]{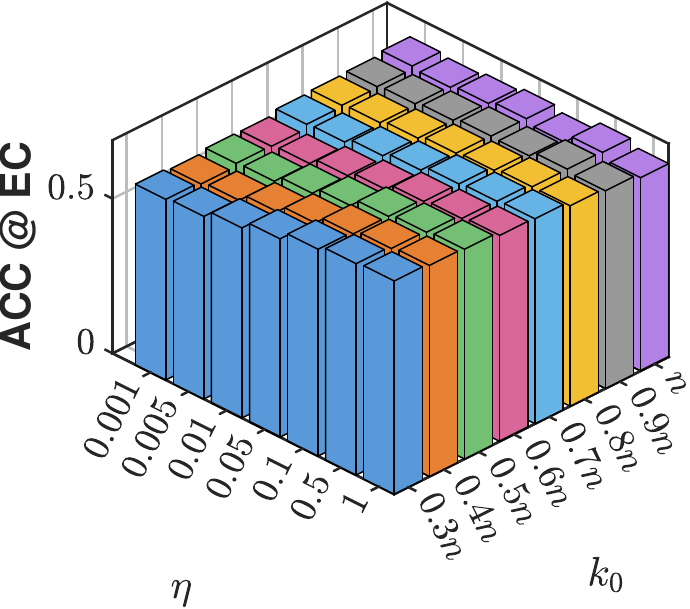}
    % US
    \includegraphics[width=0.24\linewidth]{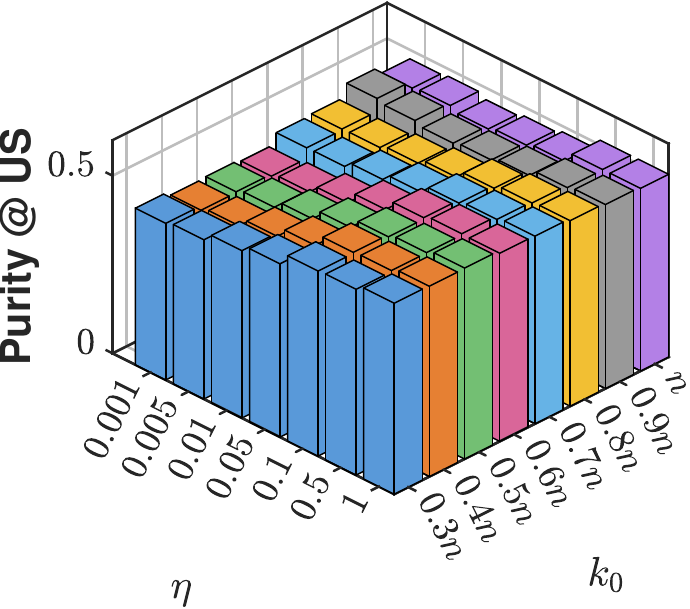}
    \includegraphics[width=0.24\linewidth]{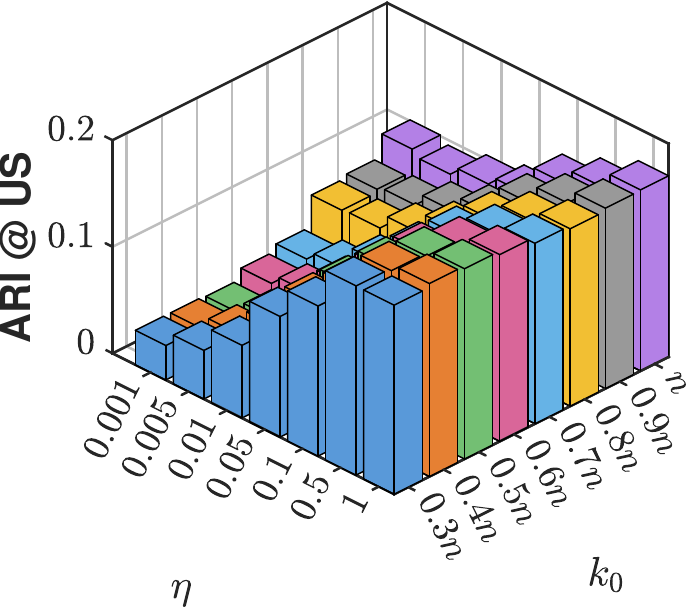}
    \includegraphics[width=0.24\linewidth]{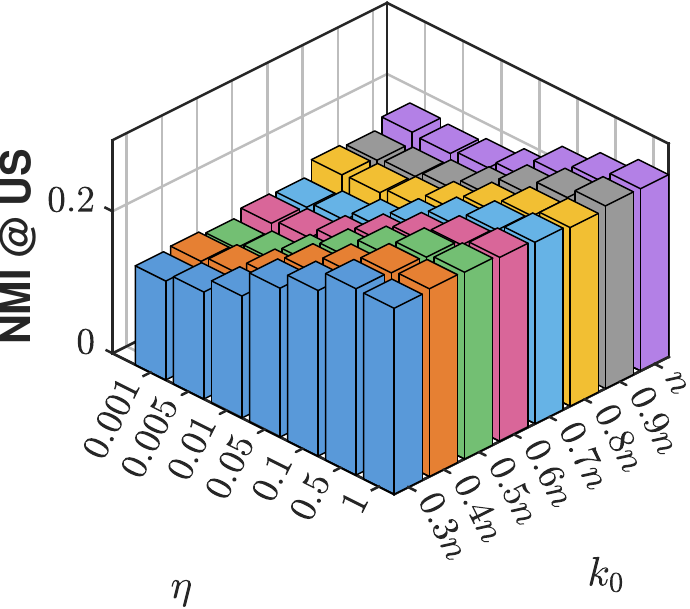}
    \includegraphics[width=0.24\linewidth]{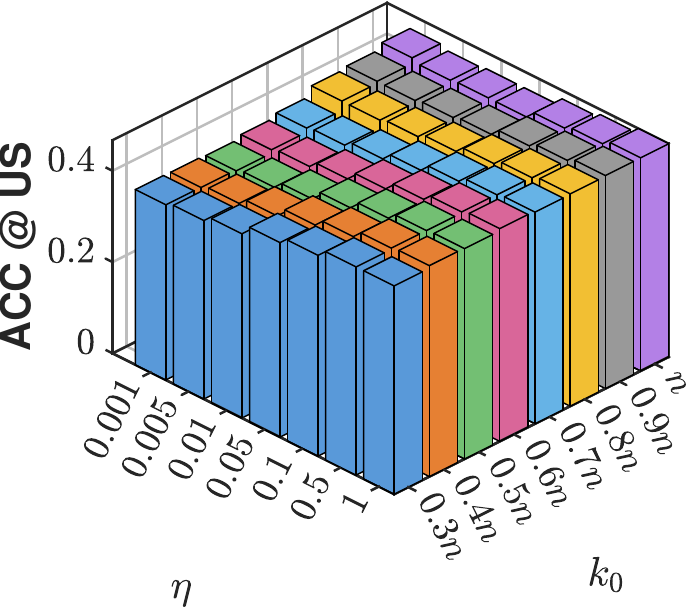}
    % VE
    \includegraphics[width=0.24\linewidth]{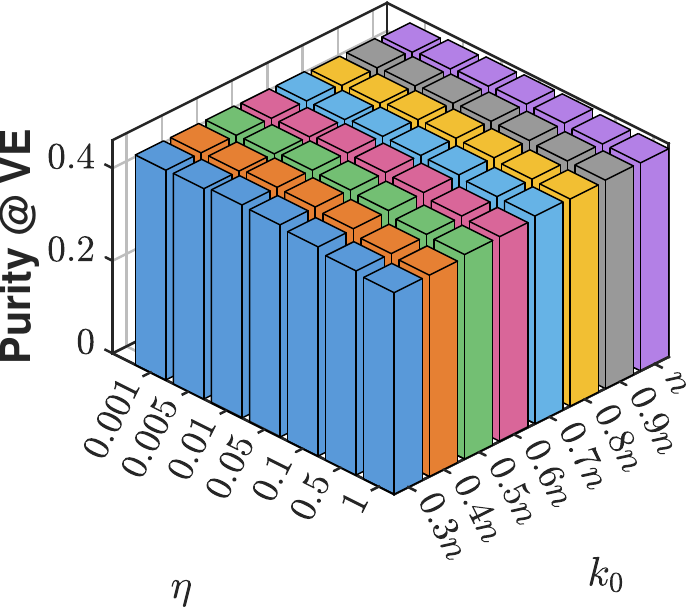}
    \includegraphics[width=0.24\linewidth]{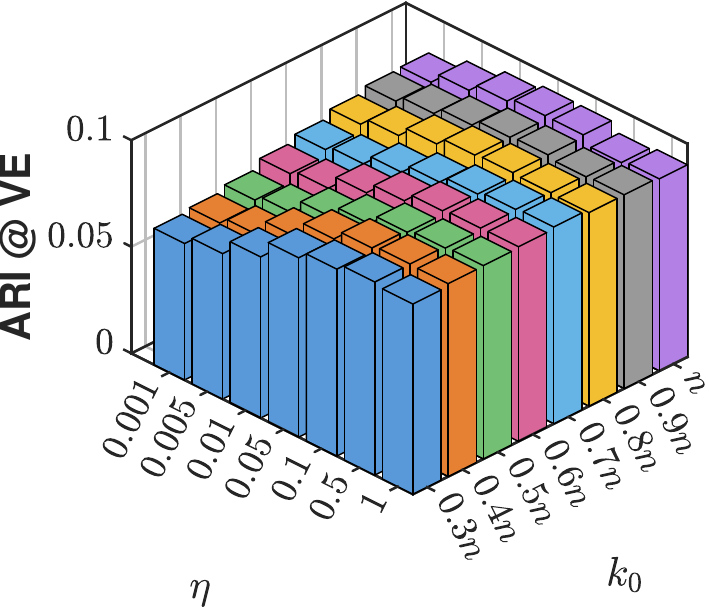}
    \includegraphics[width=0.24\linewidth]{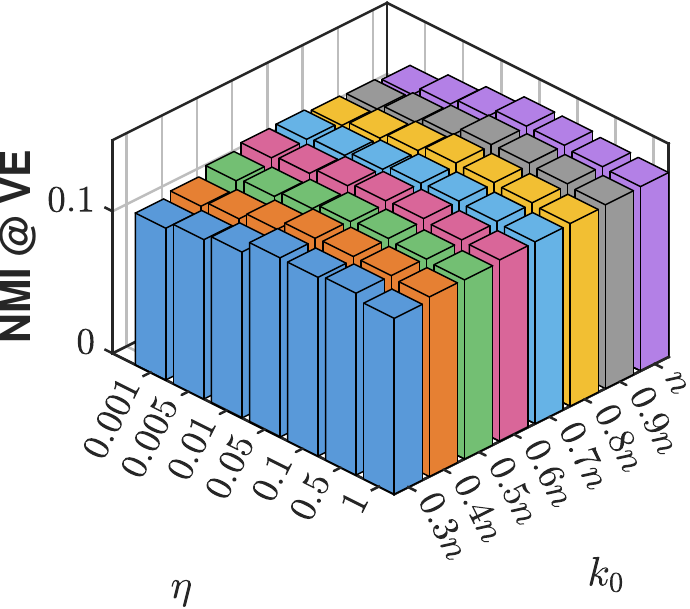}
    \includegraphics[width=0.24\linewidth]{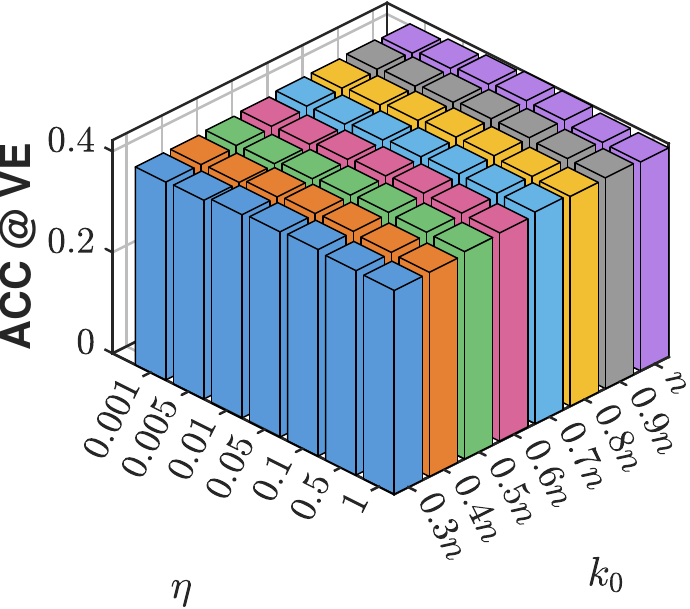}
    % EP
    \includegraphics[width=0.24\linewidth]{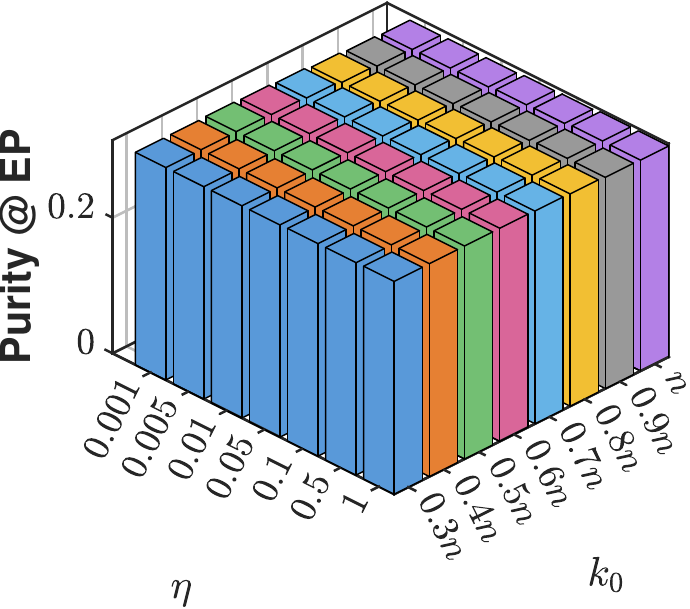}
    \includegraphics[width=0.24\linewidth]{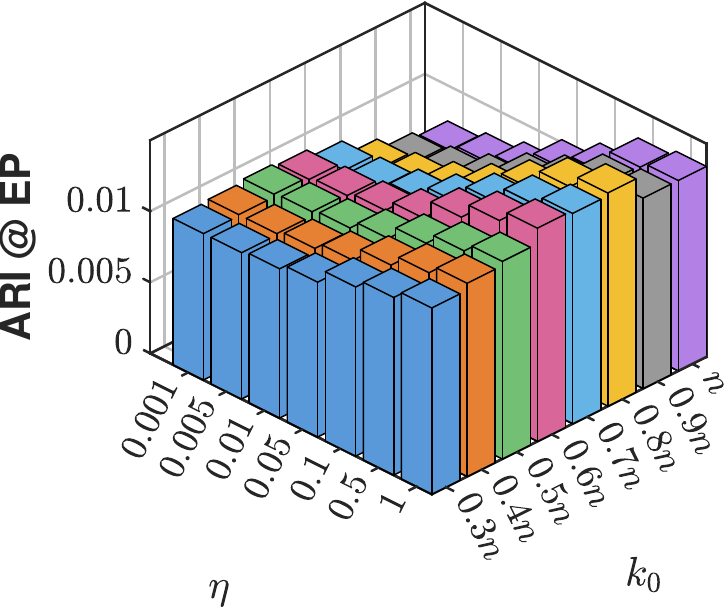}
    \includegraphics[width=0.24\linewidth]{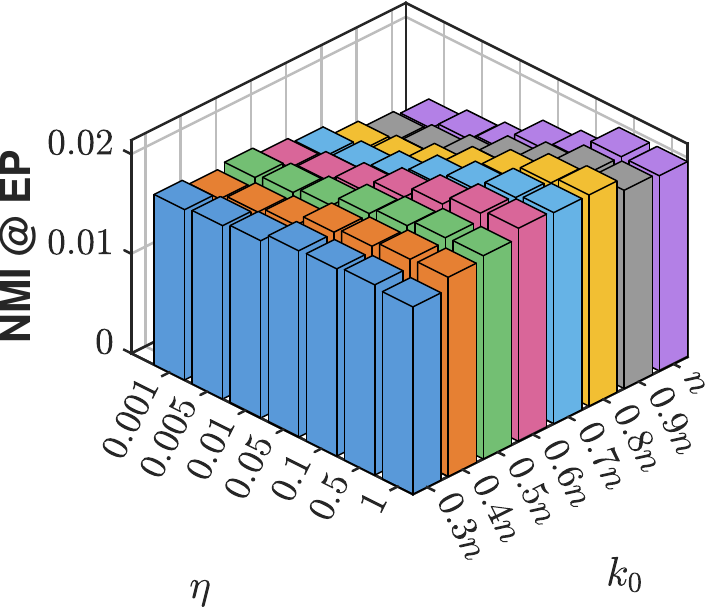}
    \includegraphics[width=0.24\linewidth]{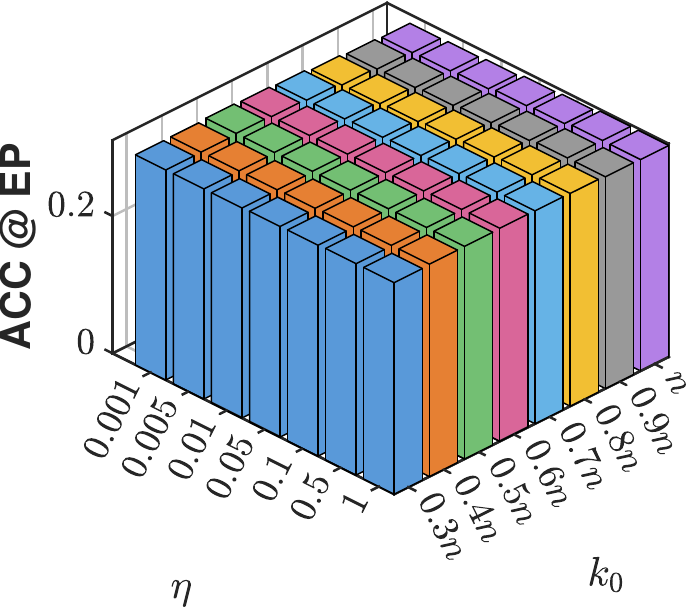}
    % YE
    \includegraphics[width=0.24\linewidth]{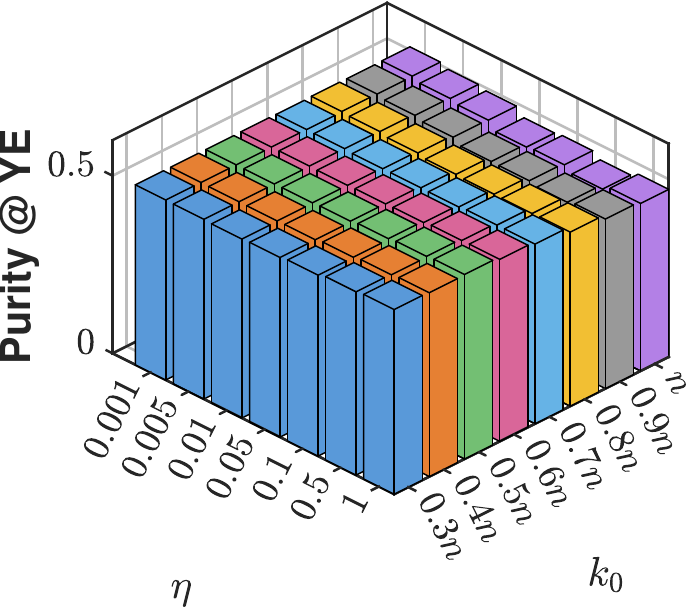}
    \includegraphics[width=0.24\linewidth]{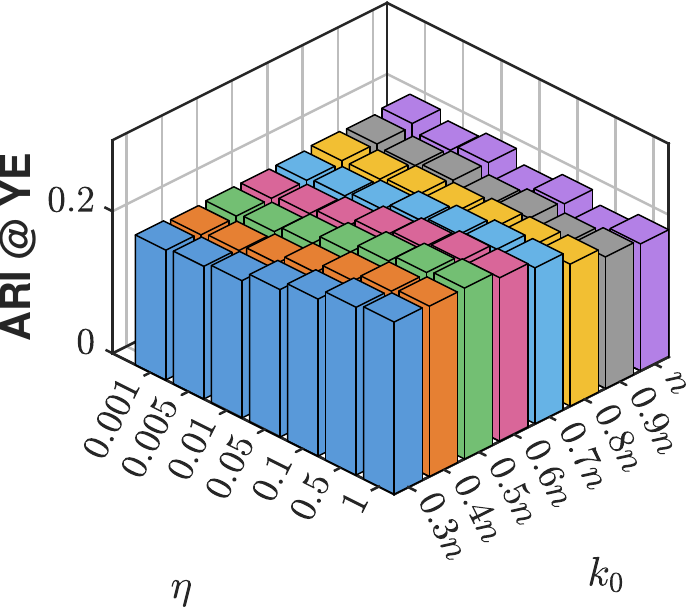}
    \includegraphics[width=0.24\linewidth]{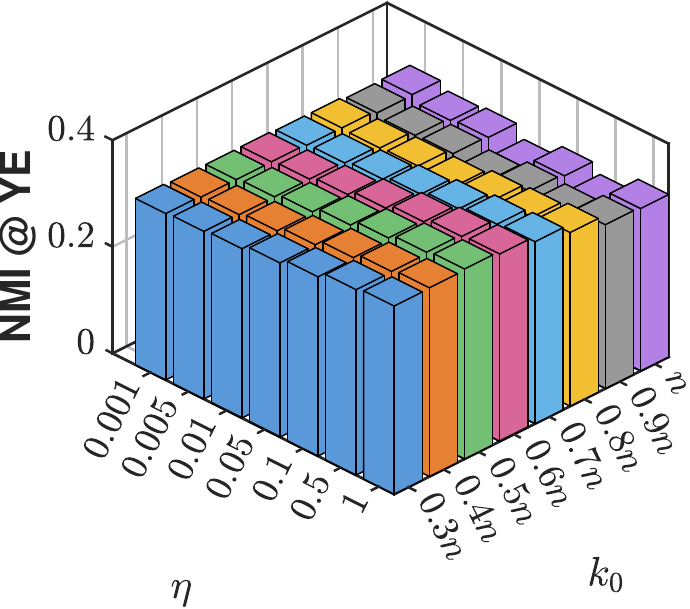}
    \includegraphics[width=0.24\linewidth]{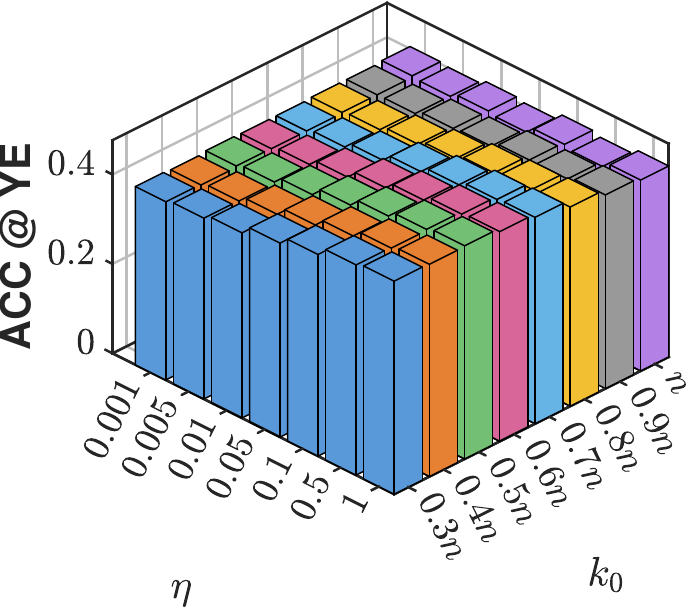}
    % CA
    \includegraphics[width=0.24\linewidth]{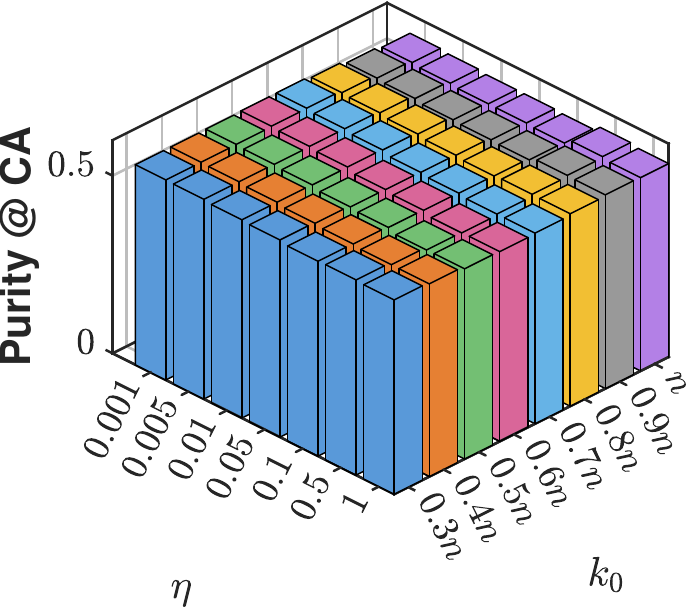}
    \includegraphics[width=0.24\linewidth]{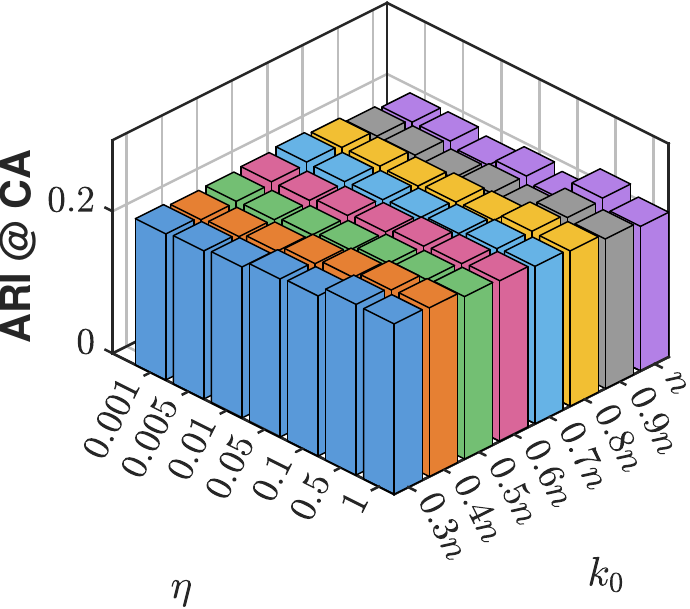}
    \includegraphics[width=0.24\linewidth]{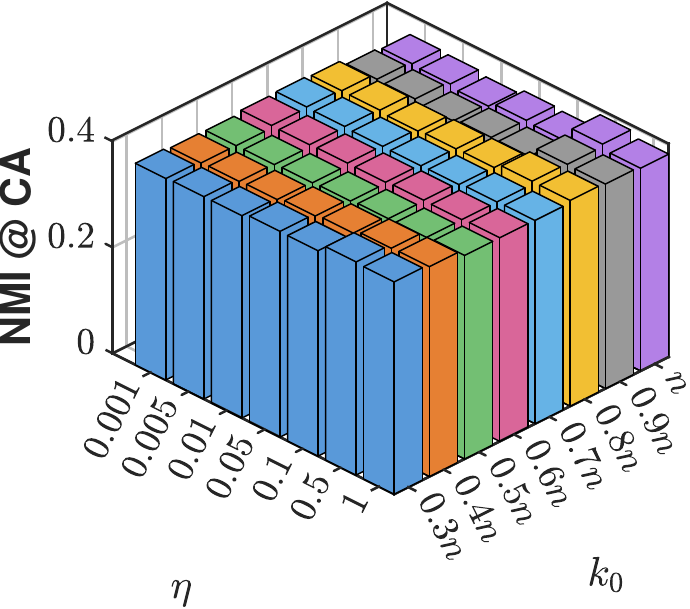}
    \includegraphics[width=0.24\linewidth]{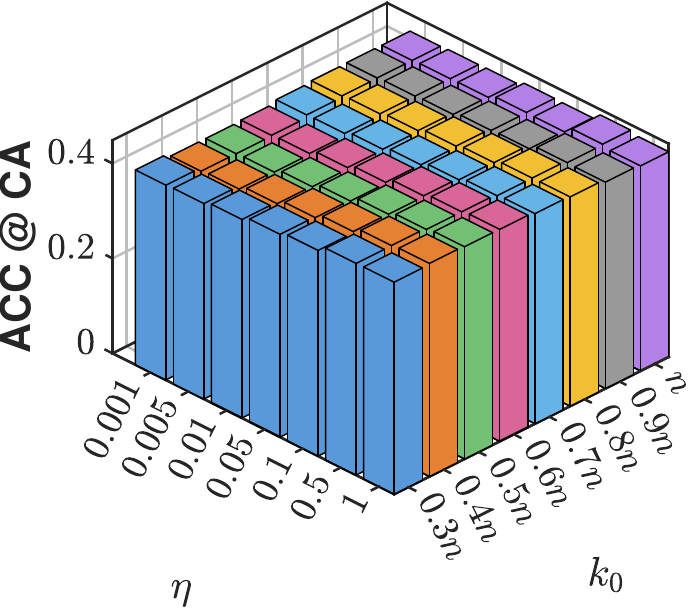}
    
    \caption{Performance of GOLD with different $\eta - k_0$ value combinations on six datasets.}
    \label{fig:hyper_appendix}
\end{figure*}

In GOLD, both FCPL and MCPL are governed by two hyperparameters, the learning rate $\eta$ and the initial number of clusters $k_0$. Because the competitive penalization relies on an over-specified initialization to activate effective split–merge dynamics, settings with too few initial clusters ($k_0 = 0.1n$ and $0.2n$) are omitted. As illustrated in Fig. \ref{fig:hyper_appendix}, across six representative datasets and four validity indices (Purity, ARI, NMI, and ACC), the performance surfaces display a broad, flat plateau, revealing that GOLD is largely insensitive to moderate changes of $\eta$ and $k_0$. Within the interior region, increasing $\eta$ from $0.01$ to $0.5$ and enlarging $k_0$ from roughly $0.4n$ to $0.8n$ yield only gentle variations, suggesting that the competitive penalties regularize optimization and prevent over-fragmentation at initialization. Degradation mainly occurs near the extremes. Very small step sizes (e.g., $\eta = 0.001$) slow down the update dynamics, making the procedure require many iterations and more easily settle in suboptimal partitions, which is reflected more noticeably by ARI and NMI due to their higher sensitivity to mismatched assignments. Conversely, an overly aggressive step (e.g., $\eta = 1$) tends to induce oscillatory behavior and unstable assignments in the coupled FCPL–MCPL updates, causing occasional drops across all metrics. Similarly, very large $k_0$ (e.g., $n$) introduces excessive prototype competition at the start; the subsequent pruning becomes volatile and delays consolidation, while too small $k_0$ (e.g., $0.3n$) deprives the model of the fine-grained structure needed for early discrimination and makes later refinement difficult. Despite dataset-specific score levels, the qualitative trends are consistent across all six datasets and all indices, reinforcing the conclusion that GOLD maintains robust clustering quality over a wide range of $\eta$ and $k_0$ and that only extreme configurations should be avoided in practice.

\end{document}